\documentclass{article}

\PassOptionsToPackage{numbers, compress}{natbib}

\usepackage[utf8]{inputenc} 
\usepackage[T1]{fontenc}    
\usepackage{etoc} 
\usepackage{microtype}      
\usepackage{xspace}

\usepackage{amsmath, amssymb, amsfonts} 
\usepackage{amsbsy}
\usepackage{nicefrac}       
\usepackage{utfsym}

\usepackage[dvipsnames,table]{xcolor} 
\usepackage{graphicx}
\usepackage{wrapfig}
\usepackage{contour}

\usepackage{booktabs}       
\usepackage{multirow}
\usepackage{array}
\usepackage{makecell}
\usepackage{arydshln}
\usepackage{float}
\usepackage{subcaption}
\usepackage{graphicx}
\newcolumntype{L}[1]{>{\raggedright\arraybackslash}m{#1}}

\newcolumntype{V}{!{\vrule width 0.6pt}}
\newcolumntype{D}{!{\vrule width 0.6pt\hspace{2pt}\vrule width 0.6pt}}
\newcolumntype{C}[1]{>{\centering\arraybackslash}m{#1}}
\usepackage{url}            
\usepackage[hidelinks]{hyperref}       
\usepackage{graphicx}
\usepackage{subcaption}
\usepackage[export]{adjustbox}

\definecolor{mygray}{gray}{0.9}

 \usepackage[main, final]{neurips_2026}

\usepackage{xcolor}

\newcommand{\OFBDUp}[1]{{\color{green!60!black}\uparrow\,#1}}
\newcommand{\OFBDDown}[1]{{\color{red}\downarrow\,#1}}

\newcommand{\ie}{\textit{i.e.}\xspace}
\newcommand{\eg}{\textit{e.g.}\xspace}

\title{OFBD: Object-Focused Background Debiasing for Long-Tailed Learning}

\author{%
\textbf{Shenghan Chen}\textsuperscript{1,2}
\thanks{This work was done while Shenghan Chen was a visiting student at Westlake University.}
\quad
\textbf{Yiming Liu}\textsuperscript{2}
\quad
\textbf{Zhipeng Deng}\textsuperscript{1}
\quad
\textbf{Haolin Wang}\textsuperscript{3}
\\
\textbf{Jiale Zhou}\textsuperscript{1}
\quad
\textbf{Zhijian Wu}\textsuperscript{1}
\quad
\textbf{Xiankai Lu}\textsuperscript{2}
\quad
\textbf{Yafei Ou}\textsuperscript{5}
\thanks{Yafei Ou and Yefeng Zheng are co-corresponding authors.}
\quad
\textbf{Yefeng Zheng}\textsuperscript{1}\footnotemark[2]
\\[1.5mm]
\textsuperscript{1}Westlake University, Hangzhou, China
\\
\textsuperscript{2}Shandong University, Jinan, China
\\
\textsuperscript{3}Hokkaido University, Sapporo, Japan
\\
\textsuperscript{5}RIKEN, Japan
}

\begin{document}

\maketitle

\begin{abstract}
  Balancing performance trade-offs on long-tailed data distributions remains a long-standing challenge in visual recognition. Existing methods mainly improve tail classes through re-balancing, representation learning, or data augmentation, but the underlying cause of tail class degradation is still insufficiently explored. In this paper, we find that standard long-tailed training induces background-biased representation and optimization: tail classes suffer larger background distribution shifts and become increasingly driven by background gradients. This reveals that tail degradation is not merely caused by insufficient samples, but also by the learning of irrelevant background features. To tackle this issue, we propose Object-Focused Background Debiasing (OFBD), a framework that mitigates background bias from both distribution and optimization perspectives. Specifically, Foreground-guided CutMix preserves target-related foregrounds while diversifying complementary backgrounds, and Background-guided Feature Rectification suppresses background-biased features without learnable parameters or additional training. Extensive experiments show that our method improves overall accuracy, achieves significant tail-class gains, and can serve as a plug-in for mainstream long-tailed methods without external data or pretrained recognition models. The code is available at: https://ofbd-neurips2026-longtail-learning.github.io/
\end{abstract}

\section{Introduction}
\label{Intro}

The prevailing paradigms in visual recognition ~\cite{lecun2015deep, voulodimos2018deep, esteva2021deep} owe much of their success to large-scale, carefully curated datasets~\cite{russakovsky2015imagenet, lin2014microsoft} with balanced class distributions. However, real-world data naturally exhibits highly imbalanced or long-tailed distributions~\cite{RAMW600}. When optimized on such skewed distributions, standard training protocols inevitably become biased toward data-abundant head classes, resulting in severe performance degradation on data-scarce tail classes. 

Existing methods addressing the class imbalance issue mainly fall into three paradigms. First, class re-balancing strategies such as resampling~\cite{liu2008exploratory, peng2019trainable} or reweighting~\cite{hong2021disentangling, cui2019class}, amplify tail signals but may overfit tail classes~\cite{shi2023re}. Second, architectural improvements correct feature and classifier bias via contrastive representation learning~\cite{zhu2022balanced, wang2021contrastive, kang2020exploring}; and via logit calibration~\cite{menonlong} or expert routing~\cite{wanglong,chen2023adamv,zhang2022self}. Third, information augmentation techniques synthesize or transfer semantic variations for tail classes~\cite{LLMAutoDA, shao2024diffult}, beyond conventional augmentations~\cite{alomar2023data} (\eg, cropping, flipping). Despite these improvements, existing methods mainly compensate for data quantity, leaving the degradation of tail representations during standard optimization~\cite{li2024feature} largely unexplored.

Interestingly, recent advances in self-supervised~\cite{jaiswal2020survey, liu2021self} and few-shot learning~\cite{wang2020generalizing} demonstrate that models can learn highly transferable representations from extremely limited data~\cite{dufew}. Such findings indicate that sample scarcity alone is insufficient to explain the profound representational collapse observed in tail classes. This paradox raises a critical question: \textbf{If limited data is not an inherent barrier, why do long-tailed models fail to capture the discriminative features of tail classes?}


\begin{figure}[t]
    \centering
    \vspace{-3em}
    \includegraphics[width=\linewidth]{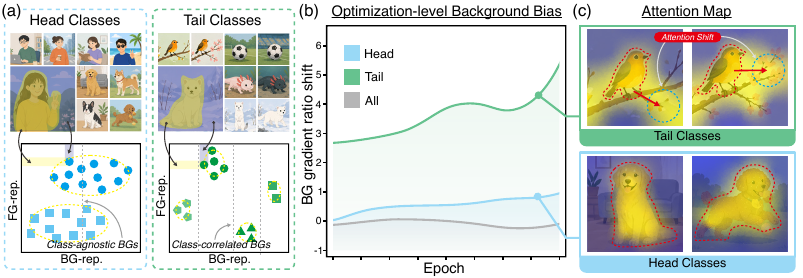}
    \caption{
    Motivation of OFBD. Tail-class degradation stems from both sample scarcity and background bias in distribution and optimization. (a) Head classes possess diverse, class-agnostic backgrounds (BGs), making background representations (BG-Rep.) uninformative for classification. Conversely, tail classes often co-occur with class-correlated BGs (\eg, snow fox in snowy scenes), which makes BG-Rep. a predictive shortcut and encourages reliance on BG cues over sparse foregrounds. Gray dashed lines denote BG-based decision boundaries. (b) Optimization-level background bias. ``Head'' and ``Tail'' represent the evaluation results under the long-tailed setting, while ``All'' indicates the overall result on the full balanced dataset. (c) Attention maps reveal that tail predictions co-activate foregrounds and backgrounds, whereas head predictions remain object-focused.
    }
    \label{fig:1}
\end{figure}

To investigate the underlying mechanism, we visualize the spatial representations learned by representative long-tailed networks~\cite{menonlong, zhu2022balanced}. As illustrated in Fig.~\ref{fig:1}(c), we observe a striking phenomenon: standard imbalanced training severely compromises the spatial inductive bias of the network, causing it to co-activate target foregrounds with irrelevant background regions. This empirical evidence reveals that tail-class degradation extends beyond simple data scarcity; crucially, it induces a profound background bias in both representation and optimization. Consequently, discriminative foreground features are marginalized by spurious contextual cues, leading to degradation on tail classes~\cite{tang2020long,mo2021object}. Therefore, resolving the long-tailed dilemma necessitates a dual approach: enhancing target-related foreground while explicitly suppressing the exploitation of irrelevant background evidence.

Suppressing irrelevant background evidence requires understanding its entanglement with targets. Building upon recent insights~\cite{mo2021object, shi2026vision}, we attribute this background bias to two structural forms of input-level co-occurrence: object-scene (\eg, fish with water) and object-object (\eg, racket with person) entanglements (Fig.~\ref{fig:1}(a)). These co-occurrences bias optimization toward spurious background shortcuts, resulting in background-biased representations and skewed BG feature distributions (Fig.~\ref{fig:1}(b)). We formalize this contextual and structural analysis in Sec.~\ref{sec: analysis}. Although spatial mixup strategies like CutMix~\cite{Yun_2019_ICCV} can disrupt these co-occurrences, prior works have shown that naive random replacements may introduce irrelevant noise~\cite{pan2024enhanced,jian2025supervised}. To tackle this issue, we propose Object-Focused Background Debiasing (OFBD), a framework that mitigates background bias from both distribution and optimization perspectives. Specifically, \textit{Foreground-guided CutMix} (FG-CutMix) uses a reinforcement-learning selector to handle discrete and non-differentiable region selection, preserving target-related foregrounds while replacing backgrounds to break foreground-background co-occurrence. Meanwhile, \textit{Background-guided Feature Rectification} (BFR) estimates background contribution from channel-wise feature statistics and down-weights background-biased spatial features in a parameter-free manner, avoiding head-class-dominated learnable rectification.

Overall, the main contributions of this paper are summarized as follows:
\begin{itemize}
\item \textbf{Novel Perspective:} We systematically investigate the inherent background bias in long-tailed recognition. Through structural analysis and empirical validation, we reveal how this bias shifts feature distributions and misguides optimization dynamics.

\item \textbf{Dual Debiasing Framework:} We propose FG-CutMix and BFR to address the two biases. FG-CutMix integrates reinforcement learning~\cite{shao2024deepseekmath} to preserve target foregrounds, while BFR rectifies features to reduce background reliance in a parameter-free manner.

\item \textbf{Superior Performance \& Plug-and-Play Versatility:} Extensive experiments demonstrate that without relying on external data or pretrained vision models, OFBD significantly boosts tail-class performance and overall accuracy. Furthermore, it inherently serves as a highly efficient plug-in module for existing mainstream long-tailed models.
\end{itemize}

\section{Related Work}
\label{Related}

\textbf{Long-tailed Learning: } Real-world data are often long-tailed~\cite{cui2019class,zhu2024rectify,narayan2025segface,li2025conmix,hong2021disentangling}. Existing methods include class re-balancing~\cite{hong2021disentangling,zhang2021learning,liu2008exploratory,ren2020balanced,shi2023parameter}, information augmentation~\cite{shao2024diffult,li2021metasaug,liu2021gistnet,wei2021crest,zhao2024ltgc}, and architectural improvements~\cite{huang2016learning,dong2017class,samuel2021distributional,wu2020solving,Song_2025_CVPR}. Yet they mainly compensate class quantity: re-balancing changes sample/loss weights, augmentation adds diversity without separating foreground features from background features, and architectures rarely rectify feature-level background bias. We instead mitigate distribution- and optimization-level background bias by decoupling target-related foreground features from spurious background features.

\textbf{Background Bias in Visual Recognition: }
Deep neural networks are notoriously prone to background bias, often relying on spurious contextual correlations (shortcuts) rather than intrinsic object semantics~\cite{geirhos2020shortcut, geirhos2018imagenet, xiaonoise, moon2021masker}. While various debiasing strategies have been proposed~\cite{minderer2020automatic, wang2019learning,kim2024discovering,li2023mitigating,chakraborty2024visual}, they predominantly assume balanced data distributions, leaving their behavior under severe class imbalance unexplored. Our work bridges this critical gap by revealing that standard long-tailed optimization inherently exacerbates these background biases. In response, we introduce a dual debiasing framework, offering a principled approach to robust and unbiased long-tailed recognition.

\textbf{RL-driven Data Augmentation: }
Reinforcement Learning (RL) has been used to search augmentation policies~\cite{cubuk2019autoaugment,lim2019fast,LLMAutoDA}, select transformation operations or magnitudes~\cite{hataya2020faster,dai2025auggpt}, and identify informative regions~\cite{wang2020glance,wang2025diversity,li2024empowering}. These methods optimize validation performance, augmentation strength, or region informativeness for general robustness, but are not tailored to long-tailed recognition, where tail classes are more vulnerable to spurious background reliance. In contrast, our RL selector is designed for FG-CutMix under long-tailed training. By selecting foreground regions that preserve target semantics after background replacement, the selector disrupts foreground-background co-occurrence and reduces tail-class background reliance.

\section{Analysis and Motivation}
\label{sec: analysis}
In this section, we analyze background bias from two complementary perspectives and provide empirical evidence. At the distribution level, we study how the learned background feature distribution deviates from that of the full balanced dataset. At the optimization level, we examine how training becomes increasingly driven by background-related gradients.

\subsection{Foreground-Background Decomposition}
\label{sec: analysis_b}
Following prior works~\cite{shetty2019not, qiao2019tell, wang2023context}, we decompose an input image $x$ into the target-related foreground and the target-irrelevant background, including co-occurring scenes and objects. Let $F=\Phi(x)\in\mathbb{R}^{D\times H\times W}$ be the feature map extracted by the encoder. Since foreground and background are spatial concepts, we first separate foreground and background regions on $F$ according to the decomposition of $x$, and then aggregate them into $f_{fg}$ and $f_{bg}$, respectively. The image-level features can be written as $f=f_{fg}+f_{bg}$, where $f_{fg}$ and $f_{bg}$ denote foreground and background features.

\subsection{Distribution-level Background Bias}
\label{sec: analysis_d}
For each class $y$, we use $p_y$ and $q_y$ to denote the background distributions under long-tailed and full-dataset training, respectively. The corresponding background feature means are $\mu_{bg}(y)=\mathbb{E}_{b\sim p_y}[f_{bg}]$ and $\bar{\mu}_{bg}(y)=\mathbb{E}_{b\sim q_y}[f_{bg}]$. We then define the background bias of class $y$ as $\Delta D_{bg}(y):=\mu_{bg}(y)-\bar{\mu}_{bg}(y)$, which measures the shift of the learned background feature mean from the full-dataset background feature mean. Since tail classes have far fewer samples than head classes, their empirical background distributions are more likely to deviate from $q_y$. Accordingly, we assume
$\|p_y^{\mathrm{tail}}-q_y\|_1 > \|p_y^{\mathrm{head}}-q_y\|_1$,
which yields a larger background distribution shift for tail classes:
\begin{equation}
\label{eq:gradient}
\|\Delta D_{bg}^{\mathrm{tail}}\|_1 >
\|\Delta D_{bg}^{\mathrm{head}}\|_1.
\end{equation}

\subsection{Optimization-level Background Bias}
\label{sec: analysis_o}
We next consider background bias from the optimization perspective. Let $g_t$ denote the gradient signal at epoch $t$. Consistent with the feature decomposition above, we first separate foreground and background regions on $F$, and then aggregate their gradient magnitudes into $g_{fg,t}$ and $g_{bg,t}$, respectively. We define the background gradient ratio and its shift for class $y$ as:
\begin{equation}
\label{eq:bg_ratio}
R_t(y)=\frac{G_{bg,t}(y)}{G_{fg,t}(y)+G_{bg,t}(y)},\quad
R_0(y)=\frac{1}{T_0}\sum_{t=1}^{T_0}R_t(y),\quad
\Delta R_t(y)=R_t(y)-R_0(y),
\end{equation}
where $G_{fg,t}(y)=\|g_{fg,t}(y)\|_1$ and $G_{bg,t}(y)=\|g_{bg,t}(y)\|_1$. Since the absolute value of $R_t(y)$ can be affected by the initial foreground-background composition of each class, $\Delta R_t(y)$ better captures whether optimization progressively leans toward background features. Under long-tailed training, tail classes receive weaker foreground supervision, while background cues are easier to exploit and continue to contribute to the loss. We therefore assume:
\begin{equation}
\label{eq:bg_ratio_a}
\Delta R_t^{\mathrm{tail}}>\Delta R_t^{\mathrm{head}}.
\end{equation}
Detailed derivations and assumptions motivating these hypotheses (Eq.~\ref{eq:gradient} and Eq.~\ref{eq:bg_ratio_a}) are provided in the Appendix~\ref{sec:derivations}. We empirically validate these claims below.

\subsection{Empirical Evidence}
\label{Evidence}

To validate the above analysis, we empirically examine background bias from both distribution and optimization perspectives. Visualized results and experimental details are provided in Appendix~\ref{sec:bg_bias_verification}.

\textbf{Observation 1}: Tail classes suffer larger distribution-level shift.
Fig.~\ref{fig:bg_distribution_shift} shows larger tail-class background distribution shift than head classes due to limited samples and less diverse backgrounds, motivating FG-CutMix to diversify backgrounds while preserving target foregrounds.

\textbf{Observation 2}: Long-tailed training amplifies optimization-level bias.
Fig.~\ref{fig:bg_gradient_shift} shows a larger background-gradient-ratio increase under long-tailed training, especially for tail classes. This indicates that insufficient and imbalanced foreground supervision shifts optimization toward easily exploitable background features, motivating BFR to suppress background-biased gradients.

\textbf{Observation 3}: Tail-class background influence is more unstable.
Fig.~\ref{fig:bg_gradient_shift} further shows stronger fluctuations in tail-class background-gradient ratios, motivating parameter-free BFR and the reference selector in FG-CutMix for stable debiasing and foreground selection.


\section{Method}
\label{sec: method}

Guided by Sec.~\ref{sec: analysis}, our framework addresses background bias from two aspects. FG-CutMix targets the distribution-level shift $\Delta D_{bg}$ by preserving foregrounds while replacing backgrounds. BFR targets the optimization-level shift $\Delta R_t$ by down-weighting background-biased features. Together, they mitigate biased background distributions and background-driven optimization in long-tailed training. We provide a theoretical interpretation of the effectiveness of both modules in Appendix~\ref{sec:ofbd_bias_mitigation}.


\subsection{Foreground-guided CutMix}
Unlike naive CutMix, which may exacerbate bias by sampling background-only regions~\cite{mo2021object}, our Foreground-guided CutMix (FG-CutMix) explicitly preserves target foregrounds. Furthermore, its RL-based selector introduces minimal computational overhead, ensuring practicality for large-scale training. Detailed experiments evaluating the computational cost are provided in Appendix~\ref{sec:efficiency_analysis}.


\subsubsection{RL-based Foreground Region Selection}
The key challenge in FG-CutMix is selecting foreground regions effective for background replacement. CAM-based~\cite{zhou2016learning,jung2021towards} and SAM-based~\cite{kirillov2023segment,ravisam} strategies can localize complete foregrounds, but accurate localization may not yield the best mixing region. Since CutMix benefits from partial-view recognition and regional perturbation~\cite{Yun_2019_ICCV}, effective regions should balance target preservation and background replacement, as overly complete regions may weaken background variation and regularization. Since foreground-region selection is discrete and non-differentiable, we use an RL-based selector to learn an adaptive policy, capture class-specific foreground importance.

\textbf{RL-based foreground candidate generation.}
Given the feature map $F=\Phi(x)\in\mathbb{R}^{D\times H\times W}$, we first randomly generate a region proposal set $\mathcal{P}(x)=\{b_1,\dots,b_N\}$ with different locations and scales. Each proposal $b_i$ corresponds to a candidate crop region on the input image. Instead of ranking these proposals by local energy~\cite{wang2025mixa, Huy_2025_ICCV}, we introduce an RL-based region selector $\pi_\theta$ to select $K$ proposals from $\mathcal{P}(x)$ as foreground candidates. Specifically, $\pi_\theta$ predicts a selection probability over $\mathcal{P}(x)$, and $K$ proposals are sampled without replacement to form the foreground candidate pool (FG candidate pool) $\mathcal{R}_\theta(x)=\{r_1,\dots,r_k,\dots,r_K\}$. This allows the selector to learn which randomly generated regions are more suitable for preserving target-related foregrounds in FG-CutMix.

\textbf{Reward design.}
Given the selected candidate pool, we construct mixed samples 
$\{\tilde{x}_1,\dots,\tilde{x}_K\}$ and obtain class prediction probabilities 
$p_k\in[0,1]^C$ from the classifier. Let $y$ and $y^{\mathrm{bg}}$ denote the target label and the background-source label, respectively. The reward for candidate $r_k$ is defined as:
\begin{equation}
\label{eq:reward}
\rho_k=\mathbf{1}\!\left(p_k[y] > p_k[y^{\mathrm{bg}}]\right).
\end{equation}
Here, $p_k$ provides the reward signal and $o_k=\pi_\theta(r_k\mid F)$ denotes the selector confidence, encouraging regions that preserve target semantics over background-induced semantics.

\textbf{Policy learning.}
The region selector $\pi_\theta$ defines a selection distribution over the proposal set $\mathcal{P}(x)$ and samples $K$ proposals $\mathcal{R}_\theta(x)$ as foreground candidates. Since proposal selection is discrete and non-differentiable, we optimize $\pi_\theta$ via policy gradient~\cite{konda1999actor,schulman2017proximal,wu2021coordinated}. Since selected candidates within each image are not directly comparable due to varying classifier confidence~\cite{zhang2023deep} and potential background bias, we follow GRPO~\cite{shao2024deepseekmath} and normalize rewards within each candidate pool. Since tail-class background gradients fluctuate more (Sec.~\ref{Evidence}), we use the previous-epoch selector as a frozen reference $\pi_{\mathrm{ref}}$ to stabilize foreground selection. The policy learning objective is:
\begin{equation}
\label{eq:rl_loss}
\begin{gathered}
\mathcal{L}_{\mathrm{RL}}
=
-\frac{1}{K}\sum_{k=1}^{K}
\left[
\min\!\left(\hat{s}_{1,k}A_k,\hat{s}_{2,k}A_k\right)
-\beta \mathcal{D}_{\mathrm{KL}}\!\left(\pi_\theta\|\pi_{\mathrm{ref}}\right)
-\alpha \mathrm{CE}(o_k,\rho_k)
\right], \\[4pt]
\mathcal{D}_{\mathrm{KL}}\!\left(\pi_\theta\|\pi_{\mathrm{ref}}\right)
=
\frac{\pi_{\mathrm{ref}}(r_k\mid F)}
{\pi_\theta(r_k\mid F)}
-\log
\frac{\pi_{\mathrm{ref}}(r_k\mid F)}
{\pi_\theta(r_k\mid F)}
-1, \\[4pt]
\hat{s}_{1,k}
=
\frac{\pi_\theta(r_k\mid F)}
{\pi_{\mathrm{ref}}(r_k\mid F)},
\qquad
\hat{s}_{2,k}
=
\mathrm{clip}
\left(
\frac{\pi_\theta(r_k\mid F)}
{\pi_{\mathrm{ref}}(r_k\mid F)},
1-\epsilon_c,
1+\epsilon_c
\right), \\[4pt]
A_k
=
\frac{\rho_k-\mathrm{mean}(\rho_1,\ldots,\rho_K)}
{\mathrm{std}(\rho_1,\ldots,\rho_K)+\epsilon},
\qquad
o_k=\pi_\theta(r_k\mid F),
\end{gathered}
\end{equation}
where $\theta$ denotes the parameters of the region selector $\pi_\theta$, $\epsilon_c$ clamps the policy ratio, and $\beta$ controls the KL penalty. Following prior works~\cite{schulman2017proximal,shao2024deepseekmath}, the frozen reference selector $\pi_{\mathrm{ref}}$ and KL term stabilize policy learning by limiting abrupt policy shifts, while the clipped objective prioritizes high-advantage foreground candidates without overly large updates. The auxiliary term, weighted by $\alpha$, encourages region selector confidence to align with the reward signal.

\begin{figure}[t]
    \centering
    \vspace{-3em}
    \includegraphics[width=\linewidth]{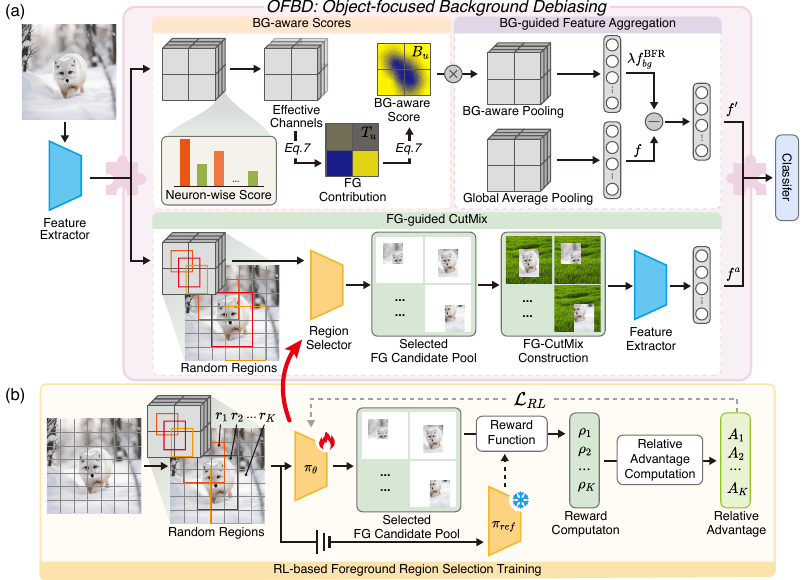}
    \caption{
    Overview of OFBD. The pink box denotes OFBD as a plug-in module for mainstream long-tailed models. (a) FG-CutMix selects target-related foregrounds from random regions and replaces complementary backgrounds, while BFR estimates background-aware scores to rectify features before classification. (b) The region selector $\pi_\theta$ is optimized via reinforcement learning, and a frozen reference selector $\pi_{\mathrm{ref}}$ stabilizes the training of the selector.
}
    \label{fig:2}
\end{figure}

\subsubsection{FG-CutMix Sample Construction}
Given a selected foreground candidate $r_k\in\mathcal{R}(x)$, we preserve the target-related region in $x$ and replace its complement with content from another image $x'$. Let $M_k$ be the binary mask of $r_k$, where $M_k=1$ denotes the preserved region, and $M_k=0$ denotes the complement. The mixed sample is $\tilde{x}_k := M_k \odot x + (1-M_k)\odot x'$, where $\odot$ denotes element-wise multiplication. This preserves target foregrounds, varies surrounding backgrounds, and reduces foreground-background co-occurrence. Following prior works~\cite{pan2024enhanced,liu2025long}, we correct the mixed label as $\tilde{y}_k$ for reliable supervision.


\subsection{Background-guided Feature Rectification}

Although FG-CutMix alleviates distribution-level background bias by restoring background diversity, optimization may still favor irrelevant background cues that are easier to exploit than sparse tail-class foreground evidence. We therefore introduce Background-guided Feature Rectification (BFR), which uses background-aware scores to down-weight biased feature locations in the final representation. To avoid head-class-dominated gradients from learnable parameters, BFR is strictly parameter-free. Inspired by SimAM~\cite{yang2021simam} and LaSt-ViT~\cite{shi2026vision}, BFR estimates foreground contribution from channel-wise feature statistics: larger normalized deviations from stable channel means indicate higher foreground contribution. Effective channel aggregation further prevents sparse tail-class foreground channels from being diluted and derives background-aware scores for rectification.

\subsubsection{Background-aware Scores}

Given the feature map $F=\Phi(x)\in\mathbb{R}^{D\times H\times W}$, let $u$ denote a spatial location and $F_{d,u}$ denote the activation at channel $d$ and location $u$, we compute a neuron-wise score $S_{d,u}$:
\begin{equation}
\label{eq:energy}
S_{d,u}
=
\mathrm{sigmoid}\left(
\frac{(F_{d,u}-\mu_d)^2}{4(\sigma_d^2+\epsilon)}
\right),
\qquad
\bar{S}_u=\frac{1}{D}\sum_{d=1}^{D}S_{d,u},
\end{equation}
where $\mu_d$ and $\sigma_d^2$ are the mean and variance of channel $d$ over spatial locations, respectively. The detailed derivation is provided in Appendix~\ref{sec:score_derivation}. We then retain effective channels $\mathcal{C}_u=\{d\mid S_{d,u}\geq\bar{S}_u\}$ and compute the foreground contribution $T_u$ and background-aware score $B_u$ as:
\begin{equation}
\label{eq:background score}
T_u=
\frac{1}{|\mathcal{C}_u|}
\sum_{d\in\mathcal{C}_u}S_{d,u},
\qquad
B_u=
\mathrm{sigmoid}\left(
\frac{\bar{T}-T_u}{\sigma_T+\epsilon}
\right),
\end{equation}
where $\bar{T}$ and $\sigma_T$ are the mean and standard deviation of $T_u$ over all feature locations, respectively. A larger $B_u$ indicates stronger background bias at location $u$.

\subsubsection{Background-guided Feature Aggregation}

Based on the background-aware score $B_u$, we estimate and remove the background-related component from the image-level feature. Let $f$ denote the standard image-level feature:
\begin{equation}
f=\frac{1}{HW}\sum_{u}F_u,
\qquad
f_{bg}^{\mathrm{BFR}}=\frac{1}{HW}\sum_{u}B_uF_u.
\end{equation}
Here, $f_{bg}^{\mathrm{BFR}}$ denotes the estimated background-biased component. The rectified feature is then obtained by removing this estimated background-biased feature:
\begin{equation}
\label{eq:fr}
f'=f-\lambda f_{bg}^{\mathrm{BFR}}
=
\frac{1}{HW}\sum_{u}(1-\lambda B_u)F_u,
\end{equation}
where $\lambda$ controls the strength of background rectification. In this way, BFR softly down-weights background-biased feature locations instead of relying on a hard foreground-background mask.

Notably, BFR introduces no learnable parameters and \textbf{does not require additional training}, making it less affected by label imbalance in long-tailed recognition. By relying on feature statistics rather than extra supervision, BFR provides a lightweight way to rectify background-biased features.

\subsection{Training Objective}
For a given training sample $(x, y)$, FG-CutMix generates $K$ augmented samples and aggregates them into a single feature $f^a$ with target label $\tilde{y}$. Meanwhile, BFR produces the rectified feature $f'$ for the original sample $x$. The overall objective is:
\begin{equation}
\label{eq:objective}
\mathcal{L}
=
\mathcal{L}_{\mathrm{cls}}
+
\gamma \mathcal{L}_{\mathrm{RL}},
\quad
\mathcal{L}_{\mathrm{cls}}
=
\frac{1}{2}
\left[
\ell_{\mathrm{LT}}\big(h(f^{a}),\tilde{y}\big)
+
\ell_{\mathrm{LT}}\big(h(f'),y\big)
\right],
\end{equation}
where $h(\cdot)$ is the classifier head, $\ell_{\mathrm{LT}}$ denotes the underlying long-tailed objective, $\mathcal{L}_{\mathrm{RL}}$ is the reinforcement learning loss (Eq.~\ref{eq:rl_loss}), and $\gamma$ controls the strength of policy learning.

\begin{table*}[t]
\centering
\caption{Comparisons on CIFAR-LT~\cite{cao2019learning} datasets with varying imbalance ratios ($r$). Results are reported across all classes as well as different groups based on training sample sizes ("Many", "Med", and "Few"). For each method, the first row reports the original baseline accuracy, and the second row shows the performance after applying OFBD.}
\label{tab:sel_cifar_lt_combined}
\scriptsize
\renewcommand{\arraystretch}{1.06}
\setlength{\tabcolsep}{6pt}
\resizebox{\textwidth}{!}{
\begin{tabular}{l|ccc|ccc|ccc}
\toprule
& \multicolumn{3}{c|}{CIFAR-100-LT} 
& \multicolumn{3}{c|}{CIFAR-10-LT}
& \multicolumn{3}{c}{Statistic$(r=100)$} \\
\cmidrule(lr){2-4} \cmidrule(lr){5-7} \cmidrule(lr){8-10}
Method
& r=100$\uparrow$ & r=50$\uparrow$ & r=10$\uparrow$
& r=100$\uparrow$ & r=50$\uparrow$ & r=10$\uparrow$
& Many$\uparrow$ & Med.$\uparrow$ & Few$\uparrow$ \\
\midrule

CE
& 38.32 & 43.94 & 55.71
& 70.45 & 74.82 & 86.44
& 65.19 & 37.06 & 9.14 \\
\rowcolor{gray!18}
\quad +OFBD
& $44.83_{\OFBDUp{6.51}}$
& $47.79_{\OFBDUp{3.85}}$
& $58.22_{\OFBDUp{2.51}}$
& $74.33_{\OFBDUp{3.88}}$
& $78.68_{\OFBDUp{3.86}}$
& $87.30_{\OFBDUp{0.86}}$
& $65.48_{\OFBDUp{0.29}}$
& $42.09_{\OFBDUp{5.03}}$
& $23.93_{\OFBDUp{14.79}}$ \\

LDAM-DRW~\cite{cao2019learning} {\color{gray}(NeurIPS'19)}
& 42.09 & 46.57 & 58.64
& 77.06 & 81.09 & 88.02
& 61.51 & 41.67 & 20.22 \\
\rowcolor{gray!18}
\quad +OFBD
& $45.81_{\OFBDUp{3.72}}$
& $49.55_{\OFBDUp{2.98}}$
& $59.09_{\OFBDUp{0.45}}$
& $79.61_{\OFBDUp{2.55}}$
& $82.43_{\OFBDUp{1.34}}$
& $88.76_{\OFBDUp{0.74}}$
& $64.93_{\OFBDUp{3.42}}$
& $45.81_{\OFBDUp{4.14}}$
& $26.61_{\OFBDUp{6.39}}$ \\

BBN~\cite{zhou2020bbn} {\color{gray}(CVPR'20)}
& 42.45 & 47.13 & 59.23
& 79.78 & 81.24 & 88.32
& 42.63 & 50.70 & 32.60 \\
\rowcolor{gray!18}
\quad +OFBD
& $44.81_{\OFBDUp{2.36}}$
& $48.32_{\OFBDUp{1.19}}$
& $60.57_{\OFBDUp{1.34}}$
& $80.69_{\OFBDUp{0.91}}$
& $84.41_{\OFBDUp{3.17}}$
& $89.28_{\OFBDUp{0.96}}$
& $43.94_{\OFBDUp{1.31}}$
& $51.83_{\OFBDUp{1.13}}$
& $37.64_{\OFBDUp{5.04}}$ \\

BCL~\cite{zhu2022balanced} {\color{gray}(CVPR'22)}
& 51.84 & 56.57 & 64.03
& 84.31 & 87.21 & 90.92
& 66.85 & 52.54 & 33.07 \\
\rowcolor{gray!18}
\quad +OFBD
& $54.19_{\OFBDUp{2.35}}$
& $57.94_{\OFBDUp{1.37}}$
& $64.91_{\OFBDUp{0.88}}$
& $86.34_{\OFBDUp{2.03}}$
& $88.12_{\OFBDUp{0.91}}$
& $91.63_{\OFBDUp{0.71}}$
& $66.79_{\OFBDDown{0.06}}$
& $53.46_{\OFBDUp{0.92}}$
& $40.34_{\OFBDUp{7.27}}$ \\

SBCL~\cite{hou2023subclass} {\color{gray}(CVPR'23)}
& 44.93 & 48.65 & 57.83
& 74.84 & 80.05 & 84.48
& 64.35 & 45.27 & 22.14 \\
\rowcolor{gray!18}
\quad +OFBD
& $46.92_{\OFBDUp{1.99}}$
& $49.08_{\OFBDUp{0.43}}$
& $58.44_{\OFBDUp{0.61}}$
& $76.52_{\OFBDUp{1.68}}$
& $81.43_{\OFBDUp{1.38}}$
& $84.89_{\OFBDUp{0.41}}$
& $65.26_{\OFBDUp{0.91}}$
& $45.35_{\OFBDUp{0.08}}$
& $27.36_{\OFBDUp{5.22}}$ \\

GBG~\cite{li2024long} {\color{gray}(AAAI'24)}
& 51.92 & 56.92 & 64.32
& 85.12 & 87.64 & 91.23
& 66.94 & 53.15 & 32.96 \\
\rowcolor{gray!18}
\quad +OFBD       
& $54.23_{\OFBDUp{2.31}}$
& $58.37_{\OFBDUp{1.45}}$
& $64.98_{\OFBDUp{0.66}}$
& $86.57_{\OFBDUp{1.45}}$
& $88.25_{\OFBDUp{0.61}}$
& $91.61_{\OFBDUp{0.38}}$
& $66.81_{\OFBDDown{0.13}}$
& $53.57_{\OFBDUp{0.42}}$
& $40.32_{\OFBDUp{7.36}}$ \\

MKP~\cite{chen2026reframing} {\color{gray}(CVPR'26)}
& 53.21 & 57.63 & 68.74
& 86.31 & 88.26 & 92.53
& 67.28 & 54.86 & 34.91 \\
\rowcolor{gray!18}
\quad +OFBD
& $54.42_{\OFBDUp{1.21}}$
& $58.41_{\OFBDUp{0.78}}$
& $68.12_{\OFBDDown{0.62}}$
& $86.78_{\OFBDUp{0.47}}$
& $88.42_{\OFBDUp{0.16}}$
& $92.02_{\OFBDDown{0.51}}$
& $67.57_{\OFBDUp{0.29}}$
& $55.95_{\OFBDUp{1.09}}$
& $37.29_{\OFBDUp{2.38}}$ \\

\bottomrule
\end{tabular}
}
\end{table*}

\section{Experiments}
\label{experiment}
\subsection{Experiment Setup}
\label{sec:exp_details}


\textbf{Datasets and Metrics.} Our proposed framework is evaluated on four long-tailed benchmarks: CIFAR-10-LT~\cite{cao2019learning}, CIFAR-100-LT~\cite{cao2019learning}, ImageNet-LT~\cite{liu2019large}, and iNaturalist 2018~\cite{van2018inaturalist}. Following standard evaluation protocols~\cite{LLMAutoDA,zhao2025learning}, we report Top-1 accuracy. For CIFAR datasets, we evaluate effectiveness of OFBD across three imbalance ratios ($r \in \{100, 50, 10\}$) and decompose results into ``Many'' ($>100$ samples), ``Med.'' ($20 \sim 100$), and ``Few'' ($<20$) categories for further analysis. For ImageNet-LT and iNaturalist 2018, we assess various backbones to verify architecture generality.

\textbf{Compared Methods.} 
To demonstrate the effectiveness and plug-and-play versatility of OFBD, we integrate it into representative, open-source long-tailed methods to ensure reproducibility. These span standard training (CE), re-balancing techniques (LDAM-DRW~\cite{cao2019learning}, BBN~\cite{zhou2020bbn}), contrastive learning (SBCL~\cite{hou2023subclass}, BCL~\cite{zhu2022balanced}), and advanced representation optimization (GBG~\cite{li2024long}, MKP~\cite{chen2026reframing}). By consistently using standard ResNet backbones, this integration fairly and directly validates OFBD's ability to boost performance by mitigating background bias across diverse learning paradigms.

\textbf{Implementation.} 
For fairness, we reproduce baselines with official codes and fixed seeds. We use ResNet-32 for CIFAR10/100-LT~\cite{cao2019learning}, ResNet-50~\cite{he2016deep} for ImageNet-LT and iNaturalist 2018, and ViT-B/16~\cite{dosovitskiyimage} for cross-architecture evaluation. All models are trained on an RTX 4090 GPU with a batch size of 256. For OFBD, we set $\lambda=0.5$, $\gamma=0.8$, $\alpha=1.0$, and $\beta=0.02$ across datasets. Extensive ablation studies, further analyses, and implementation details are deferred to Appendix~\ref{sec:add_experiments}.

\subsection{Effectiveness as a Plug-in Module}
\label{sec:main_results}
\textbf{CIFAR100-LT and CIFAR10-LT~\cite{cao2019learning}.}
Table ~\ref{tab:sel_cifar_lt_combined} reports results on CIFAR-10/100-LT after integrating OFBD into existing methods. Across 42 experimental settings, OFBD improves 40 of them, with the largest gains of $+6.51\%$ on CIFAR-100-LT and $+3.88\%$ on CIFAR-10-LT. For the standard CE baseline, OFBD improves CIFAR-100-LT by $+6.51\%$, $+3.85\%$, and $+2.51\%$ under $r=100$, $50$, and $10$, respectively, and improves CIFAR-10-LT by $+3.88\%$, $+3.86\%$, and $+0.86\%$. For re-balancing methods, OFBD improves LDAM-DRW by up to $+3.72\%$ and BBN by up to $+2.36\%$ on CIFAR-100-LT. For contrastive-learning methods, OFBD improves BCL and SBCL by up to $+2.35\%$ and $+1.99\%$, respectively. For representation-optimization methods, OFBD improves GBG by up to $+2.31\%$ and MKP by $+1.21\%$ under high imbalance. These consistent gains verify the plug-in effectiveness of OFBD across diverse long-tailed learning paradigms.

The Many/Med./Few statistics on CIFAR-100-LT with $r=100$ further show that OFBD mainly benefits data-scarce classes while preserving head-class performance. OFBD improves Few-class accuracy by $+14.79\%$ on CE, $+6.39\%$ on LDAM-DRW, $+5.04\%$ on BBN, $+7.27\%$ on BCL, $+5.22\%$ on SBCL, $+7.36\%$ on GBG, and $+2.38\%$ on MKP. In contrast, Many-class changes remain small for strong baselines (\eg, $-0.06\%$ on BCL, $-0.13\%$ on GBG, and $+0.29\%$ on MKP). These results indicate that OFBD alleviates tail-class degradation without sacrificing head-class recognition.

\begin{table*}[t]
\centering
\scriptsize
\renewcommand{\arraystretch}{1.05}

\begin{minipage}[t]{0.43\textwidth}
\centering
\captionof{table}{Top-1 accuracy on ImageNet-LT~\cite{liu2019large} and iNaturalist 2018~\cite{van2018inaturalist}.}
\label{tab:imagenet_inat_vit_results}
\setlength{\tabcolsep}{3pt}
\begin{tabular}{L{2.46cm}C{1.55cm} C{1.35cm}}
\toprule
Method & ImageNet-LT & iNature2018\\
\midrule

\multicolumn{3}{c}{\textit{ResNet-50~\cite{he2016deep} backbone}} \\
\midrule
CE
& 44.80 & 65.95 \\
\rowcolor{gray!18}
\quad +OFBD & $47.39_{\OFBDUp{2.59}}$ & $69.76_{\OFBDUp{3.81}}$ \\

BCL~\cite{zhu2022balanced} {\color{gray}(CVPR'22)}
& 55.67 & 71.81 \\
\rowcolor{gray!18}
\quad +OFBD & $56.81_{\OFBDUp{1.14}}$ & $72.34_{\OFBDUp{0.53}}$ \\

GCL~\cite{li2022long} {\color{gray}(CVPR'22)}
& 55.46 & 70.76 \\
\rowcolor{gray!18}
\quad +OFBD & $56.34_{\OFBDUp{0.88}}$ & $71.92_{\OFBDUp{1.16}}$ \\

GBG~\cite{li2024long} {\color{gray}(AAAI'24)}
& 57.13 & 71.92 \\
\rowcolor{gray!18}
\quad +OFBD & $57.64_{\OFBDUp{0.51}}$ & $72.36_{\OFBDUp{0.44}}$ \\

MKP~\cite{chen2026reframing} {\color{gray}(CVPR'26)}
& 57.92 & 74.41 \\
\rowcolor{gray!18}
\quad +OFBD & $58.51_{\OFBDUp{0.59}}$ & $74.63_{\OFBDUp{0.22}}$ \\

\midrule
\multicolumn{3}{c}{\textit{ViT-B~\cite{dosovitskiyimage} backbone}} \\
\midrule
ViT~\cite{dosovitskiyimage} {\color{gray}(ICLR'21)}
& 37.51 & 54.23 \\
\rowcolor{gray!18}
\quad +OFBD & $39.17_{\OFBDUp{1.66}}$ & $56.62_{\OFBDUp{2.39}}$ \\

DeiT-LT~\cite{rangwani2024deit} {\color{gray}(CVPR'24)}
& 55.64 & 72.92 \\
\rowcolor{gray!18}
\quad +OFBD & $56.21_{\OFBDUp{0.57}}$ & $74.56_{\OFBDUp{1.64}}$ \\

\bottomrule
\end{tabular}
\end{minipage}
\hfill
\begin{minipage}[t]{0.52\textwidth}
\centering
\captionof{table}{Ablation study on main components.}
\label{tab:ablation_modules}
\setlength{\tabcolsep}{3pt}
\begin{tabular}{l@{\hspace{4pt}\vrule\hspace{2pt}\vrule\hspace{4pt}}c c c c}
\toprule
Method & Many$\uparrow$ & Med.$\uparrow$ & Few$\uparrow$ & All$\uparrow$ \\
\midrule

CE & 65.12 & 36.60 & 9.06 & 38.32 \\
\rowcolor{gray!12}
+FG-CutMix+BFR & 65.48 & 42.09 & 23.93 & $44.83_{\OFBDUp{6.51}}$ \\

\midrule

BCL & 66.85 & 52.92 & 33.07 & 51.84 \\
+FG-CutMix & 67.11 & 52.11 & 38.84 & $53.38_{\OFBDUp{1.54}}$ \\
+BFR & 66.96 & 53.65 & 35.86 & $52.97_{\OFBDUp{1.13}}$ \\
\rowcolor{gray!18}
+FG-CutMix+BFR & 66.79 & 53.46 & 40.34 & $54.19_{\OFBDUp{2.35}}$ \\

\bottomrule
\end{tabular}

\vspace{1em}

\captionof{table}{Further analysis of CutMix region selection. All variants use the same CutMix pipeline but differ in how the preserved region is selected.}
\label{tab:rl_selection_ablation}
\setlength{\tabcolsep}{2pt}
\begin{tabular}{>{\raggedright\arraybackslash}p{0.65\linewidth}@{\hspace{4pt}\vrule width 0.45pt\hspace{2pt}\vrule width 0.45pt\hspace{4pt}}c}
\toprule
CutMix Variant & Acc. (\%) \\
\midrule
Standard CutMix (random selection) & 52.17 \\
CutMix w/ SAM-based foreground selection& 52.43 \\
CutMix w/ CAM-based foreground selection& 52.63 \\
\midrule
\rowcolor{gray!18}
FG-CutMix w/ RL-based foreground selection& \textbf{54.19} \\
\bottomrule
\end{tabular}
\end{minipage}

\end{table*}

\noindent \textbf{ImageNet-LT and iNaturalist 2018~\cite{liu2019large,van2018inaturalist}.}
Table~\ref{tab:imagenet_inat_vit_results} reports results on large-scale long-tailed datasets with both ResNet-50 and ViT-B backbones. With ResNet-50, OFBD consistently improves all baselines on both datasets. The gains are particularly large for the vanilla CE baseline, improving on ImageNet-LT from $44.80\%$ to $47.39\%$ ($+2.59\%$) and on iNaturalist 2018 from $65.95\%$ to $69.76\%$ ($+3.81\%$), showing that OFBD effectively complements standard training. For stronger baselines, OFBD still brings stable gains: $+1.14\%$, $+0.88\%$, $+0.51\%$, and $+0.59\%$ on ImageNet-LT for BCL, GCL, GBG, and MKP, and $+0.53\%$, $+1.16\%$, $+0.44\%$, and $+0.22\%$ on iNaturalist 2018. With ViT-B, OFBD also improves ViT and DeiT-LT by $+1.66\%$/$+0.57\%$ on ImageNet-LT and $+2.39\%$/$+1.64\%$ on iNaturalist 2018, respectively. These consistent improvements across datasets, baselines, and CNN/Transformer backbones verify the generality of OFBD as a plug-in framework.

\subsection{Ablation Study and Analysis}
\label{sec: ablation}
In this section, we present key ablation and analysis experiments to characterize the effectiveness, design choices, and debiasing behavior of OFBD. More experiments are provided in Appendix~\ref{sec:add_experiments}.

\noindent\textbf{Component Analysis.}
We conduct component ablation on CIFAR100-LT ($r=100$) with ResNet-32. We choose CE and BCL~\cite{zhu2022balanced} as representative baselines: CE reflects standard long-tailed training, while BCL represents a strong contrastive-learning baseline. As shown in Table~\ref{tab:ablation_modules}, adding the full FG-CutMix+BFR framework to CE improves the overall accuracy from $38.32\%$ to $44.83\%$ ($+6.51\%$), mainly due to the large Few-class gain from $9.06\%$ to $23.93\%$. On BCL, FG-CutMix and BFR improve the overall accuracy by $+1.54\%$ and $+1.13\%$, respectively. Combining them further improves BCL from $51.84\%$ to $54.19\%$ ($+2.35\%$), with Few-class accuracy increasing from $33.07\%$ to $40.34\%$. These results show that both components are effective, especially for tail classes.


\begin{figure*}[t] 
  \centering
  
 \begin{subfigure}[b]{0.54\textwidth}
    \centering
    \setlength{\tabcolsep}{1pt} 
    \renewcommand{\arraystretch}{1.2} 
    
    \footnotesize 
    
    \begin{tabular}{ccccc}
      & Salamander & Mousetrap & Cowboy Boot & Pool Table\\
      \noalign{\vspace{2mm}} 
      
      \rotatebox[origin=c]{90}{CE} & 
      \raisebox{-0.5\height}{\includegraphics[width=0.23\linewidth]{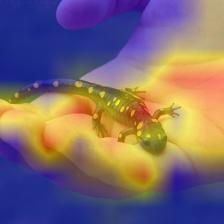}} & 
      \raisebox{-0.5\height}{\includegraphics[width=0.23\linewidth]{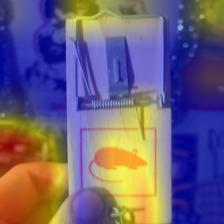}} & 
      \raisebox{-0.5\height}{\includegraphics[width=0.23\linewidth]{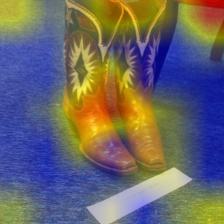}} &
      \raisebox{-0.5\height}{\includegraphics[width=0.23\linewidth]{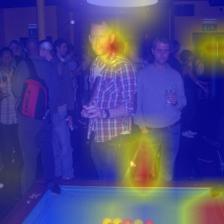}}\\
      
      \noalign{\vspace{3mm}} 
      
      \rotatebox[origin=c]{90}{CE + OFBD} & 
      \raisebox{-0.5\height}{\includegraphics[width=0.23\linewidth]{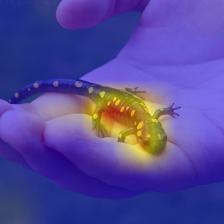}} & 
      \raisebox{-0.5\height}{\includegraphics[width=0.23\linewidth]{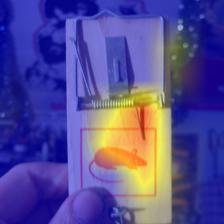}} & 
      \raisebox{-0.5\height}{\includegraphics[width=0.23\linewidth]{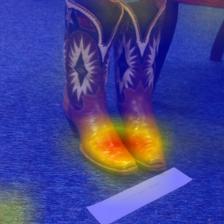}} &
      \raisebox{-0.5\height}{\includegraphics[width=0.23\linewidth]{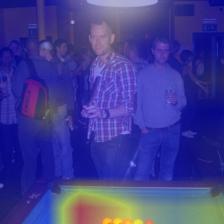}} \\
      
      \noalign{\vspace{2mm}} 
      
      & \textbf{Head Class} & \textbf{Head Class} & \textbf{Tail Class} & \textbf{Tail Class}\\
    \end{tabular}
    \vspace{2mm} 
    \caption{Attention Map}
    \label{fig:sub_attention}
\end{subfigure}
  \hfill 
  \begin{subfigure}[b]{0.38\textwidth} 
    \centering
    \includegraphics[width=0.95\linewidth]{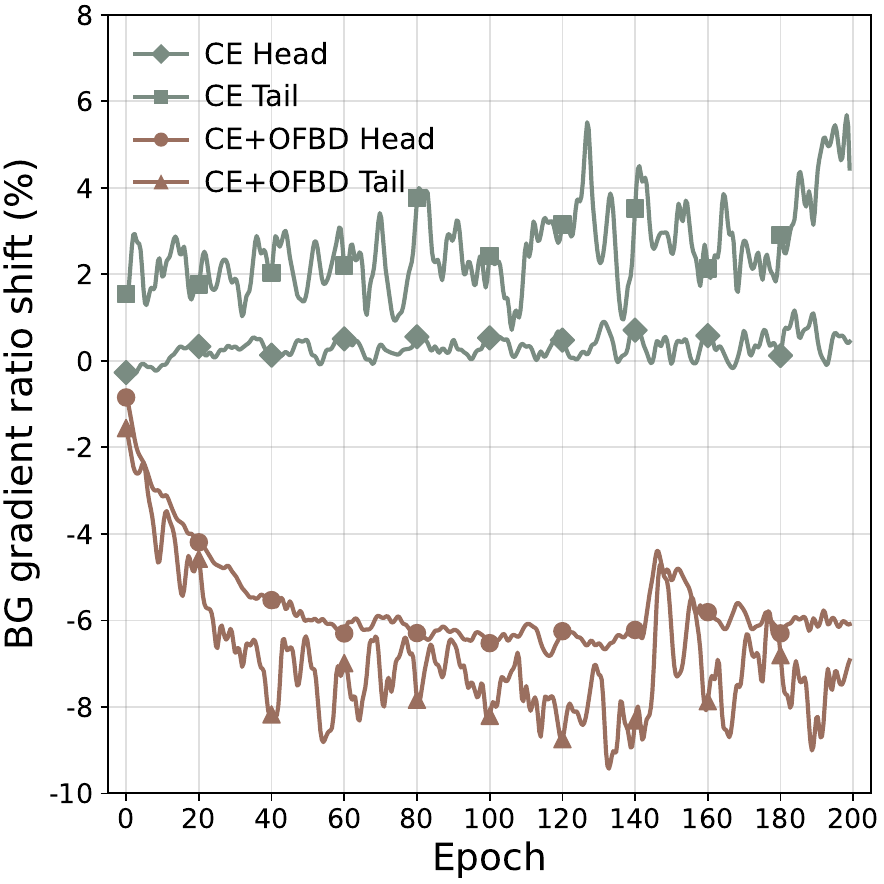}
    \caption{Background Gradient Radio Shift}
    \label{fig:sub_curve}
  \end{subfigure}
  
  \caption{Visualization of background debiasing on CIFAR100-LT ($r=100$). 
     (a) Pretrained-CAM attention maps of CE and CE+OFBD on three representative head, medium, and tail classes, showing that OFBD reduces background co-activation. More visualization results are provided in Appendix~\ref{sec:extended_cam}.
     (b) Background gradient ratio (Eq.~\ref{eq:bg_ratio}) shift during training. OFBD reduces optimization-level background bias (Sec.~\ref{sec: analysis_o}), especially for tail classes.}
  \label{fig:overall_comparison}
\end{figure*}


\noindent\textbf{Analysis of RL-based foreground region selection.}
Built upon the BCL~\cite{zhu2022balanced}, we further analyze the region selection strategy in FG-CutMix on CIFAR100-LT ($r=100$). As shown in Table~\ref{tab:rl_selection_ablation}, standard CutMix with random regions achieves only $52.17\%$, indicating that random replacement may preserve irrelevant backgrounds or discard target-related evidence. Using pretrained SAM- and CAM-based foreground regions selection improves the accuracy to $52.43\%$ and $52.63\%$, respectively, but the gains remain limited. In contrast, our RL-based foreground regions selection achieves the best accuracy of $54.19\%$, outperforming random, SAM-based, and CAM-based selection by $+2.02\%$, $+1.76\%$, and $+1.56\%$, respectively. These results show that accurate foreground localization is not necessarily the most effective mixing strategy, while RL-based selection better balances target preservation and background replacement. Further ablation analysis is provided in Appendix~\ref{sec:fgcutmix_gradient_analysis}.

\noindent\textbf{Visualization of Background Debiasing.}
We further visualize whether OFBD reduces background-biased representation and optimization. As shown in Fig.~\ref{fig:sub_attention}, CE predictions co-activate target foregrounds and surrounding backgrounds. With OFBD, attention becomes more object-focused, indicating reduced reliance on irrelevant background cues. Fig.~\ref{fig:sub_curve} shows that CE produces a positive and increasing background-gradient-ratio shift, particularly for tail classes, while CE+OFBD suppresses this shift for both head and tail classes, especially for tail classes. Overall, these results confirm that OFBD successfully mitigates background bias, encouraging the model to focus more on target semantics rather than spurious background cues.

\begin{table*}[t]
\centering
\scriptsize
\renewcommand{\arraystretch}{1.05}


\begin{minipage}[t]{0.49\textwidth}
\centering

\captionof{table}{Validation of $B_u$ on ImageNet-LT.}
\label{tab:bu_validation}

\setlength{\tabcolsep}{3pt}

\begin{tabular}{l@{\hspace{4pt}\vrule\hspace{2pt}\vrule\hspace{4pt}}c c c}
\toprule
Group
&$B_u$-FG$\downarrow$
&$B_u$-BG$\uparrow$
&BG AUROC$\uparrow$\\
\midrule

Overall&0.3657&0.6321&0.7182\\
Many&0.3639&0.6340&0.7247\\
Medium&0.3651&0.6315&0.7161\\
Few&0.3726&0.6292&0.7049\\

\bottomrule
\end{tabular}

\end{minipage}
\hfill
\begin{minipage}[t]{0.49\textwidth}
\centering

\captionof{table}{Background intervention analysis on ImageNet-LT.}
\label{tab:bg_intervention}

\setlength{\tabcolsep}{2.5pt}

\begin{tabular}{l@{\hspace{4pt}\vrule\hspace{2pt}\vrule\hspace{4pt}}c c c}
\toprule
Method 
& FG-only$\uparrow$
& BG-only$\downarrow$
& BG-swap$\uparrow$\\
\midrule

BCL~\cite{zhu2022balanced}
&40.18&20.15&31.21\\

\rowcolor{gray!18}
+OFBD
&$41.99_{\OFBDUp{1.81}}$
&$18.91_{\OFBDDown{1.24}}$
&$38.23_{\OFBDUp{7.02}}$\\

\bottomrule
\end{tabular}

\end{minipage}

\vspace{2mm}


\begin{minipage}[t]{0.49\textwidth}
\centering

\captionof{table}{Foreground selection comparison.}
\label{tab:rl_selection}

\setlength{\tabcolsep}{3pt}

\begin{tabular}{l@{\hspace{4pt}\vrule\hspace{2pt}\vrule\hspace{4pt}}c c c c}
\toprule

Method
&Overall$\uparrow$
&Many$\uparrow$
&Med.$\uparrow$
&Few$\uparrow$\\

\midrule

SaliencyMix~\cite{uddin2020saliencymix}
&52.85&67.62&53.47&34.90\\

SnapMix~\cite{huang2021snapmix}
&52.01&67.45&53.94&31.75\\

PuzzleMix~\cite{kim2020puzzle}
&52.52&67.06&53.80&34.06\\

ResizeMix~\cite{qin2020resizemix}
&52.03&67.24&53.91&32.96\\

Attentive CutMix~\cite{walawalkar2020attentive}
&53.19&\textbf{68.92}&53.99&33.91\\

\midrule

\rowcolor{gray!18}
RL selector
&\textbf{54.21}&66.51&\textbf{54.00}&\textbf{40.10}\\

\bottomrule
\end{tabular}

\end{minipage}
\hfill
\begin{minipage}[t]{0.49\textwidth}
\centering

\captionof{table}{Results with different imbalance ratios.}
\label{tab:imbalance_ratio}

\setlength{\tabcolsep}{3pt}

\begin{tabular}{c@{\hspace{4pt}\vrule\hspace{2pt}\vrule\hspace{4pt}}l c c c c}
\toprule
$r$ & Method & Many$\uparrow$ &  Few$\uparrow$ & Overall$\uparrow$\\
\midrule

200 & BCL~\cite{zhu2022balanced}
& 66.60
& 26.33
& 46.30 \\

\rowcolor{gray!18}
& +OFBD
& $65.37_{\OFBDDown{1.23}}$
& $30.08_{\OFBDUp{3.75}}$
& $47.20_{\OFBDUp{0.90}}$ \\

\midrule

300 & BCL~\cite{zhu2022balanced}
& 68.00
& 26.09
& 44.62 \\

\rowcolor{gray!18}
& +OFBD
& $65.10_{\OFBDDown{2.90}}$
& $29.59_{\OFBDUp{3.50}}$
& $45.78_{\OFBDUp{1.16}}$ \\

\midrule

400 & BCL~\cite{zhu2022balanced}
& 68.33
& 24.04
& 42.99 \\

\rowcolor{gray!18}
& +OFBD
& $64.60_{\OFBDDown{3.73}}$
& $27.70_{\OFBDUp{3.66}}$
& $44.16_{\OFBDUp{1.17}}$ \\

\bottomrule
\end{tabular}

\end{minipage}

\end{table*}

\subsection{Further Analysis of Background Debiasing}

\textbf{Background Score Validation.}
We evaluate whether $B_u$ captures background-related information using CAM-guided SAM masks. Specifically, CAM responses from a pretrained classifier are used as point prompts for SAM to obtain foreground/background masks. As shown in Table~\ref{tab:bu_validation}, $B_u$ assigns higher average scores to background regions than foreground regions and achieves BG AUROC above 0.70 across all groups, demonstrating its ability to identify background-biased features.

\textbf{Background Intervention.}
To verify whether OFBD reduces background shortcut reliance, we conduct intervention-based evaluations on ImageNet-LT~\cite{liu2019large} using foreground-only, background-only, and background-swapped inputs with fixed post-hoc foreground masks. As shown in Table~\ref{tab:bg_intervention}, OFBD improves FG-only accuracy by +1.81 points while reducing BG-only accuracy by 1.24 points, indicating a more object-focused representation. Moreover, it achieves a +7.02-point gain under BG-swap, demonstrating improved robustness against background bias under distribution shifts.

\textbf{Foreground Selection Analysis.}
We compare the proposed RL-based foreground selector with representative mixing strategies, including SaliencyMix~\cite{uddin2020saliencymix}, SnapMix~\cite{huang2021snapmix}, PuzzleMix~\cite{kim2020puzzle}, ResizeMix~\cite{qin2020resizemix}, and Attentive CutMix~\cite{walawalkar2020attentive}. As shown in Table~\ref{tab:rl_selection}, the RL selector achieves the best overall accuracy and Few-shot performance, improving Few-shot accuracy by +5.20 points over the strongest heuristic baseline, demonstrating the effectiveness of adaptive foreground preservation for long-tailed recognition. This improvement highlights the advantage of learning region selection over manually designed heuristic strategies.

\textbf{Performance under Different Imbalance Ratios.}
We further evaluate OFBD under more severe imbalance settings by increasing the imbalance ratio on CIFAR-100-LT. As shown in Table~\ref{tab:imbalance_ratio}, OFBD consistently improves overall and Few-shot accuracy across different imbalance ratios. Specifically, OFBD improves overall accuracy by +0.90/+1.16/+1.17 points under $r=200/300/400$, while achieving +3.75/+3.50/+3.66 points gains on Few-shot classes. These results demonstrate that OFBD remains effective under increasingly challenging long-tailed distributions.

\section{Conclusion}
Revisiting long-tailed recognition via background bias, we reveal that tail degradation arises from sample scarcity combined with background-biased representation and optimization (i.e., larger distribution shifts and gradient reliance). To mitigate this, we propose Object-Focused Background Debiasing (OFBD), a plug-in framework. Specifically, FG-CutMix preserves target foregrounds while diversifying backgrounds, and parameter-free BFR suppresses background-biased features. Experiments across four datasets demonstrate that OFBD improves mainstream methods, boosting tail accuracy without sacrificing head performance on both CNN and Transformer backbones. Its plug-in design inspires extensions to other long-tailed tasks like object detection and trajectory prediction.

\vspace{-1em}
\paragraph{Limitation:} the performance of OFBD on domains with highly complex scenes (\eg, medical imaging datasets) remains to be explored.

\bibliographystyle{plainnat}
\bibliography{refs}

\begin{thebibliography}{111}
\providecommand{\natexlab}[1]{#1}
\providecommand{\url}[1]{\texttt{#1}}
\expandafter\ifx\csname urlstyle\endcsname\relax
  \providecommand{\doi}[1]{doi: #1}\else
  \providecommand{\doi}{doi: \begingroup \urlstyle{rm}\Url}\fi

\bibitem[Alomar et~al.(2023)Alomar, Aysel, and Cai]{alomar2023data}
Khaled Alomar, Halil~Ibrahim Aysel, and Xiaohao Cai.
\newblock {Data Augmentation in Classification and Segmentation: A Survey and New Strategies}.
\newblock \emph{Journal of Imaging}, 9\penalty0 (2):\penalty0 46, 2023.

\bibitem[Arora et~al.(2019)Arora, Cohen, Hu, and Luo]{arora2019implicit}
Sanjeev Arora, Nadav Cohen, Wei Hu, and Yuping Luo.
\newblock {Implicit Regularization in Deep Matrix Factorization}.
\newblock \emph{Advances in Neural Information Processing Systems}, 32, 2019.

\bibitem[Ba et~al.(2016)Ba, Kiros, and Hinton]{ba2016layer}
Jimmy~Lei Ba, Jamie~Ryan Kiros, and Geoffrey~E Hinton.
\newblock {Layer Normalization}.
\newblock \emph{arXiv preprint arXiv:1607.06450}, 2016.

\bibitem[Cao et~al.(2019)Cao, Wei, Gaidon, Arechiga, and Ma]{cao2019learning}
Kaidi Cao, Colin Wei, Adrien Gaidon, Nikos Arechiga, and Tengyu Ma.
\newblock {Learning Imbalanced Datasets with Label-Distribution-Aware Margin Loss}.
\newblock \emph{Advances in Neural Information Processing Systems}, 32, 2019.

\bibitem[Chakraborty et~al.(2024)Chakraborty, Wang, Gao, Zheng, Zhang, and De~la Torre]{chakraborty2024visual}
Rwiddhi Chakraborty, Yinong Wang, Jialu Gao, Runkai Zheng, Cheng Zhang, and Fernando De~la Torre.
\newblock {Visual Data Diagnosis and Debiasing with Concept Graphs}.
\newblock \emph{Advances in Neural Information Processing Systems}, 37:\penalty0 106383--106410, 2024.

\bibitem[Chaudhari et~al.(2021)Chaudhari, Mithal, Polatkan, and Ramanath]{chaudhari2021attentive}
Sneha Chaudhari, Varun Mithal, Gungor Polatkan, and Rohan Ramanath.
\newblock {An Attentive Survey of Attention Models}.
\newblock \emph{ACM Transactions on Intelligent Systems and Technology}, 12\penalty0 (5):\penalty0 1--32, 2021.

\bibitem[Chaudhuri et~al.(2025)Chaudhuri, Dutta, Bui, and Georgescu]{chaudhuricloser}
Abhra Chaudhuri, Anjan Dutta, Tu~Bui, and Serban Georgescu.
\newblock {A Closer Look at Multimodal Representation Collapse}.
\newblock In \emph{International Conference on Machine Learning}, 2025.

\bibitem[Chen et~al.(2026)Chen, Liu, Wang, Wang, and Lu]{chen2026reframing}
Shenghan Chen, Yiming Liu, Yanzhen Wang, Yujia Wang, and Xiankai Lu.
\newblock {Reframing Long-Tailed Learning via Loss Landscape Geometry}.
\newblock \emph{arXiv preprint arXiv:2603.21217}, 2026.

\bibitem[Chen et~al.(2023)Chen, Chen, Du, Rashwan, Yang, Chen, Wang, and Li]{chen2023adamv}
Tianlong Chen, Xuxi Chen, Xianzhi Du, Abdullah Rashwan, Fan Yang, Huizhong Chen, Zhangyang Wang, and Yeqing Li.
\newblock {Adamv-Moe: Adaptive Multi-task Vision Mixture-of-Experts}.
\newblock In \emph{IEEE/CVF International Conference on Computer Vision}, pages 17346--17357, 2023.

\bibitem[Cubuk et~al.(2019)Cubuk, Zoph, Mane, Vasudevan, and Le]{cubuk2019autoaugment}
Ekin~D Cubuk, Barret Zoph, Dandelion Mane, Vijay Vasudevan, and Quoc~V Le.
\newblock {AutoAugment: Learning Augmentation Strategies from Data}.
\newblock In \emph{IEEE/CVF Conference on Computer Vision and Pattern Recognition}, pages 113--123, 2019.

\bibitem[Cui et~al.(2019)Cui, Jia, Lin, Song, and Belongie]{cui2019class}
Yin Cui, Menglin Jia, Tsung-Yi Lin, Yang Song, and Serge Belongie.
\newblock {Class-Balanced Loss Based on Effective Number of Samples}.
\newblock In \emph{IEEE/CVF Conference on Computer Vision and Pattern Recognition}, 2019.

\bibitem[Dai et~al.(2025)Dai, Liu, Liao, Huang, Cao, Wu, Zhao, Xu, Zeng, Liu, et~al.]{dai2025auggpt}
Haixing Dai, Zhengliang Liu, Wenxiong Liao, Xiaoke Huang, Yihan Cao, Zihao Wu, Lin Zhao, Shaochen Xu, Fang Zeng, Wei Liu, et~al.
\newblock {AugGPT: Leveraging Chatgpt for Text Data Augmentation}.
\newblock \emph{IEEE Transactions on Big Data}, 11\penalty0 (3):\penalty0 907--918, 2025.

\bibitem[Dong et~al.(2017)Dong, Gong, and Zhu]{dong2017class}
Qi~Dong, Shaogang Gong, and Xiatian Zhu.
\newblock {Class Rectification Hard Mining for Imbalanced Deep Learning}.
\newblock In \emph{IEEE/CVF International Conference on Computer Vision}, 2017.

\bibitem[Dosovitskiy et~al.(2020)Dosovitskiy, Beyer, Kolesnikov, Weissenborn, Zhai, Unterthiner, Dehghani, Minderer, Heigold, Gelly, et~al.]{dosovitskiyimage}
Alexey Dosovitskiy, Lucas Beyer, Alexander Kolesnikov, Dirk Weissenborn, Xiaohua Zhai, Thomas Unterthiner, Mostafa Dehghani, Matthias Minderer, Georg Heigold, Sylvain Gelly, et~al.
\newblock {An Image is Worth 16x16 Words: Transformers for Image Recognition at Scale}.
\newblock In \emph{International Conference on Learning Representations}, 2020.

\bibitem[Du et~al.(2021)Du, Hu, Kakade, Lee, and Lei]{dufew}
Simon~Shaolei Du, Wei Hu, Sham~M Kakade, Jason~D Lee, and Qi~Lei.
\newblock {Few-Shot Learning via Learning the Representation, Provably}.
\newblock In \emph{International Conference on Learning Representations}, 2021.

\bibitem[Esteva et~al.(2021)Esteva, Chou, Yeung, Naik, Madani, Mottaghi, Liu, Topol, Dean, and Socher]{esteva2021deep}
Andre Esteva, Katherine Chou, Serena Yeung, Nikhil Naik, Ali Madani, Ali Mottaghi, Yun Liu, Eric Topol, Jeff Dean, and Richard Socher.
\newblock {Deep Learning-enabled Medical Computer Vision}.
\newblock \emph{npj Digital Medicine}, 4\penalty0 (1):\penalty0 5, 2021.

\bibitem[Geirhos et~al.(2018)Geirhos, Rubisch, Michaelis, Bethge, Wichmann, and Brendel]{geirhos2018imagenet}
Robert Geirhos, Patricia Rubisch, Claudio Michaelis, Matthias Bethge, Felix~A Wichmann, and Wieland Brendel.
\newblock {ImageNet-trained CNNs are Biased towards Texture; Increasing Shape Bias Improves Accuracy and Robustness}.
\newblock In \emph{International Conference on Learning Representations}, 2018.

\bibitem[Geirhos et~al.(2020)Geirhos, Jacobsen, Michaelis, Zemel, Brendel, Bethge, and Wichmann]{geirhos2020shortcut}
Robert Geirhos, J{\"o}rn-Henrik Jacobsen, Claudio Michaelis, Richard Zemel, Wieland Brendel, Matthias Bethge, and Felix~A Wichmann.
\newblock {Shortcut Learning in Deep Neural Networks}.
\newblock \emph{Nature Machine Intelligence}, 2\penalty0 (11):\penalty0 665--673, 2020.

\bibitem[Hataya et~al.(2020)Hataya, Zdenek, Yoshizoe, and Nakayama]{hataya2020faster}
Ryuichiro Hataya, Jan Zdenek, Kazuki Yoshizoe, and Hideki Nakayama.
\newblock {Faster AutoAugment: Learning Augmentation Strategies Using Backpropagation}.
\newblock In \emph{European Conference on Computer Vision}, pages 1--16. Springer, 2020.

\bibitem[He et~al.(2016)He, Zhang, Ren, and Sun]{he2016deep}
Kaiming He, Xiangyu Zhang, Shaoqing Ren, and Jian Sun.
\newblock {Deep Residual Learning for Image Recognition}.
\newblock In \emph{IEEE/CVF Conference on Computer Vision and Pattern Recognition}, pages 770--778, 2016.

\bibitem[Hong et~al.(2021)Hong, Han, Choi, Seo, Kim, and Chang]{hong2021disentangling}
Youngkyu Hong, Seungju Han, Kwanghee Choi, Seokjun Seo, Beomsu Kim, and Buru Chang.
\newblock {Disentangling Label Distribution for Long-Tailed Visual Recognition}.
\newblock In \emph{IEEE/CVF Conference on Computer Vision and Pattern Recognition}, 2021.

\bibitem[Hou et~al.(2023)Hou, Zhang, Wang, and Zhou]{hou2023subclass}
Chengkai Hou, Jieyu Zhang, Haonan Wang, and Tianyi Zhou.
\newblock {Subclass-balancing Contrastive Learning for Long-tailed Recognition}.
\newblock In \emph{IEEE/CVF International Conference on Computer Vision}, pages 5395--5407, 2023.

\bibitem[Hou et~al.(2021)Hou, Zhou, and Feng]{hou2021coordinate}
Qibin Hou, Daquan Zhou, and Jiashi Feng.
\newblock {Coordinate Attention for Efficient Mobile Network Design}.
\newblock In \emph{IEEE/CVF Conference on Computer Vision and Pattern Recognition}, pages 13708--13717. IEEE, 2021.

\bibitem[Huang et~al.(2016)Huang, Li, Loy, and Tang]{huang2016learning}
Chen Huang, Yining Li, Chen~Change Loy, and Xiaoou Tang.
\newblock {Learning Deep Representation for Imbalanced Classification}.
\newblock In \emph{IEEE/CVF Conference on Computer Vision and Pattern Recognition}, 2016.

\bibitem[Huang et~al.(2021)Huang, Wang, and Tao]{huang2021snapmix}
Shaoli Huang, Xinchao Wang, and Dacheng Tao.
\newblock {Snapmix: Semantically Proportional Mixing for Augmenting Fine-grained Data}.
\newblock In \emph{AAAI Conference on Artificial Intelligence}, volume~35, pages 1628--1636, 2021.

\bibitem[Huy et~al.(2025)Huy, Huynh, Xie, Qi, Chen, Le~Nguyen, Tran, Phung, van~den Hengel, Liao, To, Verjans, and Phan]{Huy_2025_ICCV}
Ta~Duc Huy, Duy~Anh Huynh, Yutong Xie, Yuankai Qi, Qi~Chen, Phi Le~Nguyen, Sen~Kim Tran, Son~Lam Phung, Anton van~den Hengel, Zhibin Liao, Minh-Son To, Johan~W. Verjans, and Vu~Minh~Hieu Phan.
\newblock {Seeing the Trees for the Forest: Rethinking Weakly-Supervised Medical Visual Grounding}.
\newblock In \emph{IEEE/CVF International Conference on Computer Vision}, pages 24445--24455, 2025.

\bibitem[Ioffe and Szegedy(2015)]{ioffe2015batch}
Sergey Ioffe and Christian Szegedy.
\newblock {Batch Normalization: Accelerating Deep Network Training by Reducing Internal Covariate Shift}.
\newblock In \emph{International Conference on Machine Learning}, pages 448--456, 2015.

\bibitem[Jaiswal et~al.(2020)Jaiswal, Babu, Zadeh, Banerjee, and Makedon]{jaiswal2020survey}
Ashish Jaiswal, Ashwin~Ramesh Babu, Mohammad~Zaki Zadeh, Debapriya Banerjee, and Fillia Makedon.
\newblock {A Survey on Contrastive Self-Supervised Learning}.
\newblock \emph{Technologies}, 9\penalty0 (1):\penalty0 2, 2020.

\bibitem[Jian et~al.(2025)Jian, Chen, Wang, Yao, Wang, and Wu]{jian2025supervised}
Zhongquan Jian, Yanhao Chen, Yancheng Wang, Junfeng Yao, Meihong Wang, and Qingqiang Wu.
\newblock {Supervised Exploratory Learning for Long-tailed Visual Recognition}.
\newblock In \emph{IEEE/CVF International Conference on Computer Vision}, pages 1870--1880, 2025.

\bibitem[Jiang et~al.(2023)Jiang, Rahmani, Black, and Williams]{jiang2023probabilistic}
Zheheng Jiang, Hossein Rahmani, Sue Black, and Bryan~M Williams.
\newblock {A Probabilistic Attention Model with Occlusion-aware Texture Regression for 3D Hand Reconstruction from a Single RGB Image}.
\newblock In \emph{IEEE/CVF Conference on Computer Vision and Pattern Recognition}, pages 758--767, 2023.

\bibitem[Jung and Oh(2021)]{jung2021towards}
Hyungsik Jung and Youngrock Oh.
\newblock {Towards Better Explanations of Class Activation Mapping}.
\newblock In \emph{IEEE/CVF International Conference on Computer Vision}, pages 1336--1344, 2021.

\bibitem[Kang et~al.(2020)Kang, Li, Xie, Yuan, and Feng]{kang2020exploring}
Bingyi Kang, Yu~Li, Sa~Xie, Zehuan Yuan, and Jiashi Feng.
\newblock {Exploring Balanced Feature Spaces for Representation Learning}.
\newblock In \emph{International Conference on Learning Representations}, 2020.

\bibitem[Kim et~al.(2020)Kim, Choo, and Song]{kim2020puzzle}
Jang-Hyun Kim, Wonho Choo, and Hyun~Oh Song.
\newblock {Puzzle mix: Exploiting Saliency and Local Statistics for Optimal Mixup}.
\newblock In \emph{International Conference on Learning Representations}, pages 5275--5285. PMLR, 2020.

\bibitem[Kim et~al.(2024)Kim, Mo, Kim, Lee, Lee, and Shin]{kim2024discovering}
Younghyun Kim, Sangwoo Mo, Minkyu Kim, Kyungmin Lee, Jaeho Lee, and Jinwoo Shin.
\newblock {Discovering and Mitigating Visual Biases through Keyword Explanation}.
\newblock In \emph{IEEE/CVF Conference on Computer Vision and Pattern Recognition}, pages 11082--11092, 2024.

\bibitem[Kirillov et~al.(2023)Kirillov, Mintun, Ravi, Mao, Rolland, Gustafson, Xiao, Whitehead, Berg, Lo, et~al.]{kirillov2023segment}
Alexander Kirillov, Eric Mintun, Nikhila Ravi, Hanzi Mao, Chloe Rolland, Laura Gustafson, Tete Xiao, Spencer Whitehead, Alexander~C Berg, Wan-Yen Lo, et~al.
\newblock {Segment Anything}.
\newblock In \emph{IEEE/CVF International Conference on Computer Vision}, pages 4015--4026, 2023.

\bibitem[Konda and Tsitsiklis(1999)]{konda1999actor}
Vijay Konda and John Tsitsiklis.
\newblock {Actor-Critic Algorithms}.
\newblock \emph{Advances in Neural Information Processing Systems}, 12, 1999.

\bibitem[LeCun et~al.(2015)LeCun, Bengio, and Hinton]{lecun2015deep}
Yann LeCun, Yoshua Bengio, and Geoffrey Hinton.
\newblock {Deep Learning}.
\newblock \emph{Nature}, 521\penalty0 (7553):\penalty0 436--444, 2015.

\bibitem[Li et~al.(2023)Li, Liu, Zhang, and Li]{li2023mitigating}
Haoxin Li, Yuan Liu, Hanwang Zhang, and Boyang Li.
\newblock {Mitigating and Evaluating Static Bias of Action Representations in the Background and the Foreground}.
\newblock In \emph{IEEE/CVF International Conference on Computer Vision}, pages 19911--19923, 2023.

\bibitem[Li et~al.(2022)Li, Cheung, and Lu]{li2022long}
Mengke Li, Yiu-ming Cheung, and Yang Lu.
\newblock {Long-tailed Visual Recognition via Gaussian Clouded Logit Adjustment}.
\newblock In \emph{IEEE/CVF Conference on Computer Vision and Pattern Recognition}, pages 6929--6938, 2022.

\bibitem[Li et~al.(2024{\natexlab{a}})Li, Zhikai, Lu, Lan, Cheung, and Huang]{li2024feature}
Mengke Li, HU~Zhikai, Yang Lu, Weichao Lan, Yiu-ming Cheung, and Hui Huang.
\newblock {Feature Fusion from Head to Tail for Long-Tailed Visual Recognition}.
\newblock In \emph{AAAI Conference on Artificial Intelligence}, 2024{\natexlab{a}}.

\bibitem[Li et~al.(2021)Li, Gong, Liu, Wang, Qiao, and Cheng]{li2021metasaug}
Shuang Li, Kaixiong Gong, Chi~Harold Liu, Yulin Wang, Feng Qiao, and Xinjing Cheng.
\newblock {MetaSAug: Meta Semantic Augmentation for Long-Tailed Visual Recognition}.
\newblock In \emph{IEEE/CVF Conference on Computer Vision and Pattern Recognition}, 2021.

\bibitem[Li et~al.(2025)Li, Xu, Yang, Wang, Zhang, Cao, and Huang]{lifocal}
Sicong Li, Qianqian Xu, Zhiyong Yang, Zitai Wang, Linchao Zhang, Xiaochun Cao, and Qingming Huang.
\newblock {Focal-SAM: Focal Sharpness-Aware Minimization for Long-Tailed Classification}.
\newblock In \emph{International Conference on Machine Learning}, 2025.

\bibitem[Li et~al.(2024{\natexlab{b}})Li, Lyu, Shang, Wan, and Feng]{li2024long}
Weiqi Li, Fan Lyu, Fanhua Shang, Liang Wan, and Wei Feng.
\newblock {Long-Tailed Learning as Multi-Objective Optimization}.
\newblock In \emph{AAAI Conference on Artificial Intelligence}, 2024{\natexlab{b}}.

\bibitem[Li et~al.(2024{\natexlab{c}})Li, Ding, Wang, and Lee]{li2024empowering}
Yichuan Li, Kaize Ding, Jianling Wang, and Kyumin Lee.
\newblock {Empowering Large Language Models for Textual Data Augmentation}.
\newblock In \emph{Findings of the Association for Computational Linguistics: ACL 2024}, pages 12734--12751, 2024{\natexlab{c}}.

\bibitem[Li and Jia(2025)]{li2025conmix}
Zhixin Li and Yuheng Jia.
\newblock {ConMix: Contrastive Mixup at Representation Level for Long-tailed Deep Clustering}.
\newblock In \emph{International Conference on Learning Representations}, 2025.

\bibitem[Lim et~al.(2019)Lim, Kim, Kim, Kim, and Kim]{lim2019fast}
Sungbin Lim, Ildoo Kim, Taesup Kim, Chiheon Kim, and Sungwoong Kim.
\newblock {Fast AutoAugment}.
\newblock \emph{Advances in Neural Information Processing Systems}, 32, 2019.

\bibitem[Lin et~al.(2014)Lin, Maire, Belongie, Hays, Perona, Ramanan, Doll{\'a}r, and Zitnick]{lin2014microsoft}
Tsung-Yi Lin, Michael Maire, Serge Belongie, James Hays, Pietro Perona, Deva Ramanan, Piotr Doll{\'a}r, and C~Lawrence Zitnick.
\newblock {Microsoft COCO: Common Objects in Context}.
\newblock In \emph{European Conference on Computer Vision}, pages 740--755, 2014.

\bibitem[Liu et~al.(2021{\natexlab{a}})Liu, Li, Kang, Hua, and Vasconcelos]{liu2021gistnet}
Bo~Liu, Haoxiang Li, Hao Kang, Gang Hua, and Nuno Vasconcelos.
\newblock {GistNet: a Geometric Structure Transfer Network for Long-tailed Recognition}.
\newblock In \emph{IEEE/CVF International Conference on Computer Vision}, 2021{\natexlab{a}}.

\bibitem[Liu et~al.(2021{\natexlab{b}})Liu, Zhang, Hou, Mian, Wang, Zhang, and Tang]{liu2021self}
Xiao Liu, Fanjin Zhang, Zhenyu Hou, Li~Mian, Zhaoyu Wang, Jing Zhang, and Jie Tang.
\newblock {Self-Supervised Learning: Generative or Contrastive}.
\newblock \emph{IEEE Transactions on Knowledge and Data Engineering}, 35\penalty0 (1):\penalty0 857--876, 2021{\natexlab{b}}.

\bibitem[Liu et~al.(2008)Liu, Wu, and Zhou]{liu2008exploratory}
Xu-Ying Liu, Jianxin Wu, and Zhi-Hua Zhou.
\newblock {Exploratory Undersampling for Class-Imbalance Learning}.
\newblock \emph{IEEE Transactions on Systems, Man, and Cybernetics, Part B (Cybernetics)}, 39\penalty0 (2):\penalty0 539--550, 2008.

\bibitem[Liu et~al.(2025)Liu, Yang, and Wang]{liu2025long}
Yuting Liu, Liu Yang, and Yu~Wang.
\newblock {Long-Tailed Classification with Multi-Granularity Semantics}.
\newblock In \emph{IEEE/CVF International Conference on Computer Vision}, pages 4285--4294, 2025.

\bibitem[Liu et~al.(2019)Liu, Miao, Zhan, Wang, Gong, and Yu]{liu2019large}
Ziwei Liu, Zhongqi Miao, Xiaohang Zhan, Jiayun Wang, Boqing Gong, and Stella~X Yu.
\newblock {Large-Scale Long-Tailed Recognition in an Open World}.
\newblock In \emph{IEEE/CVF Conference on Computer Vision and Pattern Recognition}, pages 2537--2546, 2019.

\bibitem[Menon et~al.(2020)Menon, Jayasumana, Rawat, Jain, Veit, and Kumar]{menonlong}
Aditya~Krishna Menon, Sadeep Jayasumana, Ankit~Singh Rawat, Himanshu Jain, Andreas Veit, and Sanjiv Kumar.
\newblock {Long-Tail Learning via Logit Adjustment}.
\newblock In \emph{International Conference on Learning Representations}, 2020.

\bibitem[Minderer et~al.(2020)Minderer, Bachem, Houlsby, and Tschannen]{minderer2020automatic}
Matthias Minderer, Olivier Bachem, Neil Houlsby, and Michael Tschannen.
\newblock {Automatic Shortcut Removal for Self-supervised Representation Learning}.
\newblock In \emph{International Conference on Machine Learning}, pages 6927--6937. PMLR, 2020.

\bibitem[Mo et~al.(2021)Mo, Kang, Sohn, Li, and Shin]{mo2021object}
Sangwoo Mo, Hyunwoo Kang, Kihyuk Sohn, Chun-Liang Li, and Jinwoo Shin.
\newblock {Object-Aware Contrastive Learning for Debiased Scene Representation}.
\newblock \emph{Advances in Neural Information Processing Systems}, 34:\penalty0 12251--12264, 2021.

\bibitem[Moon et~al.(2021)Moon, Mo, Lee, Lee, and Shin]{moon2021masker}
Seung~Jun Moon, Sangwoo Mo, Kimin Lee, Jaeho Lee, and Jinwoo Shin.
\newblock {MASKER: Masked Keyword Regularization for Reliable Text Classification}.
\newblock In \emph{AAAI Conference on Artificial Intelligence}, volume~35, pages 13578--13586, 2021.

\bibitem[Narayan et~al.(2025)Narayan, Vs, and Patel]{narayan2025segface}
Kartik Narayan, Vibashan Vs, and Vishal~M Patel.
\newblock {SegFace: Face Segmentation of Long-tail Classes}.
\newblock In \emph{AAAI Conference on Artificial Intelligence}, 2025.

\bibitem[Pan et~al.(2024)Pan, Guo, Yu, and Chen]{pan2024enhanced}
Haolin Pan, Yong Guo, Mianjie Yu, and Jian Chen.
\newblock {Enhanced Long-Tailed Recognition with Contrastive Cutmix Augmentation}.
\newblock \emph{IEEE Transactions on Image Processing}, 33:\penalty0 4215--4230, 2024.

\bibitem[Park et~al.(2018)Park, Woo, Lee, and Kweon]{park2018bam}
Jongchan Park, Sanghyun Woo, Joon-Young Lee, and In~So Kweon.
\newblock {Bam: Bottleneck Attention Module}.
\newblock \emph{arXiv preprint arXiv:1807.06514}, 2018.

\bibitem[Peng et~al.(2019)Peng, Zhang, Xing, Gui, Huang, Jiang, Ding, and Chen]{peng2019trainable}
Minlong Peng, Qi~Zhang, Xiaoyu Xing, Tao Gui, Xuanjing Huang, Yu-Gang Jiang, Keyu Ding, and Zhigang Chen.
\newblock {Trainable Undersampling for Class-Imbalance Learning}.
\newblock In \emph{AAAI Conference on Artificial Intelligence}, volume~33, pages 4707--4714, 2019.

\bibitem[Qiao et~al.(2019)Qiao, Zheng, Cao, and Lau]{qiao2019tell}
Xiaotian Qiao, Quanlong Zheng, Ying Cao, and Rynson~WH Lau.
\newblock {Tell Me Where I Am: Object-level Scene Context Prediction}.
\newblock In \emph{IEEE/CVF Conference on Computer Vision and Pattern Recognition}, pages 2633--2641, 2019.

\bibitem[Qin et~al.(2020)Qin, Fang, Zhang, Liu, Wang, and Wang]{qin2020resizemix}
Jie Qin, Jiemin Fang, Qian Zhang, Wenyu Liu, Xingang Wang, and Xinggang Wang.
\newblock {Resizemix: Mixing Data with Preserved Object Information and True Labels}.
\newblock \emph{arXiv preprint arXiv:2012.11101}, 2020.

\bibitem[Rangwani et~al.(2024)Rangwani, Mondal, Mishra, Asokan, and Babu]{rangwani2024deit}
Harsh Rangwani, Pradipto Mondal, Mayank Mishra, Ashish~Ramayee Asokan, and R~Venkatesh Babu.
\newblock {DeiT-LT: Distillation Strikes Back for Vision Transformer Training on Long-tailed Datasets}.
\newblock In \emph{IEEE/CVF Conference on Computer Vision and Pattern Recognition}, pages 23396--23406, 2024.

\bibitem[Ravi et~al.(2025)Ravi, Gabeur, Hu, Hu, Ryali, Ma, Khedr, R{\"a}dle, Rolland, Gustafson, et~al.]{ravisam}
Nikhila Ravi, Valentin Gabeur, Yuan-Ting Hu, Ronghang Hu, Chaitanya Ryali, Tengyu Ma, Haitham Khedr, Roman R{\"a}dle, Chloe Rolland, Laura Gustafson, et~al.
\newblock {SAM 2: Segment Anything in Images and Videos}.
\newblock In \emph{International Conference on Learning Representations}, 2025.

\bibitem[Ren et~al.(2020)Ren, Yu, Ma, Zhao, Yi, et~al.]{ren2020balanced}
Jiawei Ren, Cunjun Yu, Xiao Ma, Haiyu Zhao, Shuai Yi, et~al.
\newblock {Balanced Meta-Softmax for Long-tailed Visual Recognition}.
\newblock In \emph{Advances in Neural Information Processing Systems}, 2020.

\bibitem[Russakovsky et~al.(2015)Russakovsky, Deng, Su, Krause, Satheesh, Ma, Huang, Karpathy, Khosla, Bernstein, et~al.]{russakovsky2015imagenet}
Olga Russakovsky, Jia Deng, Hao Su, Jonathan Krause, Sanjeev Satheesh, Sean Ma, Zhiheng Huang, Andrej Karpathy, Aditya Khosla, Michael Bernstein, et~al.
\newblock {ImageNet Large Scale Visual Recognition Challenge}.
\newblock \emph{International Journal of Computer Vision}, 115\penalty0 (3):\penalty0 211--252, 2015.

\bibitem[Sagawa et~al.(2019)Sagawa, Koh, Hashimoto, and Liang]{sagawadistributionally}
Shiori Sagawa, Pang~Wei Koh, Tatsunori~B Hashimoto, and Percy Liang.
\newblock {Distributionally Robust Neural Networks}.
\newblock In \emph{International Conference on Learning Representations}, 2019.

\bibitem[Samuel and Chechik(2021)]{samuel2021distributional}
Dvir Samuel and Gal Chechik.
\newblock {Distributional Robustness Loss for Long-tail Learning}.
\newblock In \emph{IEEE/CVF International Conference on Computer Vision}, 2021.

\bibitem[Schulman et~al.(2017)Schulman, Wolski, Dhariwal, Radford, and Klimov]{schulman2017proximal}
John Schulman, Filip Wolski, Prafulla Dhariwal, Alec Radford, and Oleg Klimov.
\newblock {Proximal Policy Optimization Algorithms}.
\newblock \emph{arXiv preprint arXiv:1707.06347}, 2017.

\bibitem[Shao et~al.(2024{\natexlab{a}})Shao, Zhu, Zhang, and Wu]{shao2024diffult}
Jie Shao, Ke~Zhu, Hanxiao Zhang, and Jianxin Wu.
\newblock {DiffuLT: Diffusion for Long-tail Recognition Without External Knowledge}.
\newblock In \emph{Advances in Neural Information Processing Systems}, 2024{\natexlab{a}}.

\bibitem[Shao et~al.(2024{\natexlab{b}})Shao, Wang, Zhu, Xu, Song, Bi, Zhang, Zhang, Li, et~al.]{shao2024deepseekmath}
Zhihong Shao, Peiyi Wang, Qihao Zhu, Runxin Xu, Junxiao Song, Xiao Bi, Haowei Zhang, Mingchuan Zhang, YK~Li, et~al.
\newblock {DeepSeekMath: Pushing the Limits of Mathematical Reasoning in Open Language Models}.
\newblock \emph{arXiv preprint arXiv:2402.03300}, 2024{\natexlab{b}}.

\bibitem[Shetty et~al.(2019)Shetty, Schiele, and Fritz]{shetty2019not}
Rakshith Shetty, Bernt Schiele, and Mario Fritz.
\newblock {Not Using the Car to See the Sidewalk--Quantifying and Controlling the Effects of Context in Classification and Segmentation}.
\newblock In \emph{IEEE/CVF Conference on Computer Vision and Pattern Recognition}, pages 8218--8226, 2019.

\bibitem[Shi et~al.(2026)Shi, Yu, and Yang]{shi2026vision}
Cheng Shi, Yizhou Yu, and Sibei Yang.
\newblock {Vision Transformers Need More Than Registers}.
\newblock \emph{arXiv preprint arXiv:2602.22394}, 2026.

\bibitem[Shi et~al.(2023{\natexlab{a}})Shi, Wei, Xiang, and Li]{shi2023re}
Jiang-Xin Shi, Tong Wei, Yuke Xiang, and Yu-Feng Li.
\newblock {How Re-sampling Helps for Long-tail Learning?}
\newblock \emph{Advances in Neural Information Processing Systems}, 36:\penalty0 75669--75687, 2023{\natexlab{a}}.

\bibitem[Shi et~al.(2023{\natexlab{b}})Shi, Wei, Zhou, Han, Shao, and Li]{shi2023parameter}
Jiang-Xin Shi, Tong Wei, Zhi Zhou, Xin-Yan Han, Jie-Jing Shao, and Yufeng Li.
\newblock {Parameter-Efficient Long-Tailed Recognition}.
\newblock \emph{CoRR}, 2023{\natexlab{b}}.

\bibitem[Song et~al.(2025)Song, Qu, Zhou, and Cheng]{Song_2025_CVPR}
Mingyang Song, Xiaoye Qu, Jiawei Zhou, and Yu~Cheng.
\newblock {From Head to Tail: Towards Balanced Representation in Large Vision-Language Models through Adaptive Data Calibration}.
\newblock In \emph{IEEE/CVF Conference on Computer Vision and Pattern Recognition}, 2025.

\bibitem[Tang et~al.(2020)Tang, Huang, and Zhang]{tang2020long}
Kaihua Tang, Jianqiang Huang, and Hanwang Zhang.
\newblock {Long-Tailed Classification by Keeping the Good and Removing the Bad Momentum Causal Effect}.
\newblock \emph{Advances in Neural Information Processing Systems}, 33:\penalty0 1513--1524, 2020.

\bibitem[Uddin et~al.(2020)Uddin, Monira, Shin, Chung, Bae, et~al.]{uddin2020saliencymix}
AFM Uddin, Mst Monira, Wheemyung Shin, TaeChoong Chung, Sung-Ho Bae, et~al.
\newblock {Saliencymix: A Saliency Guided Data Augmentation Strategy for Better Regularization}.
\newblock \emph{arXiv preprint arXiv:2006.01791}, 2020.

\bibitem[Van~Horn et~al.(2018)Van~Horn, Mac~Aodha, Song, Cui, Sun, Shepard, Adam, Perona, and Belongie]{van2018inaturalist}
Grant Van~Horn, Oisin Mac~Aodha, Yang Song, Yin Cui, Chen Sun, Alex Shepard, Hartwig Adam, Pietro Perona, and Serge Belongie.
\newblock {The Inaturalist Species Classification and Detection Dataset}.
\newblock In \emph{IEEE/CVF Conference on Computer Vision and Pattern Recognition}, pages 8769--8778, 2018.

\bibitem[Voulodimos et~al.(2018)Voulodimos, Doulamis, Doulamis, and Protopapadakis]{voulodimos2018deep}
Athanasios Voulodimos, Nikolaos Doulamis, Anastasios Doulamis, and Eftychios Protopapadakis.
\newblock {Deep Learning for Computer Vision: A Brief Review}.
\newblock \emph{Computational Intelligence and Neuroscience}, 2018\penalty0 (1):\penalty0 7068349, 2018.

\bibitem[Walawalkar et~al.(2020)Walawalkar, Shen, Liu, and Savvides]{walawalkar2020attentive}
Devesh Walawalkar, Zhiqiang Shen, Zechun Liu, and Marios Savvides.
\newblock {Attentive cutmix: An Enhanced Data Augmentation Approach for Deep Learning based Image Classification}.
\newblock \emph{arXiv preprint arXiv:2003.13048}, 2020.

\bibitem[Wang et~al.(2019)Wang, Ge, Lipton, and Xing]{wang2019learning}
Haohan Wang, Songwei Ge, Zachary Lipton, and Eric~P Xing.
\newblock {Learning Robust Global Representations by Penalizing Local Predictive Power}.
\newblock \emph{Advances in Neural Information Processing Systems}, 32, 2019.

\bibitem[Wang et~al.(2021)Wang, Han, Wei, Zhang, and Wang]{wang2021contrastive}
Peng Wang, Kai Han, Xiu-Shen Wei, Lei Zhang, and Lei Wang.
\newblock {Contrastive Learning Based Hybrid Networks for Long-Tailed Image Classification}.
\newblock In \emph{IEEE/CVF Conference on Computer Vision and Pattern Recognition}, pages 943--952, 2021.

\bibitem[Wang et~al.(2024)Wang, Zhao, Wen, Wang, Wang, Zhang, and Wang]{LLMAutoDA}
Pengkun Wang, Zhe Zhao, Haibin Wen, Fanfu Wang, Binwu Wang, Qingfu Zhang, and Yang Wang.
\newblock Llm-autoda: Large language model-driven automatic data augmentation for long-tailed problems.
\newblock In \emph{Advances in Neural Information Processing Systems}, 2024.

\bibitem[Wang et~al.(2025{\natexlab{a}})Wang, Shubham, De~La~Parra, and Kumar]{wang2025mixa}
Weitian Wang, Rai Shubham, Cecilia De~La~Parra, and Akash Kumar.
\newblock {MixA-Q: Revisiting Activation Sparsity for Vision Transformers from a Mixed-precision Quantization Perspective}.
\newblock In \emph{IEEE/CVF International Conference on Computer Vision}, pages 22143--22152, 2025{\natexlab{a}}.

\bibitem[Wang and Zhu(2023)]{wang2023context}
Xuan Wang and Zhigang Zhu.
\newblock {Context Understanding in Computer Vision: A Survey}.
\newblock \emph{{Computer Vision and Image Understanding}}, 229:\penalty0 103646, 2023.

\bibitem[Wang et~al.(2020{\natexlab{a}})Wang, Lian, Miao, Liu, and Yu]{wanglong}
Xudong Wang, Long Lian, Zhongqi Miao, Ziwei Liu, and Stella Yu.
\newblock {Long-Tailed Recognition by Routing Diverse Distribution-Aware Experts}.
\newblock In \emph{International Conference on Learning Representations}, 2020{\natexlab{a}}.

\bibitem[Wang et~al.(2020{\natexlab{b}})Wang, Yao, Kwok, and Ni]{wang2020generalizing}
Yaqing Wang, Quanming Yao, James~T Kwok, and Lionel~M Ni.
\newblock {Generalizing from a Few Examples: A Survey on Few-shot Learning}.
\newblock \emph{ACM Computing Surveys (csur)}, 53\penalty0 (3):\penalty0 1--34, 2020{\natexlab{b}}.

\bibitem[Wang et~al.(2020{\natexlab{c}})Wang, Lv, Huang, Song, Yang, and Huang]{wang2020glance}
Yulin Wang, Kangchen Lv, Rui Huang, Shiji Song, Le~Yang, and Gao Huang.
\newblock {Glance and Focus: a Dynamic Approach to Reducing Spatial Redundancy in Image Classification}.
\newblock \emph{Advances in Neural Information Processing Systems}, 33:\penalty0 2432--2444, 2020{\natexlab{c}}.

\bibitem[Wang et~al.(2025{\natexlab{b}})Wang, Zhang, Zhang, Liu, Wang, and Zhou]{wang2025diversity}
Zaitian Wang, Jinghan Zhang, Xinhao Zhang, Kunpeng Liu, Pengfei Wang, and Yuanchun Zhou.
\newblock {Diversity-oriented Data Augmentation with Large Language Models}.
\newblock In \emph{Proceedings of the 63rd Annual Meeting of the Association for Computational Linguistics (Volume 1: Long Papers)}, pages 22265--22283, 2025{\natexlab{b}}.

\bibitem[Wei et~al.(2021)Wei, Sohn, Mellina, Yuille, and Yang]{wei2021crest}
Chen Wei, Kihyuk Sohn, Clayton Mellina, Alan Yuille, and Fan Yang.
\newblock {CReST: A Class-rebalancing Self-training Framework for Imbalanced Semi-supervised Learning}.
\newblock In \emph{IEEE/CVF Conference on Computer Vision and Pattern Recognition}, 2021.

\bibitem[Woo et~al.(2018)Woo, Park, Lee, and Kweon]{woo2018cbam}
Sanghyun Woo, Jongchan Park, Joon-Young Lee, and In~So Kweon.
\newblock {CBAM: Convolutional Block Attention Module}.
\newblock In \emph{European Conference on Computer Vision}, pages 3--19, 2018.

\bibitem[Wu et~al.(2020)Wu, Morgado, Wang, Ho, and Vasconcelos]{wu2020solving}
Tz-Ying Wu, Pedro Morgado, Pei Wang, Chih-Hui Ho, and Nuno Vasconcelos.
\newblock {Solving Long-tailed Recognition with Deep Realistic Taxonomic Classifier}.
\newblock In \emph{European Conference on Computer Vision}, 2020.

\bibitem[Wu et~al.(2021)Wu, Yu, Ye, Zhang, Zhuo, et~al.]{wu2021coordinated}
Zifan Wu, Chao Yu, Deheng Ye, Junge Zhang, Hankz~Hankui Zhuo, et~al.
\newblock {Coordinated Proximal Policy Optimization}.
\newblock \emph{Advances in Neural Information Processing Systems}, 34:\penalty0 26437--26448, 2021.

\bibitem[Xiao et~al.(2020)Xiao, Engstrom, Ilyas, and Madry]{xiaonoise}
Kai~Yuanqing Xiao, Logan Engstrom, Andrew Ilyas, and Aleksander Madry.
\newblock {Noise or Signal: The Role of Image Backgrounds in Object Recognition}.
\newblock In \emph{International Conference on Learning Representations}, 2020.

\bibitem[Yang et~al.(2021)Yang, Zhang, Li, and Xie]{yang2021simam}
Lingxiao Yang, Ru-Yuan Zhang, Lida Li, and Xiaohua Xie.
\newblock {SimAm: A Simple, Parameter-Free Attention Module for Convolutional Neural Networks}.
\newblock In \emph{International Conference on Machine Learning}, pages 11863--11874. PMLR, 2021.

\bibitem[Yang et~al.(2025)Yang, Wang, Fu, Tian, Kamishima, Ikebe, Ou, and Okutomi]{RAMW600}
Songxiao Yang, Haolin Wang, Yao Fu, Ye~Tian, Tamostu Kamishima, Masayuki Ikebe, Yafei Ou, and Masatoshi Okutomi.
\newblock {RAM-W600: A Multi-Task Wrist Dataset and Benchmark for Rheumatoid Arthritis}.
\newblock In \emph{Advances in Neural Information Processing Systems}, volume~38. Curran Associates, Inc., 2025.

\bibitem[Yun et~al.(2019)Yun, Han, Oh, Chun, Choe, and Yoo]{Yun_2019_ICCV}
Sangdoo Yun, Dongyoon Han, Seong~Joon Oh, Sanghyuk Chun, Junsuk Choe, and Youngjoon Yoo.
\newblock {CutMix: Regularization Strategy to Train Strong Classifiers with Localizable Features}.
\newblock In \emph{IEEE/CVF International Conference on Computer Vision}, October 2019.

\bibitem[Zhang et~al.(2020)Zhang, Karimireddy, Veit, Kim, Reddi, Kumar, and Sra]{zhang2020adaptive}
Jingzhao Zhang, Sai~Praneeth Karimireddy, Andreas Veit, Seungyeon Kim, Sashank Reddi, Sanjiv Kumar, and Suvrit Sra.
\newblock {Why are Adaptive Methods Good for Attention Models?}
\newblock \emph{Advances in Neural Information Processing Systems}, 33:\penalty0 15383--15393, 2020.

\bibitem[Zhang et~al.(2022)Zhang, Hooi, Hong, and Feng]{zhang2022self}
Yifan Zhang, Bryan Hooi, Lanqing Hong, and Jiashi Feng.
\newblock {Self-supervised Aggregation of Diverse Experts for Test-agnostic Long-tailed Recognition}.
\newblock \emph{Advances in Neural Information Processing Systems}, 35:\penalty0 34077--34090, 2022.

\bibitem[Zhang et~al.(2023)Zhang, Kang, Hooi, Yan, and Feng]{zhang2023deep}
Yifan Zhang, Bingyi Kang, Bryan Hooi, Shuicheng Yan, and Jiashi Feng.
\newblock {Deep Long-Tailed Learning: A Survey}.
\newblock \emph{IEEE Transactions on Pattern Analysis and Machine Intelligence}, 45\penalty0 (9):\penalty0 10795--10816, 2023.

\bibitem[Zhang and Pfister(2021)]{zhang2021learning}
Zizhao Zhang and Tomas Pfister.
\newblock {Learning Fast Sample Re-weighting without Reward Data}.
\newblock In \emph{IEEE/CVF International Conference on Computer Vision}, 2021.

\bibitem[Zhao et~al.(2023)Zhao, Jiang, Hu, Zhang, and Liu]{zhao2023mdcs}
Qihao Zhao, Chen Jiang, Wei Hu, Fan Zhang, and Jun Liu.
\newblock {MDCS: More Diverse Experts with Consistency Self-Distillation for Long-tailed Recognition}.
\newblock In \emph{IEEE/CVF International Conference on Computer Vision}, pages 11563--11574. IEEE, 2023.

\bibitem[Zhao et~al.(2024)Zhao, Dai, Li, Hu, Zhang, and Liu]{zhao2024ltgc}
Qihao Zhao, Yalun Dai, Hao Li, Wei Hu, Fan Zhang, and Jun Liu.
\newblock {LTGC: Long-tail Recognition via Leveraging LLMs-driven Generated Content}.
\newblock In \emph{IEEE/CVF Conference on Computer Vision and Pattern Recognition}, 2024.

\bibitem[Zhao et~al.(2025)Zhao, Wen, Liu, Ma, Yuan, and Qi]{zhao2025learning}
Shizhen Zhao, Xin Wen, Jiahui Liu, Chuofan Ma, Chunfeng Yuan, and Xiaojuan Qi.
\newblock {Learning from Neighbors: Category Extrapolation for Long-tail Learning}.
\newblock In \emph{IEEE/CVF Conference on Computer Vision and Pattern Recognition}, pages 30483--30492, 2025.

\bibitem[Zhou et~al.(2016)Zhou, Khosla, Lapedriza, Oliva, and Torralba]{zhou2016learning}
Bolei Zhou, Aditya Khosla, Agata Lapedriza, Aude Oliva, and Antonio Torralba.
\newblock {Learning Deep Features for Discriminative Localization}.
\newblock In \emph{IEEE/CVF Conference on Computer Vision and Pattern Recognition}, pages 2921--2929, 2016.

\bibitem[Zhou et~al.(2020)Zhou, Cui, Wei, and Chen]{zhou2020bbn}
Boyan Zhou, Quan Cui, Xiu-Shen Wei, and Zhao-Min Chen.
\newblock {BBN: Bilateral-branch Network with Cumulative Learning for Long-tailed Visual Recognition}.
\newblock In \emph{IEEE/CVF Conference on Computer Vision and Pattern Recognition}, 2020.

\bibitem[Zhou et~al.(2021)Zhou, Li, Peng, Zhang, and Zhang]{zhou2021triplet}
Haoyi Zhou, Jianxin Li, Jieqi Peng, Shuai Zhang, and Shanghang Zhang.
\newblock {Triplet Attention: Rethinking the Similarity in Transformers}.
\newblock In \emph{ACM SIGKDD Conference on Knowledge Discovery and Data Mining}, pages 2378--2388, 2021.

\bibitem[Zhou et~al.(2023)Zhou, Li, Zhao, Heng, and Gong]{zhou2023class}
Zhipeng Zhou, Lanqing Li, Peilin Zhao, Pheng-Ann Heng, and Wei Gong.
\newblock {Class-conditional Sharpness-aware Minimization for Deep Long-tailed Recognition}.
\newblock In \emph{IEEE/CVF Conference on Computer Vision and Pattern Recognition}, pages 3499--3509, 2023.

\bibitem[Zhu et~al.(2022)Zhu, Wang, Chen, Chen, and Jiang]{zhu2022balanced}
Jianggang Zhu, Zheng Wang, Jingjing Chen, Yi-Ping~Phoebe Chen, and Yu-Gang Jiang.
\newblock {Balanced Contrastive Learning for Long-Tailed Visual Recognition}.
\newblock In \emph{IEEE/CVF Conference on Computer Vision and Pattern Recognition}, 2022.

\bibitem[Zhu et~al.(2024)Zhu, Fu, Shao, Liu, and Wu]{zhu2024rectify}
Ke~Zhu, Minghao Fu, Jie Shao, Tianyu Liu, and Jianxin Wu.
\newblock {Rectify the Regression Bias in Long-Tailed Object Detection}.
\newblock In \emph{European Conference on Computer Vision}, 2024.

\end{thebibliography}


\appendix
\newpage

\begin{center}
    \Large \bf OFBD: Object-Focused Background Debiasing for Long-tailed Learning
\end{center}

\vspace{1em}
\noindent \textbf{Overview:} This supplementary material provides additional mathematical derivations (Appendix~\ref{sec:derivations}), extensive visualization results (Appendix~\ref{sec:visualizations}), additional experimental results and ablations (Appendix~\ref{sec:add_experiments}), and comprehensive implementation details (Appendix~\ref{sec:implementation}) to support the main paper.

\section{More Experiment Protocols}
\label{sec:implementation}

\subsection{Implementation Details}
\label{supp:Details}

OFBD is implemented as a plug-and-play module and can be integrated into various long-tailed recognition frameworks without altering their original backbones or classifier designs.

For CIFAR-10-LT and CIFAR-100-LT, we use ResNet-32 as the backbone. Models are trained for 200 epochs with a batch size of 256 using SGD with momentum 0.9 and weight decay $5\times10^{-4}$. The learning rate is linearly warmed up to 0.15 during the first 20 epochs and decayed by 0.1 at epochs 160 and 180. When OFBD is plugged into contrastive-learning baselines, \eg, BCL~\cite{zhu2022balanced} and SBCL~\cite{hou2023subclass}, we keep their original contrastive objectives unchanged. The projection dimension is 128, the contrastive temperature is 0.1, and the classification and contrastive losses are weighted by 2 and 0.6, following the original baseline settings~\cite{hou2023subclass, zhu2022balanced}. For non-contrastive baselines, the projection head and contrastive loss are removed.

For FG-CutMix, we apply the augmentation with probability 0.5. Given an input image, we randomly generate a region proposal set $\mathcal{P}(x)=\{b_1,\ldots,b_N\}$ with different locations and scales. Instead of ranking proposals by local energy, the RL-based region selector $\pi_\theta$ predicts a selection probability over $\mathcal{P}(x)$, and $K$ proposals are sampled without replacement to form the foreground candidate pool $\mathcal{R}_\theta(x)=\{r_1,\ldots,r_K\}$. For each selected candidate, we construct a mixed sample by preserving the selected target-related region and replacing its complementary region with content from another image. The mixed label is corrected in line with prior CutMix-based augmentation methods.

For selector optimization, the classifier prediction on each mixed sample provides the reward signal. Specifically, the reward is set to 1 if the mixed sample has a higher predicted probability for the target foreground-source class than for the background-source class, and 0 otherwise. Rewards are normalized within each candidate pool to compute relative advantages. The selector is optimized by policy gradient with a clipped objective and KL regularization to a frozen reference selector $\pi_{\mathrm{ref}}$, which is initialized from the previous epoch selector to stabilize policy updates. Unless otherwise specified, we set the RL loss weight $\gamma=0.8$, the auxiliary selector supervision weight $\alpha=1.0$, the KL coefficient $\beta=0.02$, and the clipping range $\epsilon_c=0.2$.

For BFR, we use the final convolutional feature map as input. Neuron-wise importance scores are computed from channel-wise feature statistics. Effective channels are selected by comparing each neuron-wise score with the spatial mean score, and the resulting foreground contribution is used to derive background-aware scores. The background-biased component is then estimated by background-aware weighted aggregation, and the rectified representation is computed as
$
f' = f - \lambda f_{\mathrm{bg}}^{\mathrm{BFR}},
$
where $\lambda=0.5$ in all experiments. The rectified representation is fed into the classifier and, when applicable, the projection head. BFR introduces no learnable parameters and requires no external annotations, pretrained localization models, or additional supervision.

For ImageNet-LT and iNaturalist 2018, we use ResNet-50~\cite{he2016deep} and ViT-B/16~\cite{dosovitskiyimage} as backbones. Models are trained with SGD, momentum 0.9, and batch size 256. For ImageNet-LT, we train for 90 epochs with a learning rate of 0.1 and weight decay $5\times10^{-4}$. For iNaturalist 2018, we train for 100 epochs with a learning rate of 0.2 and a weight decay of $1\times10^{-4}$. Unless otherwise specified, the same OFBD plug-in hyperparameters are used across datasets.

\subsection{Dataset and Evaluation Protocol}
\label{supp:protocol}

We evaluate OFBD on CIFAR-10-LT, CIFAR-100-LT, ImageNet-LT, and iNaturalist 2018. For CIFAR-10-LT and CIFAR-100-LT, we use imbalance ratios $r \in \{100, 50, 10\}$. The original balanced test sets are used for evaluation. For ImageNet-LT and iNaturalist 2018, we follow the official train/validation splits~\cite{zhang2023deep}. We report Top-1 accuracy for all datasets. For class-frequency analysis, we report Many-, Medium-, and Few-shot accuracy, where classes with more than 100 samples are Many-shot, classes with 20--100 samples are Medium-shot, and classes with fewer than 20 samples are Few-shot. For plug-in comparisons, OFBD keeps the official baseline's backbone, objective, classifier, sampler, and training schedule.


\section{Theoretical Derivations and Analysis}
\label{sec:derivations}


\subsection{Interpretation of Distribution-level Background Bias (Eq.~\ref{eq:gradient})}
\label{sec:distribution_shift_proof}

In Sec.~\ref{sec: analysis}, we hypothesized that tail classes suffer from a larger background distribution shift in Eq.~\ref{eq:gradient} and further corroborate this hypothesis through empirical observations. Here, we provide an interpretation to mathematically substantiate this claim based on sample complexity, deeply aligned with findings on spurious causal correlations in imbalanced data~\cite{tang2020long}. Let $N_y$ denote the number of training samples for class $y$. In a long-tailed recognition, we have $N_{head} \gg N_{tail}$. 

\textbf{1. Deviation of Empirical Distributions.} \\
Let $q_y$ denote the ideal background distribution for class $y$ constructed from a fully balanced, infinite-sample dataset. Under the long-tailed setting, the model observes an empirical background distribution, denoted as $p_y$, which is estimated from only $N_y$ samples. 

According to statistical learning theory~\cite{cao2019learning,zhou2023class,lifocal}, the deviation between an empirical distribution and the true underlying distribution is bounded by the sample size. Specifically, the expected $L_1$ distance between $p_y$ and $q_y$ converges at a rate of $\mathcal{O}\left(1/\sqrt{N_y}\right)$:
\begin{equation}
    \mathbb{E} \left[ \|p_y - q_y\|_1 \right] \le \mathcal{O} \left( \frac{1}{\sqrt{N_y}} \right).
\end{equation}

Since $N_{head} \gg N_{tail}$, the expected distribution deviation for tail classes is significantly larger than that for head classes~\cite{cao2019learning}. This provides the theoretical basis for the inequality:
\begin{equation}
    \|p_y^{\text{tail}} - q_y\|_1 \gg \|p_y^{\text{head}} - q_y\|_1.
\end{equation}

\textbf{2. Bound on Background Feature Bias.} \\
Next, we relate this distribution-level deviation to the feature-level background bias $\Delta D_{bg}(y)$. By definition, the expected background feature means are
\begin{equation}
\mu_{bg}(y) = \mathbb{E}_{b \sim p_y}[f_{bg}], \quad
\bar{\mu}_{bg}(y) = \mathbb{E}_{b \sim q_y}[f_{bg}].
\end{equation}

The background bias $\Delta D_{bg}(y)$ can be expanded as
\begin{equation}
\begin{aligned}
\|\Delta D_{bg}(y)\|_1 
&= \| \mu_{bg}(y) - \bar{\mu}_{bg}(y) \|_1 \\
&= \left\| \int f_{bg} \, p_y(b) \, db - \int f_{bg} \, q_y(b) \, db \right\|_1 \\
&= \left\| \int f_{bg} \, \big(p_y(b) - q_y(b)\big) \, db \right\|_1
\end{aligned}
\end{equation}

Assuming the feature representations $f_{bg}$ extracted by a standard deep neural network are inherently bounded (e.g., strictly constrained by non-linear activation functions or normalization layers~\cite{ioffe2015batch,ba2016layer}), there exists a constant $M > 0$ such that $\sup_b \|f_{bg}\|_1 \le M$. Using Hölder's inequality, we obtain an upper bound on the feature bias:
\begin{equation}
\|\Delta D_{bg}(y)\|_1 \le \int \|f_{bg}\|_1 \cdot |p_y(b) - q_y(b)| db \le M \int |p_y(b) - q_y(b)| db = M \|p_y - q_y\|_1.
\end{equation}

This bound indicates that larger background distribution deviation can increase the possible magnitude of background feature shift, which supports the hypothesis in Eq.~\ref{eq:gradient}. In particular, since tail classes have fewer samples, their empirical background distributions can deviate more from the balanced reference, potentially leading to larger background-biased feature shifts compared with head classes. Hence, this motivates a targeted debiasing mechanism for tail classes, as standard empirical risk minimization (ERM) may overfit spurious background correlations for minority categories~\cite{sagawadistributionally}.


\subsection{Formal Analysis of Optimization-level Background Bias (Eq.~\ref{eq:bg_ratio_a})}
\label{sec:gradient_shift_proof}

We provide a concise justification for Eq.~\ref{eq:bg_ratio_a} from spectral bias and implicit low-rank regularization~\cite{arora2019implicit, chaudhuricloser}. Following Sec.~\ref{sec: analysis_b}, we decompose the feature as $f = f_{\mathrm{fg}} + f_{\mathrm{bg}}$, and define:
\begin{equation}
R_t(y) = \frac{1}{1 + \gamma_t(y)}, \quad
\gamma_t(y) = \frac{G_{\mathrm{fg},t}(y)}{G_{\mathrm{bg},t}(y)}.
\end{equation}

\textbf{Spectral bias and low-rank learning dynamics.}
Under gradient descent, deep networks exhibit spectral bias, favoring directions with larger singular values~\cite{arora2019implicit}. Let $a_k(t)$ denote the alignment of the model with the $k$-th spectral component. Its evolution approximately follows:
\begin{equation}
\frac{d}{dt} a_k(t) \propto \sigma_k^2 a_k(t),
\end{equation}
which leads to exponentially faster growth along dominant directions. Empirically, deep representations exhibit low-rank structures during training~\cite{chaudhuricloser}, meaning that most energy concentrates in a few principal components. Background features, due to spatial redundancy and shared patterns across samples, naturally form a low-rank structure with dominant singular values. In contrast, foreground features are more diverse and distributed across many weaker components. As a result, gradient descent preferentially fits background-related components early in training, establishing an inherent optimization bias toward low-rank structures.

\textbf{Effect of sample size on effective rank.}
For class $y$, the effective rank of foreground features depends on the sample size $N_y$. When $N_y$ is large (head classes), diverse foreground patterns are observed, allowing multiple spectral components of $f_{\mathrm{fg}}$ to be learned. In contrast, for tail classes with limited samples, the observed foreground variations are restricted, reducing the usable rank of $f_{\mathrm{fg}}$. This further amplifies the low-rank bias: the background components remain stable due to cross-class sharing, while the high-rank foreground structure becomes increasingly difficult to capture. Consequently, optimization for tail classes gradually shifts toward low-rank (background) directions.

\textbf{Evolution of gradient ratio.}
Combining spectral bias and effective-rank reduction, the relative strength between foreground and background gradients evolves multiplicatively over time. Following standard analyses of gradient flow in linearized regimes~\cite{arora2019implicit}, the ratio $\gamma_t(y)$ can be approximated as:
\begin{equation}
\gamma_t(y) \approx \gamma_0(y)\exp\!\left((\sigma_{\mathrm{fg}}^2 - \sigma_{\mathrm{bg}}^2)t\right),
\end{equation}
where the exponent reflects the difference in effective learning rates. For tail classes, the reduced effective rank suppresses foreground components, leading to $\sigma_{\mathrm{fg}}^2 < \sigma_{\mathrm{bg}}^2$, and thus $\gamma_t(y) \to 0$. Substituting into $R_t(y)$ gives $R_t(y) \to 1$, implying $\Delta R_t(y) > 0$. For head classes, sufficient samples preserve the foreground rank and prevent $\gamma_t(y)$ from collapsing. Meanwhile, background components are fitted early and their gradients diminish over time, leading to a non-increasing $R_t(y)$ (\ie, $\Delta R_t(y) \simeq 0$). Therefore:
\begin{equation}
\Delta R_t^{\mathrm{tail}} > \Delta R_t^{\mathrm{head}}.
\end{equation}


\subsection{Detailed Derivation of \texorpdfstring{$S_{d,u}$}{S\_\{d,u\}} in Eq.~\ref{eq:energy}}
\label{sec:score_derivation}

In long-tailed visual recognition, models tend to overfit to background feature, particularly for tail classes. Addressing this optimization-level bias typically requires extra modules. However, introducing learnable parameters often exacerbates class imbalance due to biased gradient updates. To rectify this without additional parameters, we propose estimating the inherent foreground distinctiveness of each neuron based on its feature statistics. Inspired by the energy-based local spatial formulation in SimAM~\cite{yang2021simam}, we hypothesize that target-related foreground features exhibit distinctive activation patterns compared to the widely distributed, background feature. To quantify this foreground distinctiveness, we measure the linear separability between the target neuron $F_{d,u}$ and the remaining background neurons $F_{d,v}$ ($v \neq u$). Assigning a target label of $1$ to $F_{d,u}$ and $-1$ to all other neurons, the energy function to optimize the linear transform weights ($w, b$) is formulated as: 
\begin{equation}
    e(w, b) = \frac{1}{M-1} \sum_{v \neq u} (-1 - (w F_{d,v} + b))^2 + (1 - (w F_{d,u} + b))^2 + \eta w^2,
\end{equation}
where $\eta w^2$ serves as an $L_2$ regularizer. By taking the partial derivatives of the objective with respect to $w$ and $b$ and setting them to zero, we obtain a fast closed-form solution:

\begin{equation}
    w^* = -\frac{2(F_{d,u} - \mu_t)}{(F_{d,u} - \mu_t)^2 + 2\sigma_t^2 + 2\eta}, \quad b^* = -\frac{1}{2}(F_{d,u} + \mu_t)w^*,
\end{equation}
where $\mu_t$ and $\sigma_t^2$ are the mean and variance computed over all neurons in the channel except the target $F_{d,u}$. To circumvent the computational overhead of iterative spatial calculations, we approximate them using the global channel mean $\mu_d$ and variance $\sigma_d^2$ calculated over all $M$ neurons. Substituting $w^*$ and $b^*$ back yields the minimal energy $e^*$:
\begin{equation}
    e^* = \frac{4(\sigma_d^2 + \eta)}{(F_{d,u} - \mu_d)^2 + 2\sigma_d^2 + 2\eta}.
\end{equation}

A lower minimal energy $e^*$ mathematically indicates that the target neuron $F_{d,u}$ strongly deviates from the background-dominated channel context. While prior work treats this merely as general visual saliency~\cite{yang2021simam,woo2018cbam,chaudhari2021attentive}, we repurpose it as a highly reliable, unsupervised proxy for foreground information in imbalanced learning. Thus, the inherent foreground contribution of the neuron can be represented by the inverse of the minimal energy ($1/e^*$):
\begin{equation}
    \frac{1}{e^*} = \frac{(F_{d,u} - \mu_d)^2 + 2\sigma_d^2 + 2\eta}{4(\sigma_d^2 + \eta)} = \frac{(F_{d,u} - \mu_d)^2}{4(\sigma_d^2 + \eta)} + 0.5.
\end{equation}

By omitting the constant scaling offset ($0.5$) and replacing the regularizer $\eta$ with $\epsilon$ for numerical stability, we map this dynamic distinctiveness through a sigmoid function. This directly yields the bounded score $S_{d,u}$ utilized in our main framework (Eq. ~\ref{eq:energy}):
\begin{equation}
    S_{d,u} = \text{sigmoid}\left( \frac{(F_{d,u} - \mu_d)^2}{4(\sigma_d^2 + \epsilon)} \right).
\end{equation}

Additionally, we emphasize a fundamental distinction from conventional attention modules~\cite{zhang2020adaptive,jiang2023probabilistic} regarding \textbf{gradient bias and parameterization}. Standard attention modules rely on \textit{learnable} parameters to generate dynamic scores. Under severe long-tailed imbalance, the gradient updates for these parameters are overwhelmingly dominated by head classes. Consequently, to minimize the overall empirical risk, these modules inevitably learn to attend to and universally amplify spurious head-class backgrounds, inadvertently acting as ``bias amplifiers'' In contrast, our approach derives $S_{d,u}$ purely from the \textit{intrinsic spatial statistics} of the feature map, rendering it strictly parameter-free and immune to class-imbalanced gradient domination. By aggregating these statistical scores into a deterministic background penalty map $B_u$ (Eq.~\ref{eq:background score}), our BFR module shifts from generic feature reweighting to explicit spatial decoupling. It mathematically penalizes spurious background-foreground co-occurrences (Eq.~\ref{eq:fr}), preventing tail-class representations from collapsing into dominant background features.


\subsection{Why OFBD Mitigates Background Bias}
\label{sec:ofbd_bias_mitigation}

We further justify that the two components of OFBD directly reduce the two bias terms derived in Sec.~\ref{sec:distribution_shift_proof} and Sec.~\ref{sec:gradient_shift_proof}. Specifically, FG-CutMix mitigates the distribution-level bias in Eq.~\ref{eq:gradient}, while BFR suppresses the optimization-level bias in Eq.~\ref{eq:bg_ratio}.

\textbf{FG-CutMix reduces distribution-level background Bias.}
From Sec.~\ref{sec:distribution_shift_proof}, the background feature bias is upper-bounded by the background distribution deviation. Thus, it is sufficient to show that FG-CutMix reduces $\|p_y-q_y\|_1$. Let $\tilde{p}_y$ denote the effective background distribution induced by FG-CutMix. Since FG-CutMix preserves the target foreground and replaces the complementary background, $\tilde{p}_y$ can be written as:
\begin{equation}
\tilde{p}_y=(1-\eta_y)p_y+\eta_y p_{\mathrm{mix}}, \quad \eta_y\in[0,1],
\end{equation}
where $p_{\mathrm{mix}}$ denotes the background distribution introduced by mixed samples. Since $p_{\mathrm{mix}}$ aggregates backgrounds from more samples, it provides a closer estimate of the ideal background distribution:
\begin{equation}
\|p_{\mathrm{mix}}-q_y\|_1 \le \|p_y-q_y\|_1 .
\end{equation}
By the convexity of the $L_1$ norm:
\begin{equation}
\|\tilde{p}_y-q_y\|_1
\le (1-\eta_y)\|p_y-q_y\|_1+\eta_y\|p_{\mathrm{mix}}-q_y\|_1
\le \|p_y-q_y\|_1 .
\end{equation}
Therefore, using the bound established in Sec.~\ref{sec:distribution_shift_proof}, we obtain:
\begin{equation}
\|\Delta \tilde{D}_{bg}(y)\|_1 \le \|\Delta D_{bg}(y)\|_1 .
\end{equation}
This shows that FG-CutMix reduces the upper bound of distribution-level background bias. Moreover, the RL reward in Eq.~\ref{eq:reward} favors regions whose mixed samples remain more predictive of the target label than the background-source label, preventing the mixed distribution from being dominated by spurious background semantics.

\begin{figure*}[t]
    \centering
    \scriptsize
    \setlength{\tabcolsep}{2pt}
    \renewcommand{\arraystretch}{1.2} 

    \begin{tabular}{@{}c cccc@{}}
        
        \raisebox{0.06\textwidth}{\rotatebox{90}{\textbf{\normalsize Head}}} &
        \includegraphics[width=0.18\textwidth]{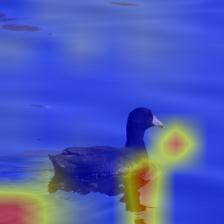} &
        \includegraphics[width=0.18\textwidth]{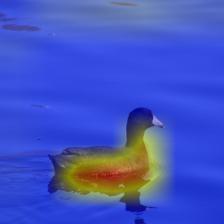} &
        \includegraphics[width=0.18\textwidth]{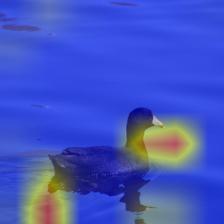} &
        \includegraphics[width=0.18\textwidth]{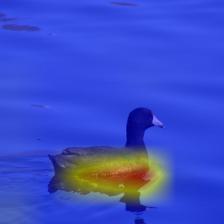} \\
        & CE: American coot & CE+OFBD: American coot & BCL: American coot & BCL+OFBD: American coot \\[1.5ex] 

        \raisebox{0.06\textwidth}{\rotatebox{90}{\textbf{\normalsize Medium}}} &
        \includegraphics[width=0.18\textwidth]{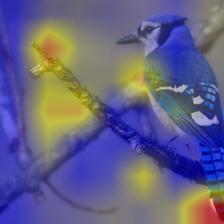} &
        \includegraphics[width=0.18\textwidth]{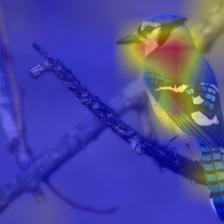} &
        \includegraphics[width=0.18\textwidth]{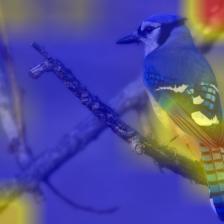} &
        \includegraphics[width=0.18\textwidth]{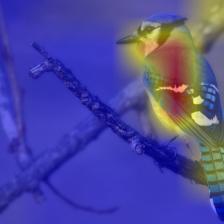} \\
        & CE: Jay & CE+OFBD: Jay & BCL: Jay & BCL+OFBD: Jay \\[1.5ex]

        \raisebox{0.06\textwidth}{\rotatebox{90}{\textbf{\normalsize Tail}}} &
        \includegraphics[width=0.18\textwidth]{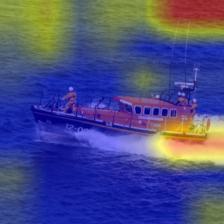} &
        \includegraphics[width=0.18\textwidth]{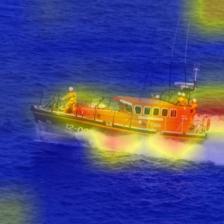} &
        \includegraphics[width=0.18\textwidth]{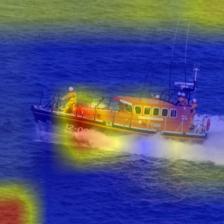} &
        \includegraphics[width=0.18\textwidth]{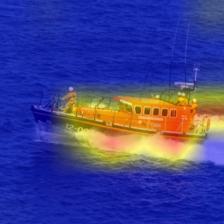} \\
        & CE: Lifeboat & CE+OFBD: Lifeboat & BCL: Lifeboat & BCL+OFBD: Lifeboat \\
        
    \end{tabular}

    \caption{
    Extended pretrained-CAM attention maps of representative \textbf{Head, Medium, and Tail} classes under CE, CE+OFBD, BCL, and BCL+OFBD. OFBD reduces background co-activation and makes attention more object-focused across different baselines and class-frequency groups.
    }
    \label{fig:more_cam_visualizations}
\end{figure*}

\textbf{BFR reduces optimization-level background bias.}
From Sec.~\ref{sec:gradient_shift_proof}, the optimization-level bias is characterized by the increase of the background gradient ratio. According to Eq.~\ref{eq:fr}, BFR rectifies the feature by assigning each spatial location a background-aware weight $(1-\lambda B_u)$. Since a larger $B_u$ indicates stronger background bias, background-dominant locations receive smaller weights than foreground-dominant locations:
\begin{equation}
B_u^{bg}>B_v^{fg} \Rightarrow 1-\lambda B_u^{bg}<1-\lambda B_v^{fg}.
\end{equation}
Therefore, the background-related feature component, and hence its gradient contribution, is more strongly suppressed than the foreground component:
\begin{equation}
\|\tilde{g}_{bg,t}(y)\|_1 \le \|g_{bg,t}(y)\|_1,\quad
\frac{\|\tilde{g}_{bg,t}(y)\|_1}{\|g_{bg,t}(y)\|_1}
<
\frac{\|\tilde{g}_{fg,t}(y)\|_1}{\|g_{fg,t}(y)\|_1}.
\end{equation}
Substituting the rectified gradients into the background gradient ratio defined in Eq.~\ref{eq:bg_ratio} yields
\begin{equation}
\tilde{R}_t(y)
=
\frac{\|\tilde{g}_{bg,t}(y)\|_1}
{\|\tilde{g}_{fg,t}(y)\|_1+\|\tilde{g}_{bg,t}(y)\|_1}
\le
\frac{\|g_{bg,t}(y)\|_1}
{\|g_{fg,t}(y)\|_1+\|g_{bg,t}(y)\|_1}
=
R_t(y).
\end{equation}
Thus,
\begin{equation}
\tilde{R}_t(y)\le R_t(y), \quad \Delta\tilde{R}_t(y)\le \Delta R_t(y).
\end{equation}
This shows that BFR suppresses the optimization-level background shift by reducing the relative contribution of background gradients.

\textbf{Joint effect.}
Combining the above results, OFBD reduces both bias terms:
\begin{equation}
\|\Delta\tilde{D}_{bg}(y)\|_1\le \|\Delta D_{bg}(y)\|_1,\quad
\Delta\tilde{R}_t(y)\le \Delta R_t(y).
\end{equation}
Hence, FG-CutMix mitigates the enlarged background distribution deviation of tail classes, while BFR suppresses their amplified background gradient. This provides a formal explanation for why OFBD alleviates both distribution-level and optimization-level background bias in long-tailed recognition.


\begin{table}[t]
    \centering
    \begin{minipage}[t]{0.48\textwidth}
        \centering
        \caption{Ablation on hyper-parameters $\beta$ and $\alpha$.}
        \label{tab:ablation_beta_alpha}
        \setlength{\tabcolsep}{6pt}
        \begin{tabular}{cc@{\hspace{1.6em}}cc}
            \toprule
            \rowcolor[gray]{0.9}
            \textbf{$\beta$} & \textbf{Acc. (\%)} & \textbf{$\alpha$} & \textbf{Acc. (\%)} \\
            \midrule
            0.005 & 53.7 & 0.1  & 53.6 \\
            0.01  & 53.9 & 0.5  & 53.8 \\
            0.015 & 54.0 & 0.75 & 54.1 \\
            \textbf{0.02}  & \textbf{54.2} & \textbf{1.0}  & \textbf{54.2} \\
            0.025 & 54.1 & 1.25 & 53.6 \\
            0.03  & 53.8 & 1.5  & 53.7 \\
            \bottomrule
        \end{tabular}
    \end{minipage}%
    \hfill
    \begin{minipage}[t]{0.48\textwidth}
        \centering
        \caption{Ablation on hyper-parameters $\lambda$ and $\gamma$.}
        \label{tab:ablation_lambda_gamma}
        \setlength{\tabcolsep}{6pt}
        \begin{tabular}{cc@{\hspace{1.6em}}cc}
            \toprule
            \rowcolor[gray]{0.9}
            \textbf{$\lambda$} & \textbf{Acc. (\%)} & \textbf{$\gamma$} & \textbf{Acc. (\%)} \\
            \midrule
            0.2 & 53.6 & 0.2 & 53.3 \\
            0.3 & 53.9 & 0.4 & 53.7 \\
            0.4 & 53.7 & 0.6 & 54.1 \\
            \textbf{0.5} & \textbf{54.2} & \textbf{0.8} & \textbf{54.2} \\
            0.6 & 54.1 & 1.0 & 53.8 \\
            0.7 & 53.9 & 1.2 & 53.4 \\
            \bottomrule
        \end{tabular}
    \end{minipage}
\end{table}
\begin{table*}[t]
\centering
\small
\renewcommand{\arraystretch}{1.15}

\begin{minipage}[t]{0.49\textwidth}
\centering
\caption{Background distribution shift on CIFAR-100-LT ($r=100$). Lower is better.}
\label{tab:bg_distribution_shift}
\setlength{\tabcolsep}{4pt}
\rowcolors{1}{white}{white}
\begin{tabular}{lcccc}
\toprule
\rowcolor[gray]{0.9}
\textbf{Method} & \textbf{Many}$\downarrow$ & \textbf{Med.}$\downarrow$ & \textbf{Few}$\downarrow$ & \textbf{All}$\downarrow$ \\
\midrule
CE & 0.79 & 0.84 & 0.86 & 0.83 \\
+ FG-CutMix & 0.78 & 0.83 & 0.80 & 0.82 \\
+ BFR & \textbf{0.79} & 0.84 & 0.81 & 0.80 \\
+ OFBD & 0.80 & \textbf{0.79} & \textbf{0.78} & \textbf{0.79} \\
\bottomrule
\end{tabular}
\end{minipage}
\hfill
\begin{minipage}[t]{0.49\textwidth}
\centering
\caption{Background gradient ratio shift on CIFAR-100-LT ($r=100$). Lower is better.}
\label{tab:bg_gradient_ratio}
\setlength{\tabcolsep}{4pt}
\rowcolors{1}{white}{white}
\begin{tabular}{lcccc}
\toprule
\rowcolor[gray]{0.9}
\textbf{Method} & \textbf{Many}$\downarrow$ & \textbf{Med.}$\downarrow$ & \textbf{Few}$\downarrow$ & \textbf{All}$\downarrow$ \\
\midrule
CE & 0.4 & 0.8 & 3.6 & 1.5 \\
+ FG-CutMix & -3.4 & -4.7 & -6.2 & -4.7 \\
+ BFR & -4.6 & -5.2 & -6.5 & -5.0 \\
+ OFBD & \textbf{-5.6} & \textbf{-6.1} & \textbf{-7.0} & \textbf{-6.2} \\
\bottomrule
\end{tabular}
\end{minipage}

\end{table*}


\section{Extensive Visualizations}
\label{sec:visualizations}

\subsection{Extended Background-bias Visualization}
\label{sec:extended_cam}

To further support the analysis in Sec.~\ref{sec: ablation}, we provide extended attention-map visualizations for representative head, medium, and tail classes. As shown in Fig.~\ref{fig:more_cam_visualizations}, we compare CE, CE+OFBD, BCL, and BCL+OFBD using pretrained-CAM attention maps. The selected samples cover different class-frequency groups, including head, medium, and tail classes. For both CE and BCL baselines, the attention maps often co-activate target objects and surrounding backgrounds, indicating background-biased representation. After applying OFBD, the activated regions become more concentrated on target-related foregrounds, while irrelevant background responses are reduced. This trend is observed across different baselines and class-frequency groups, suggesting that OFBD consistently improves object-focused representation rather than only benefiting a specific model or class group. These visualizations further validate that OFBD alleviates background reliance.

\section{Additional Experimental Results}
\label{sec:add_experiments}
\subsection{Hyper-parameters Ablation}
\label{sec:hyper_parameters_ablation}

We conduct hyper-parameter ablations on CIFAR-100-LT with imbalance ratio $100$ to analyze the sensitivity of OFBD. As shown in Tables~\ref{tab:ablation_beta_alpha} and~\ref{tab:ablation_lambda_gamma}, OFBD remains stable across a wide range of hyper-parameter values. For the RL-based foreground selector, $\beta$ controls the KL regularization strength and $\alpha$ controls the auxiliary selector supervision. The best performance is achieved at $\beta=0.02$ and $\alpha=1.0$, both reaching $54.2\%$ accuracy. Smaller values provide insufficient regularization or supervision, while overly large values slightly degrade performance.

For the overall OFBD framework, $\lambda$ controls the strength of background feature rectification and $\gamma$ controls the weight of policy learning. The best accuracy is obtained at $\lambda=0.5$ and $\gamma=0.8$, both reaching $54.4\%$. When $\lambda$ is too small, background-biased features are insufficiently suppressed; when too large, useful foreground information may also be weakened. Similarly, a moderate $\gamma$ provides the best trade-off between selector learning and classification optimization. These results show that OFBD is not sensitive to exact hyper-parameter choices and maintains competitive performance under nearby settings.

\subsection{Ablation on Candidate Pool Size}
\label{sec:candidate_pool_ablation}

We further study the effect of the candidate pool size $K$ in FG-CutMix, where $K$ denotes the number of selected foreground candidates used for reward comparison and relative-advantage estimation. Small $K$ values limit candidate diversity, reducing the selector's ability to distinguish regions with different target-preservation quality. Increasing $K$ improves the chance of selecting foregrounds that better preserve target semantics after background replacement. However, excessively large $K$ increases computation cost and may introduce redundant or noisy candidates.

Fig.~\ref{fig:k_sensitivity} visualizes the accuracy and per-epoch computation cost for different $K$ on CIFAR-100-LT with imbalance ratio $100$. Accuracy improves as $K$ increases from 2 to 6, then slightly decreases for larger $K$, while per-epoch cost rises monotonically. This indicates that a moderate candidate pool provides sufficient diversity for effective foreground selection. We set $K=6$ as the default, achieving a favorable trade-off between accuracy and efficiency.

\begin{figure*}[t]
    \centering
    \begin{subfigure}[t]{0.46\textwidth}
        \centering
        \includegraphics[height=0.22\textheight]{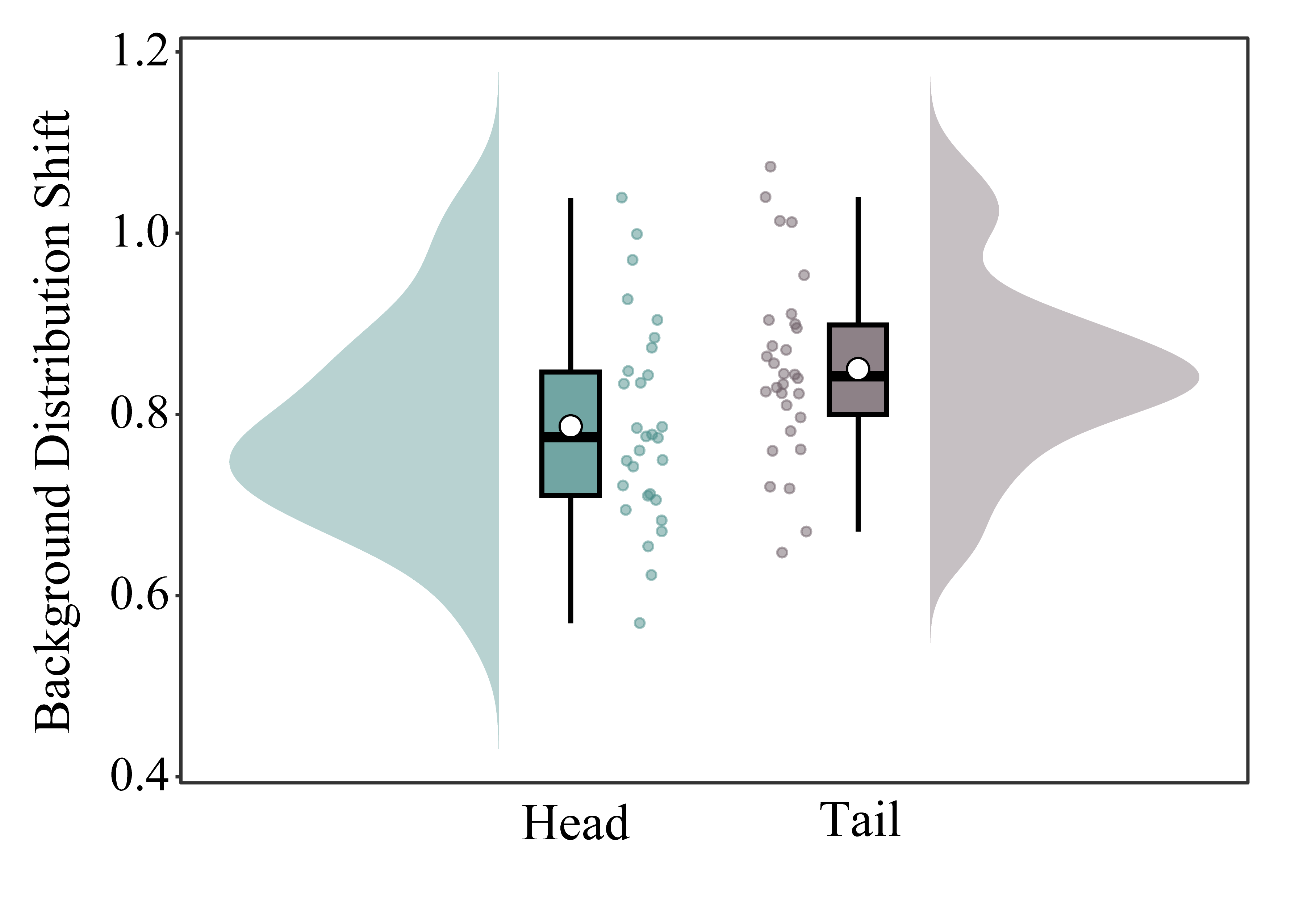}
        \caption{Distribution-level background shift.}
        \label{fig:bg_distribution_shift}
    \end{subfigure}
    \hfill
    \begin{subfigure}[t]{0.50\textwidth}
        \centering
        \includegraphics[height=0.22\textheight]{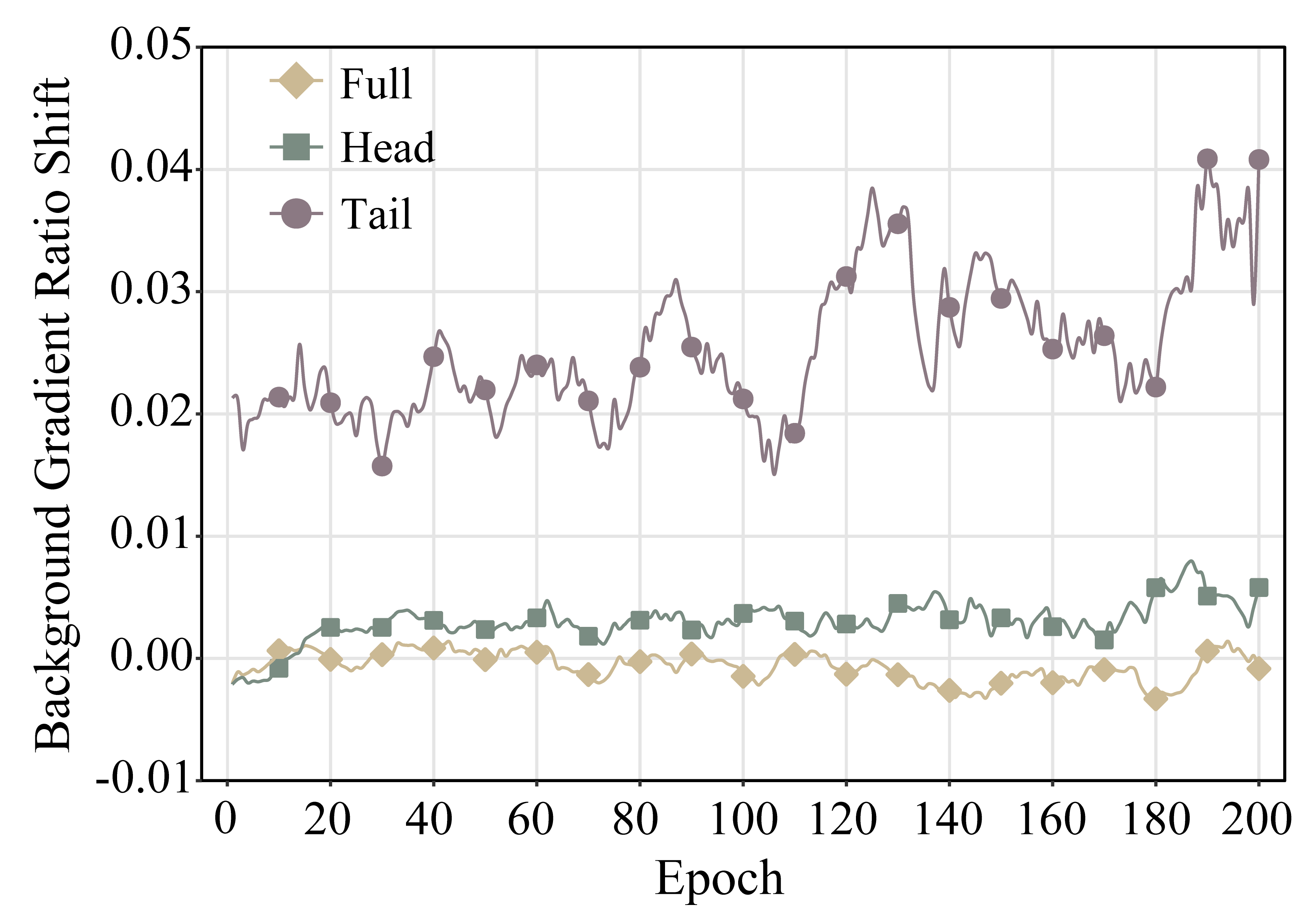}
        \caption{Optimization-level background-gradient-ratio shift.}
        \label{fig:bg_gradient_shift}
    \end{subfigure}
    \caption{Empirical verification of background bias on CIFAR-100-LT with imbalance ratio 100. Tail classes exhibit both larger background distribution shift and stronger background-gradient-ratio shift than head classes, validating the distribution-level and optimization-level bias analyzed in Sec.~\ref{sec: analysis}.}
    \label{fig:bg_bias_analysis}
\end{figure*}


\subsection{Empirical Verification of Background Bias}
\label{sec:bg_bias_verification}

\textbf{Experiment Setting.}
We train ResNet-32~\cite{he2016deep} with cross-entropy on CIFAR-100-LT~\cite{cao2019learning} and CIFAR-100 under identical settings. The CIFAR-100 model serves as the balanced reference for measuring distribution-level background shift. Background regions are obtained with a pretrained CAM~\cite{zhou2016learning,jung2021towards}.

\textbf{Results.}
As shown in Fig.~\ref{fig:bg_bias_analysis}, tail classes suffer from stronger background bias than head classes. Specifically, Fig.~\ref{fig:bg_distribution_shift} shows that tail classes have a larger background distribution shift, indicating that their learned background statistics deviate more from the balanced reference distribution. Meanwhile, Fig.~\ref{fig:bg_gradient_shift} shows that the background-gradient-ratio shift of tail classes increases faster and fluctuates more strongly during training, suggesting that tail-class optimization is more easily driven by background-related gradients. These results support our analysis (Sec.~\ref{Evidence}) that tail degradation is not only caused by sample scarcity, but also by background-biased distribution and optimization.


\subsection{Effect on Background Distribution Shift}
\label{sec:bg_distribution_shift_exp}

Based on the observed distribution-level bias, we further evaluate whether FG-CutMix can reduce the background distribution shift. We measure the deviation between the background feature distribution learned under long-tailed training and that learned from balanced training. Specifically, we train a reference model on CIFAR-100 and use it to estimate the class-wise reference background feature mean $\bar{\mu}_{bg}(y)$. For each long-tailed model, we compute the background feature mean $\mu_{bg}(y)$ on the same balanced test set, where background regions are obtained by pretrained CAM masks. The background distribution shift is measured as:
\begin{equation}
\Delta D_{bg}(y)=\|\mu_{bg}(y)-\bar{\mu}_{bg}(y)\|_1 .
\end{equation}
We report the averaged $\Delta D_{bg}(y)$ over Many-, Medium-, and Few-shot classes. Lower values indicate a smaller distribution-level background shift.

As shown in Table~\ref{tab:bg_distribution_shift}, standard long-tailed training exhibits a clear background distribution shift, especially on Few-shot classes. Standard CutMix only provides limited reduction, since random region replacement may still preserve or introduce spurious background correlations. In contrast, FG-CutMix consistently reduces $\Delta D_{bg}(y)$, indicating that preserving target-related foregrounds while diversifying backgrounds makes the learned background distribution closer to the balanced reference distribution.


\subsection{Effect on Background Gradient Ratio}
\label{sec:bg_gradient_ratio_exp}

Based on the observed optimization-level bias, we further examine whether BFR can suppress the increase of the background gradient ratio. We measure the increase of the background gradient ratio defined in Eq.~\ref{eq:bg_ratio}. Foreground and background gradient components are computed using the same foreground-background masks as in Sec.~\ref{sec:bg_bias_verification}. We report the averaged $\Delta R_t(y)$ over Many-, Medium-, and Few-shot classes. Lower values indicate that the optimization process is less dominated by background-related gradients.

As shown in Table~\ref{tab:bg_gradient_ratio}, both FG-CutMix and BFR substantially reduce the background gradient ratio compared with standard CE, with the effect most pronounced for Few-shot classes. FG-CutMix decreases the gradient shift at the input level by diversifying backgrounds, while BFR further down-weights background-biased spatial features during feature-level rectification. The full OFBD, combining FG-CutMix and BFR, achieves the largest reduction across all class groups, demonstrating the complementarity between input-level background diversification and feature-level rectification in mitigating background-driven optimization.


\begin{figure*}[t]
    \centering

    \begin{subfigure}[t]{0.46\textwidth}
        \centering
        \vspace{0pt}
        \includegraphics[width=\linewidth,height=0.22\textheight,keepaspectratio]{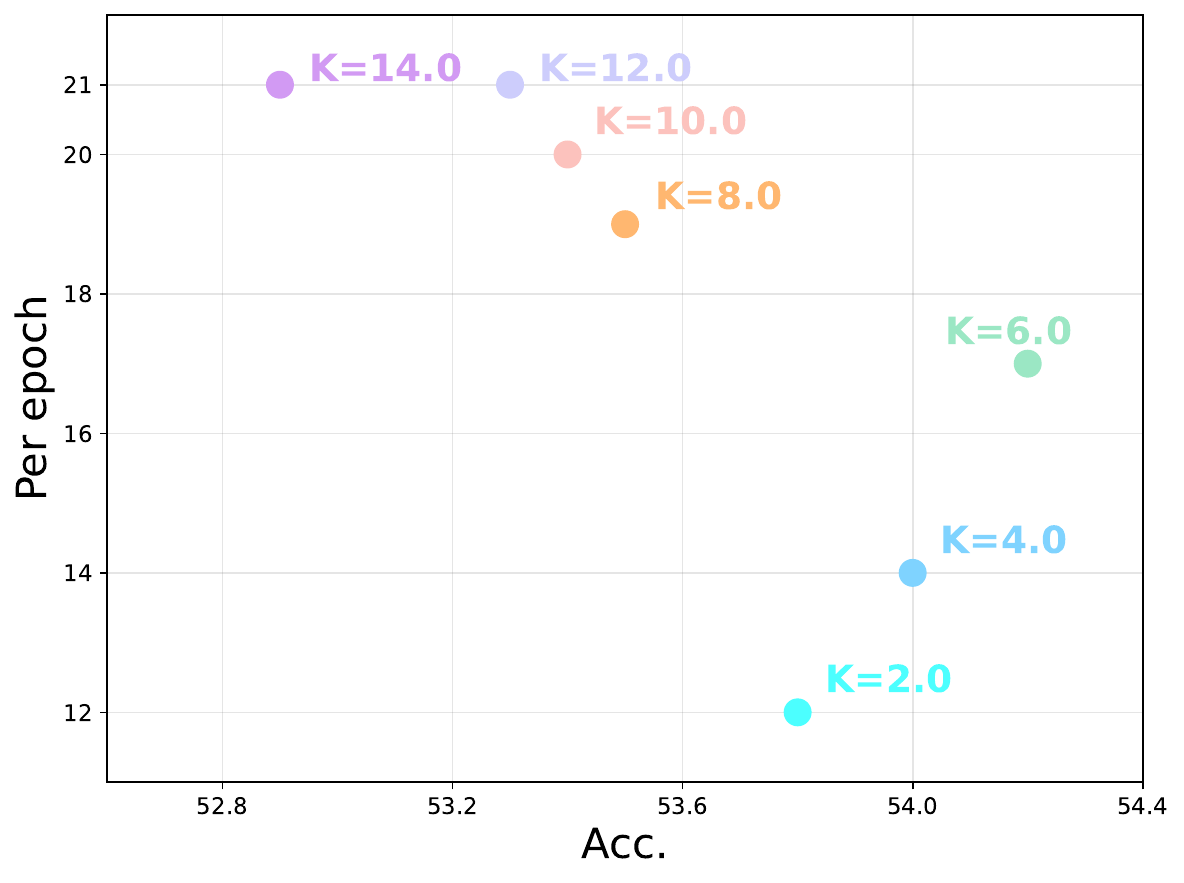}
        \caption{Sensitivity to candidate pool size $K$.}
        \label{fig:k_sensitivity}
    \end{subfigure}
    \hspace{0.04\textwidth}
    \begin{subfigure}[t]{0.46\textwidth}
        \centering
        \vspace{0pt}
        \includegraphics[width=\linewidth,height=0.22\textheight,keepaspectratio]{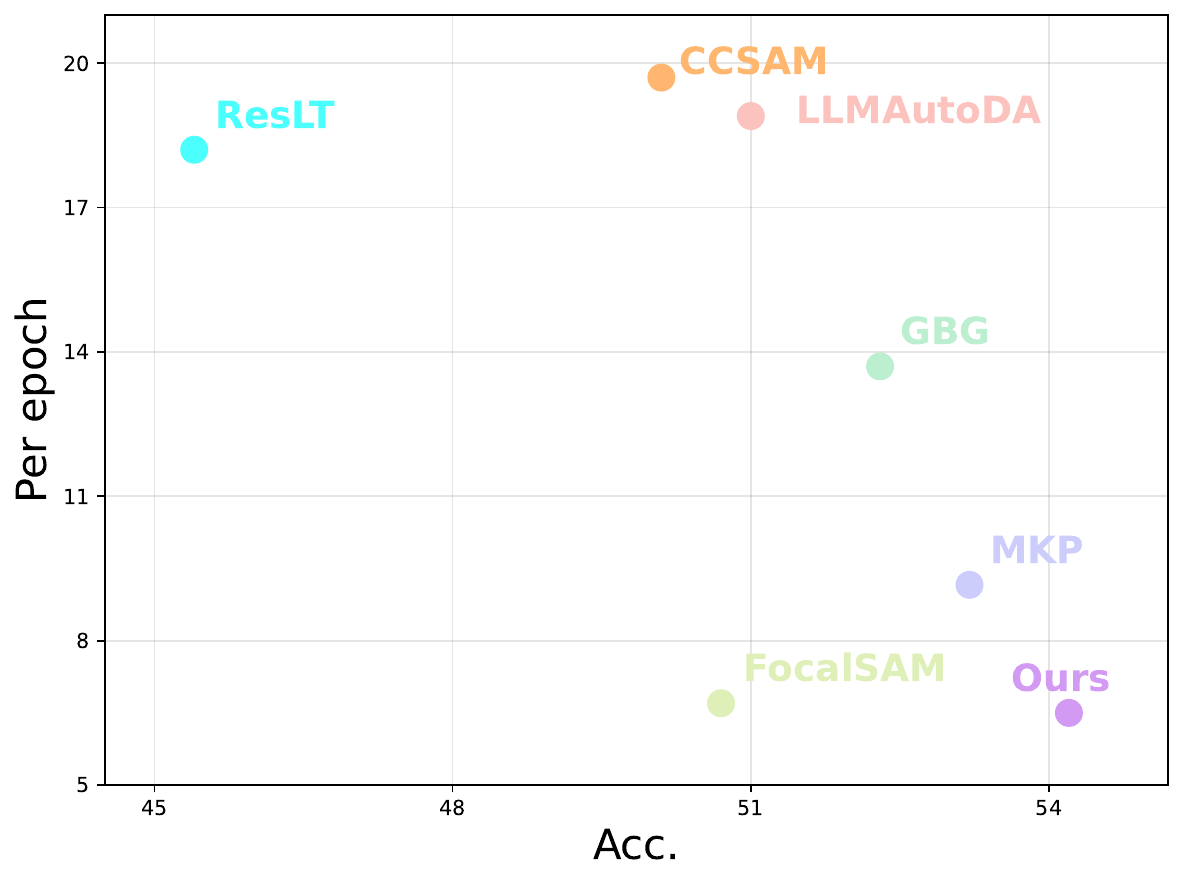}
        \caption{Efficiency-accuracy comparison.}
        \label{fig:computation_cost}
    \end{subfigure}

    \caption{
    Efficiency analysis of OFBD on CIFAR-100-LT with imbalance ratio $100$.
    (a) Accuracy and per-epoch (s) computation cost as a function of the candidate pool size $K$. Accuracy first improves and then slightly decreases for large $K$, while per-epoch cost rises monotonically. $K=6$ is selected as the default for a good efficiency-accuracy trade-off.
    (b) Efficiency-accuracy comparison with representative long-tailed methods, where per-epoch computation cost is measured in seconds. ``Ours'' denotes BCL+OFBD.
    }
    \label{fig:efficiency_analysis}
\end{figure*}


\subsection{Performance Analysis under Long-tailed and Balanced Settings}
\label{sec:performance_analysis}

\begin{figure*}[t]
\centering
\begin{minipage}[t]{0.48\textwidth}
    \centering
    \includegraphics[width=0.95\linewidth]{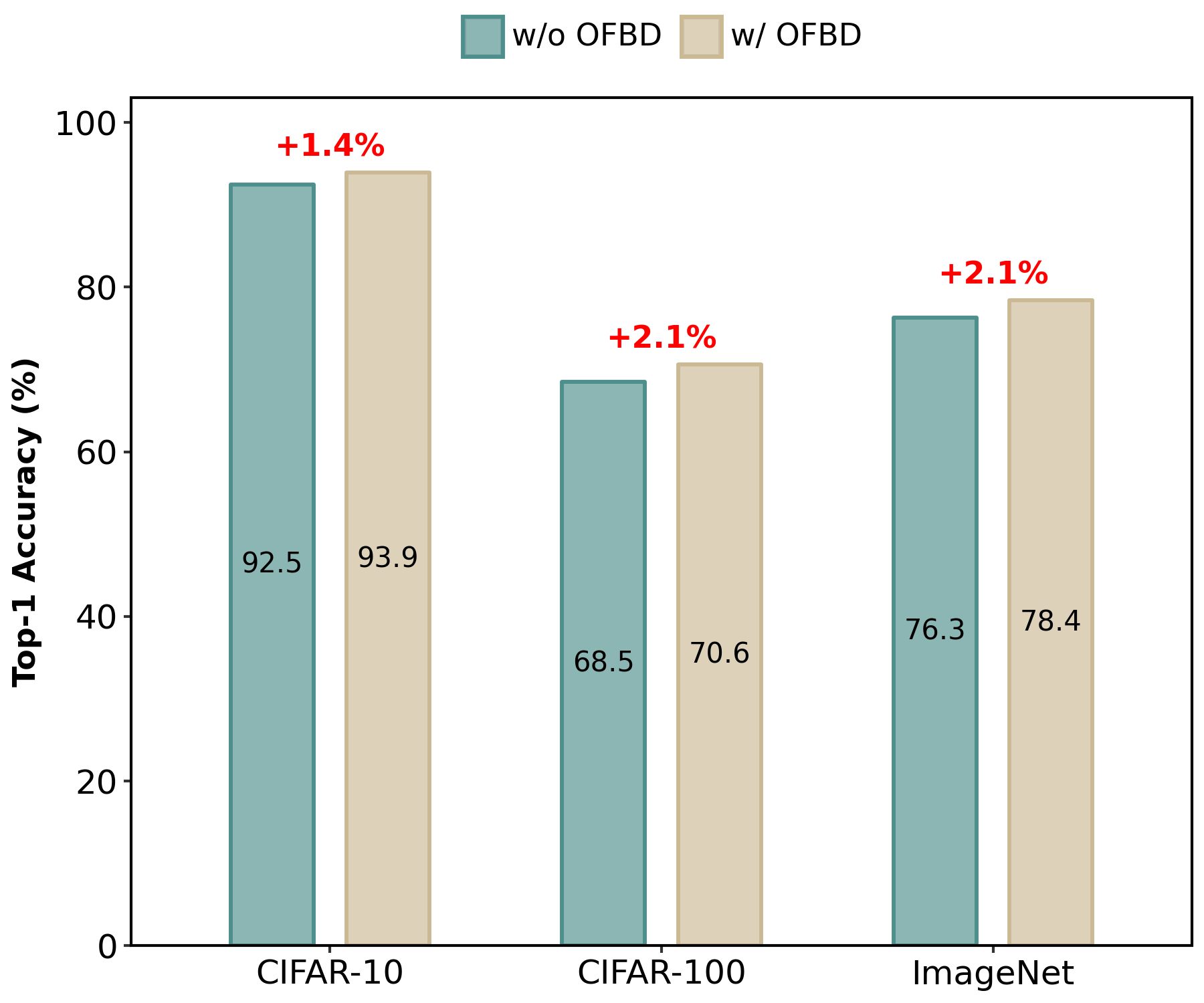}
    \caption{Performance comparison with and without OFBD on balanced datasets ($r=1$). OFBD consistently improves Top-1 accuracy on CIFAR-10, CIFAR-100, and ImageNet.}
    \label{fig:balanced_comparison}
\end{minipage}
\hfill
\begin{minipage}[t]{0.48\textwidth}
    \centering
    \includegraphics[width=0.95\linewidth]{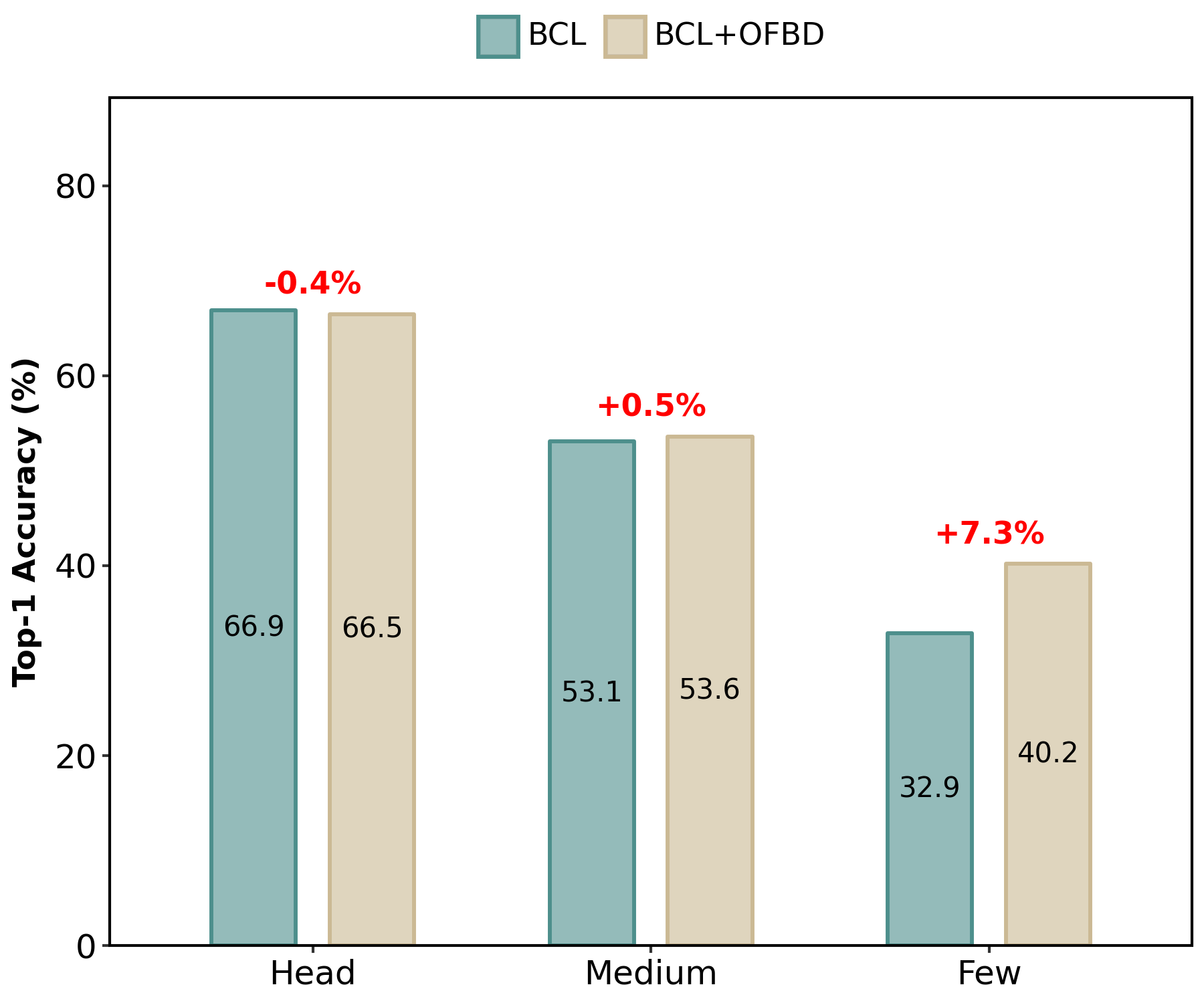}
    \caption{Head-, medium-, and tail-class performance comparison on CIFAR-100-LT with imbalance ratio 100. OFBD mainly improves tail-class accuracy while largely preserving head-class performance.}
    \label{fig:head_tail_comparison}
\end{minipage}
\end{figure*}

We further compare the head-, medium-, and tail-class performance to examine where OFBD brings the largest improvement. As shown in Fig.~\ref{fig:head_tail_comparison}, adding OFBD to BCL substantially improves Few-shot accuracy from $32.9\%$ to $40.2\%$, yielding a $+7.3\%$ gain. Meanwhile, it slightly improves Medium-shot accuracy from $53.1\%$ to $53.6\%$, with only a marginal change on Head classes from $66.9\%$ to $66.5\%$. These results indicate that OFBD mainly benefits data-scarce tail classes while largely preserving head-class performance, demonstrating its effectiveness in alleviating tail-specific background bias.

We also examine whether OFBD is effective under the balanced setting, where the imbalance ratio is $1$. As shown in Fig.~\ref{fig:balanced_comparison}, OFBD consistently improves the baseline without OFBD on balanced CIFAR-10, CIFAR-100, and ImageNet, with gains of $+1.4\%$, $+2.1\%$, and $+2.1\%$, respectively. This demonstrates that OFBD does not rely on long-tailed class distributions and can still provide object-focused regularization benefits when the class distribution is uniform.


\subsection{Further Ablation of FG-CutMix}
\label{sec:fgcutmix_gradient_analysis}

\begin{figure*}[t]
    \centering

    \begin{subfigure}[t]{0.46\textwidth}
        \centering
        \vspace{0pt}
        \includegraphics[width=\linewidth,height=0.22\textheight,keepaspectratio]{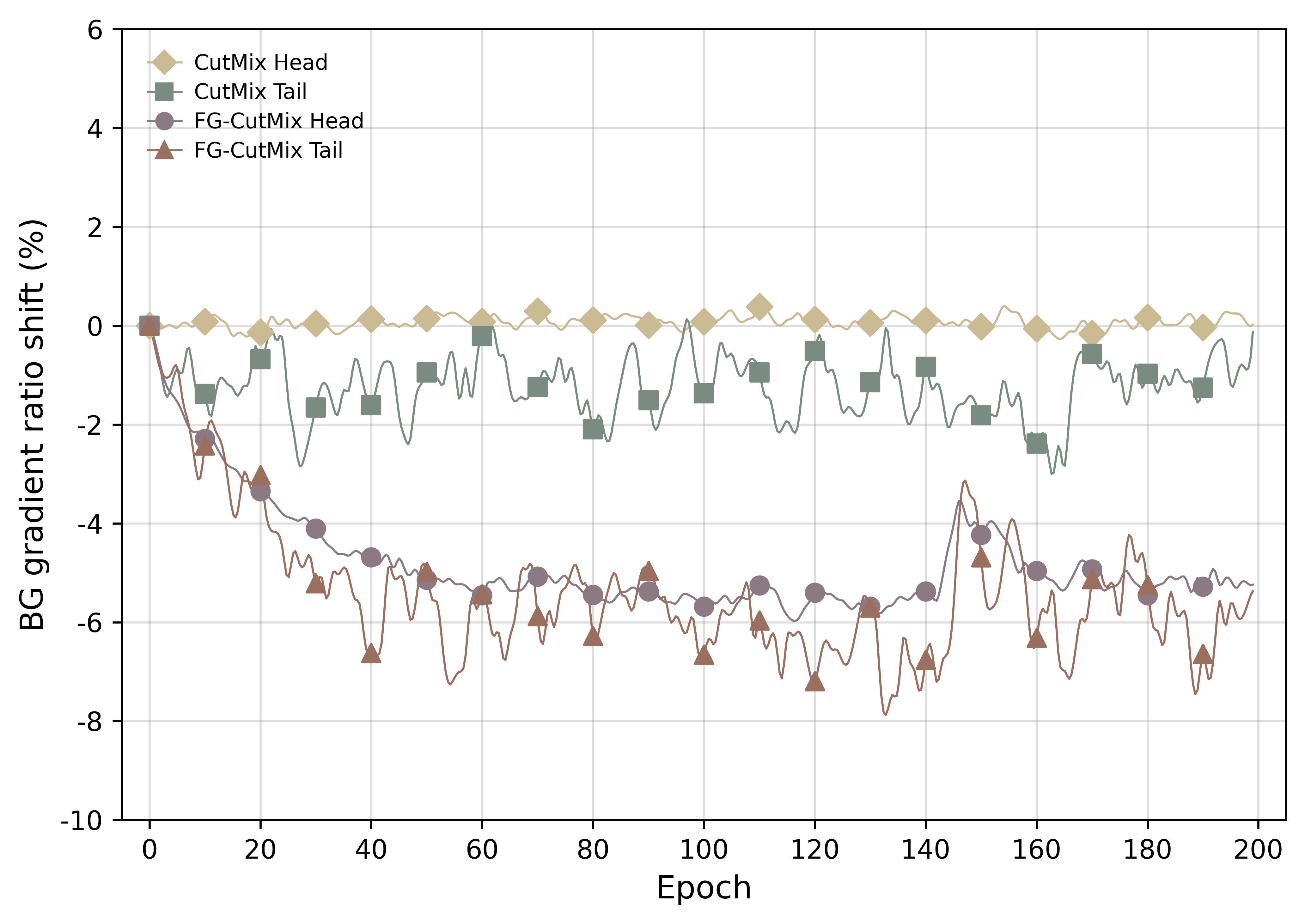}
        \caption{BG gradient ratio shift.}
        \label{fig:fgcutmix_gradient_shift}
    \end{subfigure}
    \hspace{0.04\textwidth}
    \begin{subfigure}[t]{0.46\textwidth}
        \centering
        \vspace{0pt}
        \includegraphics[width=\linewidth,height=0.22\textheight,keepaspectratio]{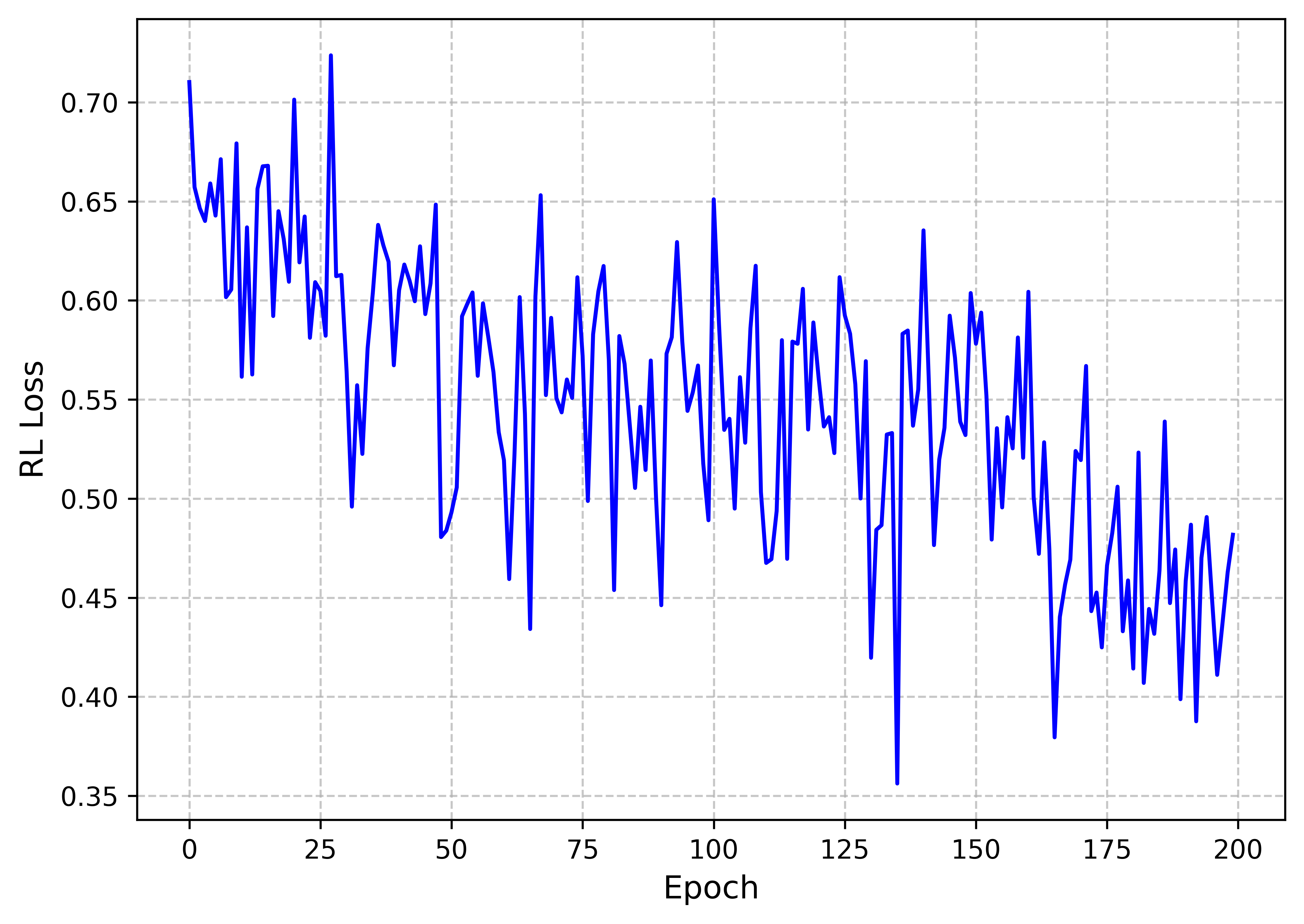}
        \caption{RL loss curve of the selector.}
        \label{fig:selector_loss_curve}
    \end{subfigure}

    \caption{
    Further analysis of FG-CutMix on CIFAR-100-LT with imbalance ratio $100$.
    (a) FG-CutMix reduces tail-class background-gradient-ratio shift compared with standard CutMix, indicating weaker background-driven optimization.
    (b) The RL loss of the foreground selector remains bounded and gradually stabilizes during training, indicating stable optimization of the RL-based selector.
    }
    \label{fig:fgcutmix_analysis}
\end{figure*}

Table~\ref{tab:rl_selection_ablation} shows that FG-CutMix with RL-based foreground selection outperforms standard CutMix with random selection. To further analyze this improvement, we compare their optimization-level background bias by measuring the background-gradient-ratio shift on CIFAR-100-LT with imbalance ratio $100$. As shown in Fig.~\ref{fig:fgcutmix_gradient_shift}, both CutMix and FG-CutMix keep the head-class shift relatively small, suggesting that head-class optimization is less sensitive to region selection.

For tail classes, however, standard CutMix exhibits a stronger background-gradient-ratio shift, indicating that random replacement may preserve background cues or discard target-related foreground evidence. In contrast, FG-CutMix consistently produces a lower tail-class shift during training. This confirms that RL-based foreground region selection not only improves accuracy, as shown in Table~\ref{tab:rl_selection_ablation}, but also reduces background-driven optimization for tail classes.

We further examine the training stability of the RL-based foreground selector. As shown in Fig.~\ref{fig:selector_loss_curve}, the selector loss exhibits stochastic fluctuations due to random proposal sampling and policy-gradient optimization, but remains bounded and gradually stabilizes throughout training. This suggests that the selector does not suffer from unstable policy updates. Together with the accuracy improvement in Table~\ref{tab:rl_selection_ablation} and the reduced tail-class background-gradient shift in Fig.~\ref{fig:fgcutmix_gradient_shift}, these results indicate that the RL-based selector can be optimized stably while learning foreground regions that better mitigate background-driven optimization.


\subsection{Efficiency Analysis}
\label{sec:efficiency_analysis}

We evaluate the efficiency-accuracy trade-off of BCL+OFBD on CIFAR-100-LT with imbalance ratio 100. As shown in Fig.~\ref{fig:computation_cost}, BCL+OFBD achieves the highest accuracy while maintaining a low per-epoch training cost. In particular, compared with MKP, BCL+OFBD reduces the computation cost by $2.66$ seconds per epoch and improves Top-1 accuracy by $1.0\%$. These results show that OFBD provides an efficient plug-in solution for long-tailed recognition, benefiting from its lightweight background-aware feature rectification and training-only foreground-guided augmentation.

\begin{figure*}[t]
    \centering

    \begin{subfigure}[t]{0.28\textwidth}
        \centering
        \includegraphics[width=\linewidth]{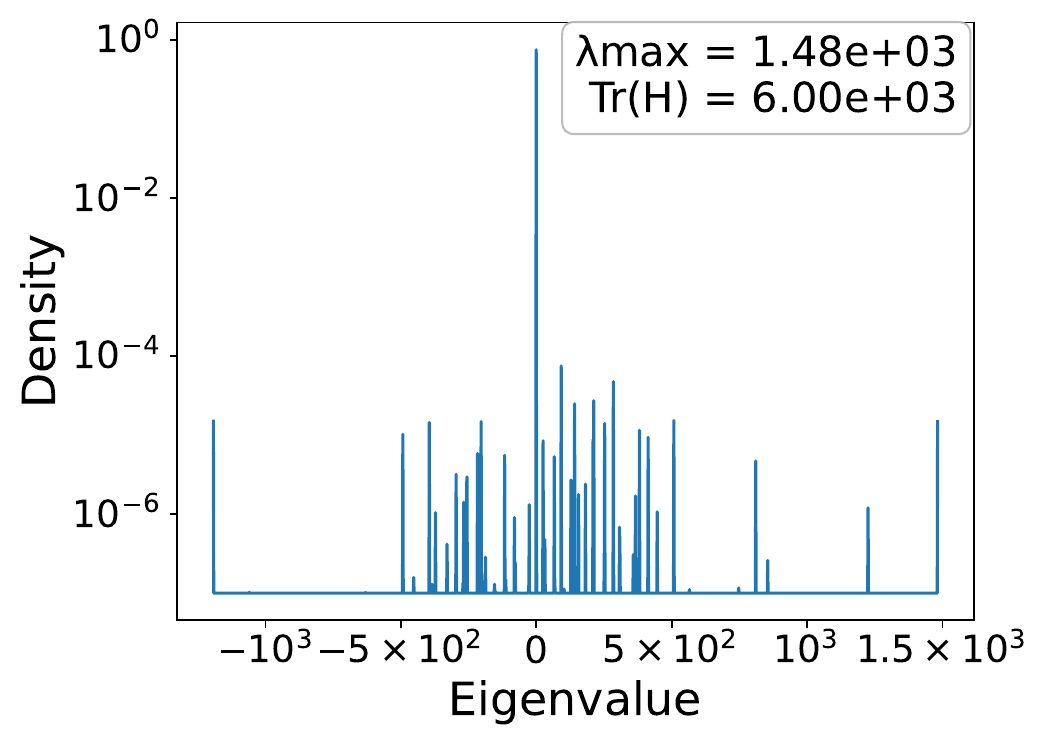}
        \caption{CE: Head Classes}
        \label{fig:ce_h}
    \end{subfigure}
    \hspace{0.04\textwidth}
    \begin{subfigure}[t]{0.28\textwidth}
        \centering
        \includegraphics[width=\linewidth]{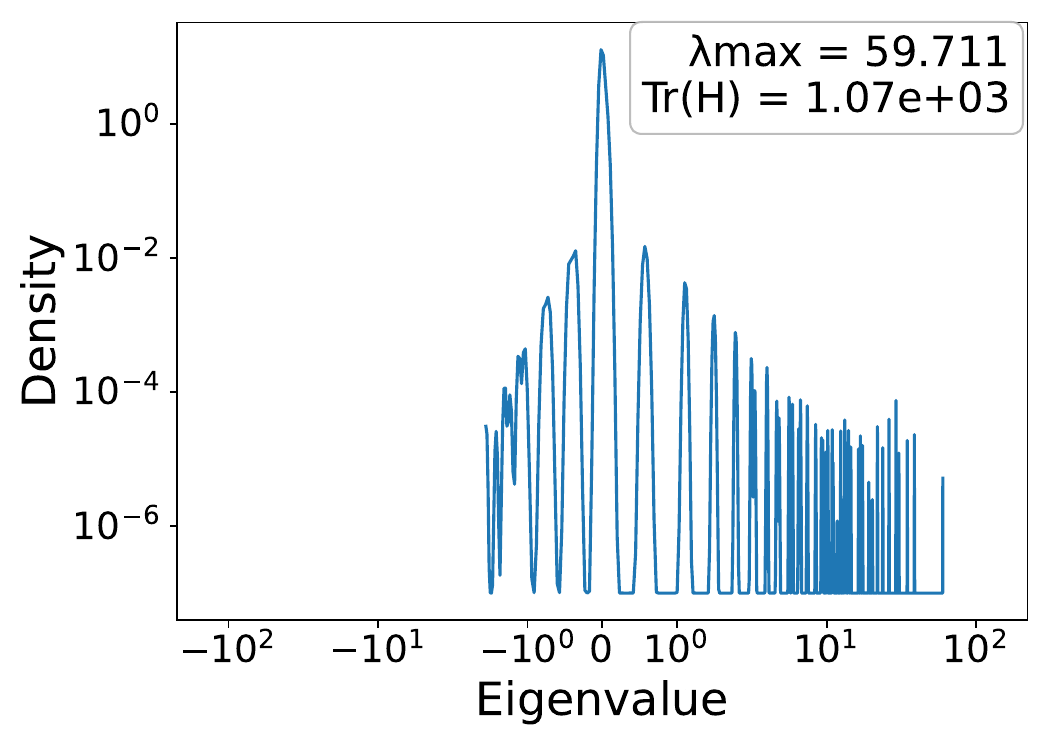}
        \caption{BCL: Head Classes}
        \label{fig:bcl_h}
    \end{subfigure}
    \hspace{0.04\textwidth}
    \begin{subfigure}[t]{0.28\textwidth}
        \centering
        \includegraphics[width=\linewidth]{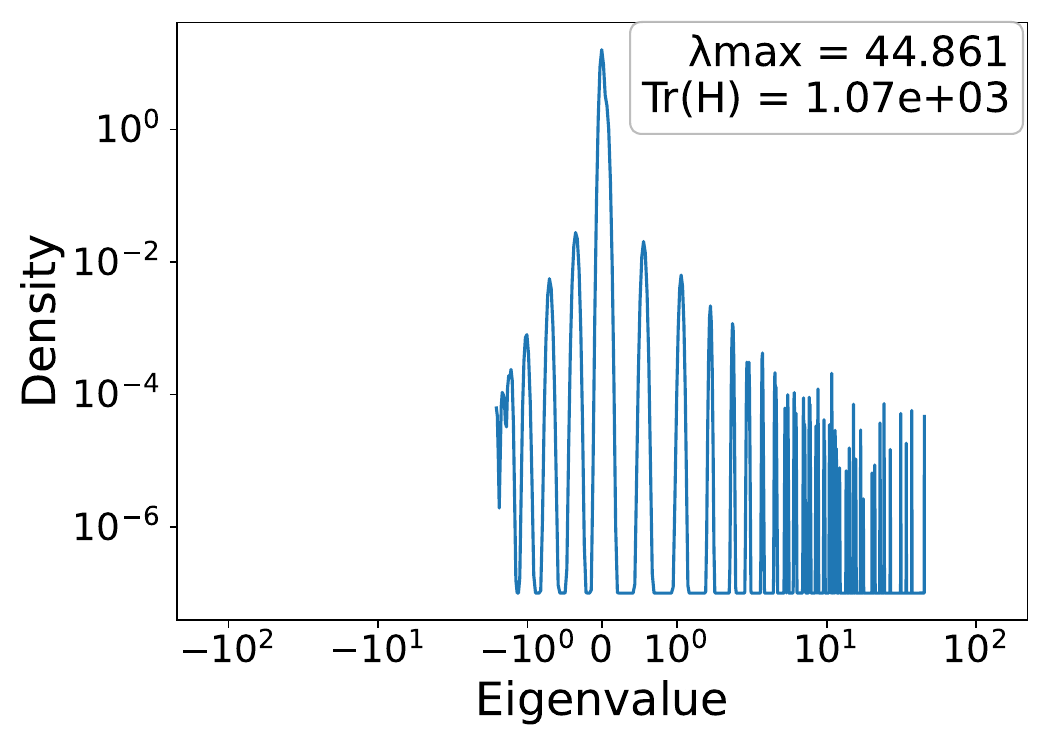}
        \caption{BCL+OFBD: Head Classes}
        \label{fig:our_h}
    \end{subfigure}

    \vspace{0.8em}

    \begin{subfigure}[t]{0.28\textwidth}
        \centering
        \includegraphics[width=\linewidth]{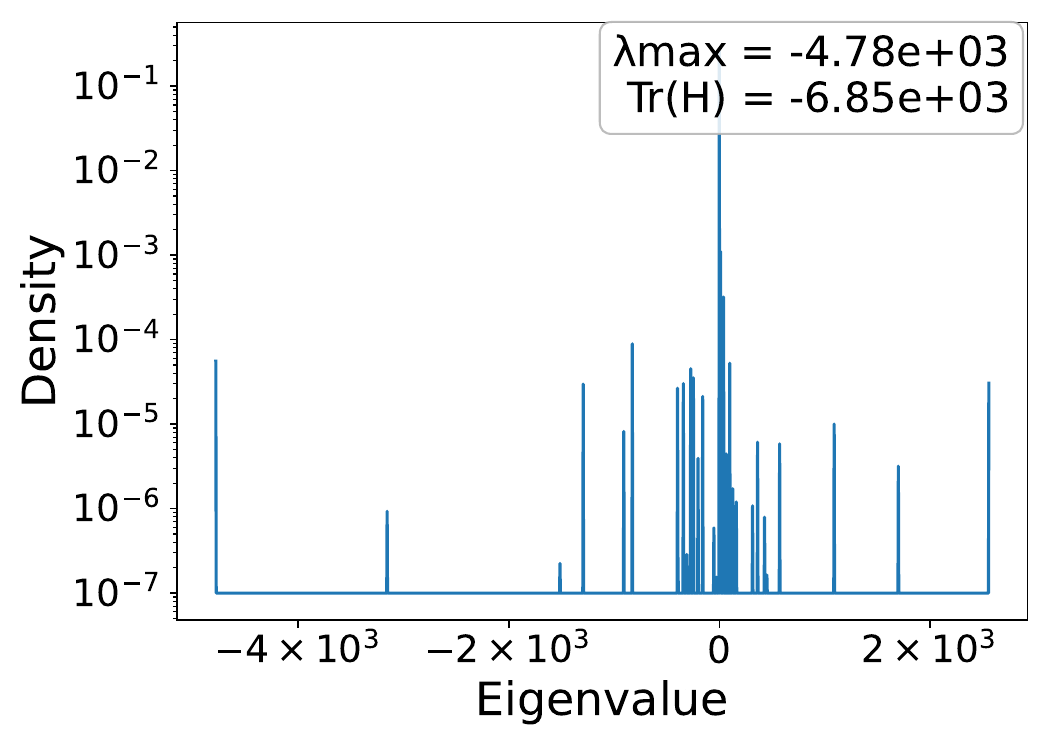}
        \caption{CE: Tail Classes}
        \label{fig:ce_t}
    \end{subfigure}
    \hspace{0.04\textwidth}
    \begin{subfigure}[t]{0.28\textwidth}
        \centering
        \includegraphics[width=\linewidth]{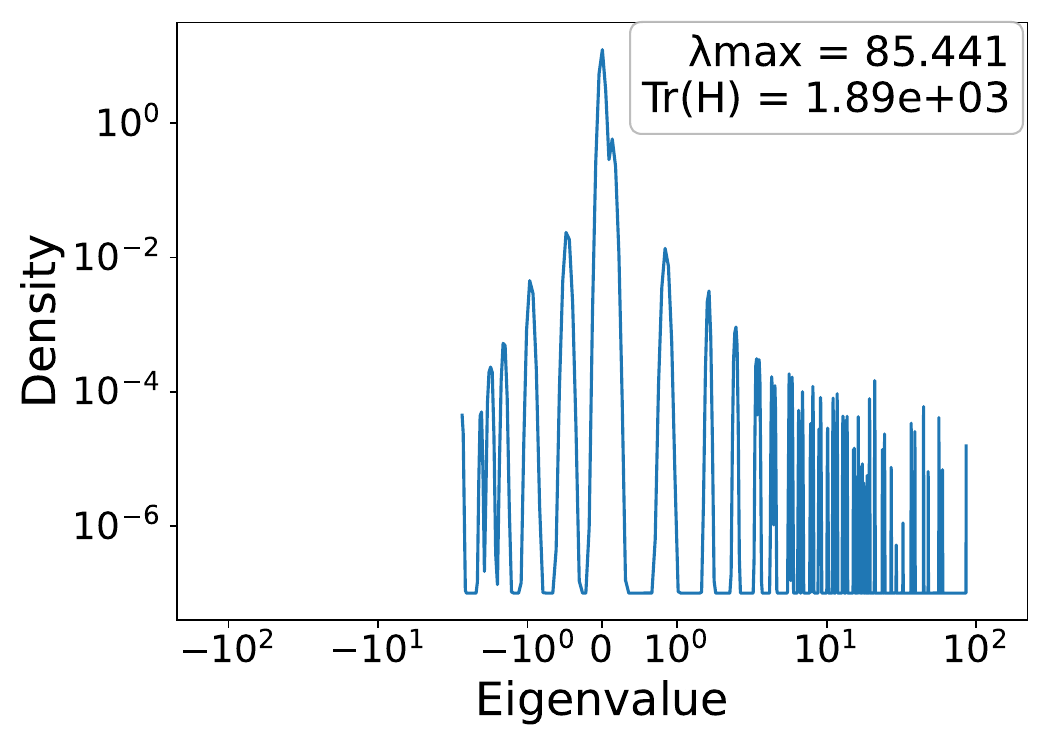}
        \caption{BCL: Tail Classes}
        \label{fig:bcl_t}
    \end{subfigure}
    \hspace{0.04\textwidth}
    \begin{subfigure}[t]{0.28\textwidth}
        \centering
        \includegraphics[width=\linewidth]{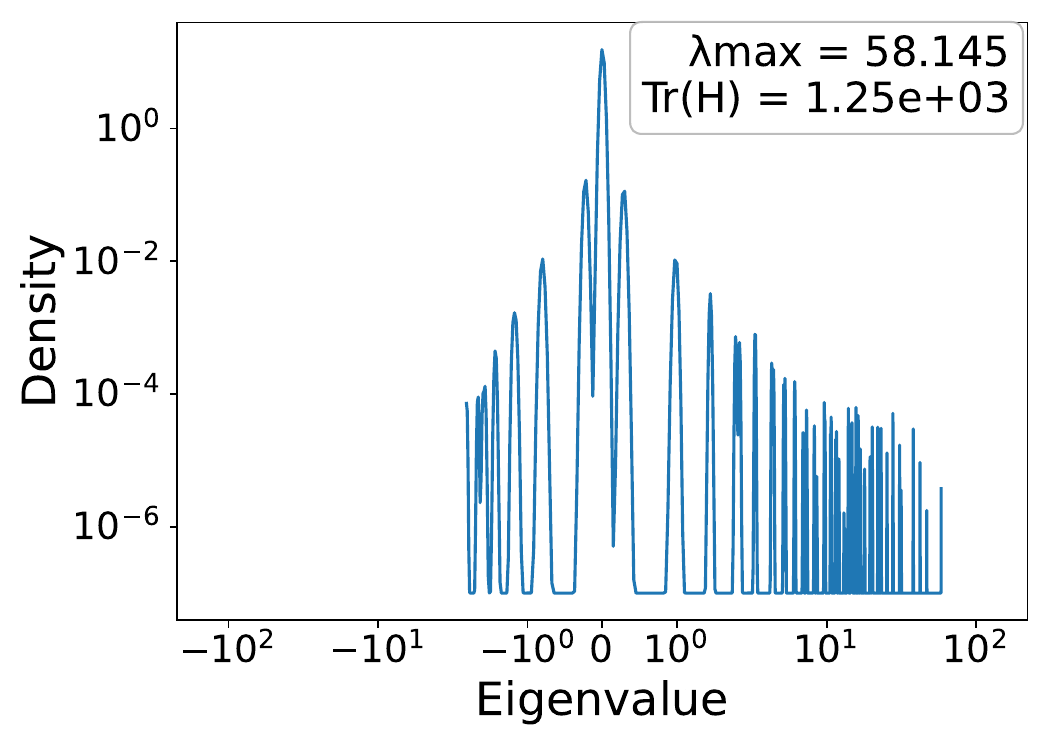}
        \caption{BCL+OFBD: Tail Classes}
        \label{fig:our_t}
    \end{subfigure}

    \caption{Eigen Spectral Density of Hessian for head and tail classes of ResNet models trained with CE, BCL, and BCL+OFBD on CIFAR-100 LT respectively. A smaller $\lambda_{\max}$ and $Tr(H)$ generally indicate a flatter loss landscape.}
    \label{fig:hessian}
\end{figure*}
\subsection{Additional Results for Eigen Spectral Density of Hessian}
This section presents additional results on the eigen spectral density of the Hessian for ResNet models trained with CE, BCL, and BCL+OFBD on CIFAR-100-LT. As shown in Fig.~\ref{fig:hessian}, we separately analyze the Hessian spectra of head and tail classes. A smaller largest eigenvalue $\lambda_{\max}$ and trace $\mathrm{Tr}(H)$ generally indicate a flatter loss landscape and better optimization stability.

Compared with the CE baseline, BCL tends to reduce the sharpness of the loss landscape to some extent, especially for tail classes, suggesting that contrastive learning helps improve the optimization behavior under long-tailed distributions. However, its Hessian spectrum still shows relatively large eigenvalues, indicating that the optimization landscape remains sharp for certain class groups. In contrast, BCL+OFBD consistently produces lower $\lambda_{\max}$ and $\mathrm{Tr}(H)$ for both head and tail classes. This suggests that BCL+OFBD effectively smooths the loss landscape, rather than only benefiting tail classes. The flatter Hessian spectrum further supports that BCL+OFBD mitigates background-biased optimization and leads to more stable learning under long-tailed training.

\subsection{Visualization of LossLandscape}
\begin{figure*}[t]
    \centering

    \begin{subfigure}[t]{0.28\textwidth}
        \centering
        \includegraphics[width=\linewidth]{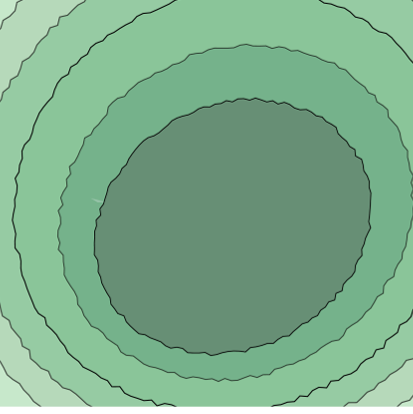}
        \caption{CE: Head Classes}
        \label{fig:loss_ce_h}
    \end{subfigure}
    \hspace{0.04\textwidth} 
    \begin{subfigure}[t]{0.28\textwidth}
        \centering
        \includegraphics[width=\linewidth]{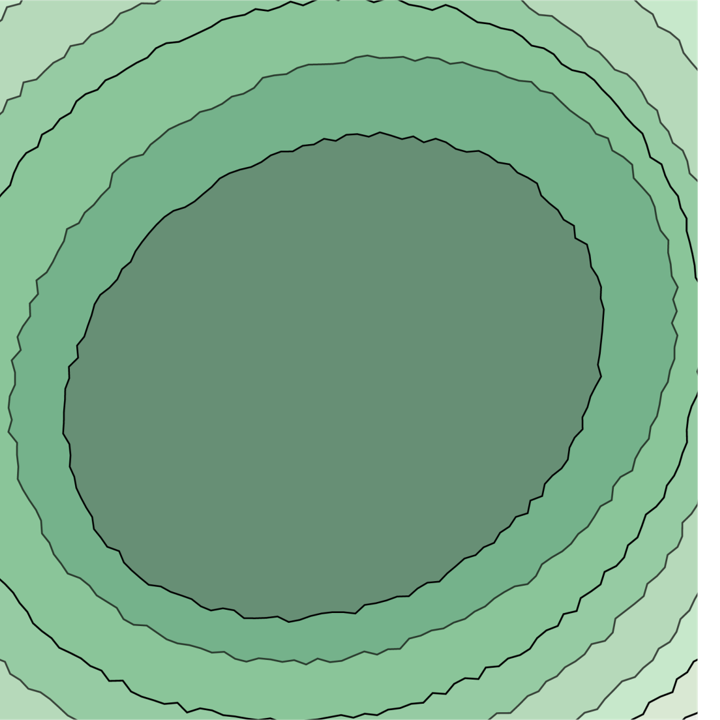}
        \caption{BCL: Head Classes}
        \label{fig:loss_bcl_h}
    \end{subfigure}
    \hspace{0.04\textwidth}
    \begin{subfigure}[t]{0.28\textwidth}
        \centering
        \includegraphics[width=\linewidth]{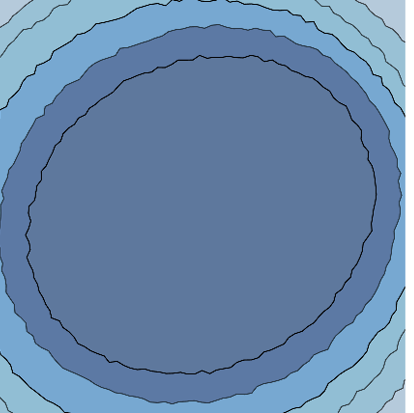}
        \caption{BCL+OFBD: Head Classes}
        \label{fig:loss_our_h}
    \end{subfigure}

    \vspace{0.8em}

    \begin{subfigure}[t]{0.28\textwidth}
        \centering
        \includegraphics[width=\linewidth]{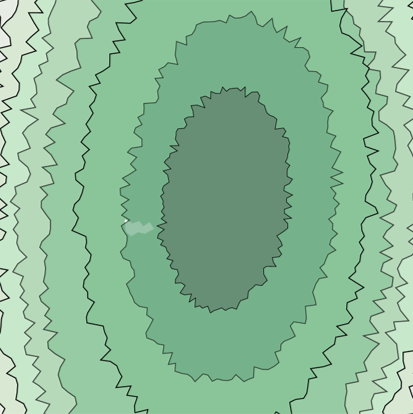}
        \caption{CE: Tail Classes}
        \label{fig:loss_ce_t}
    \end{subfigure}
    \hspace{0.04\textwidth}
    \begin{subfigure}[t]{0.28\textwidth}
        \centering
        \includegraphics[width=\linewidth]{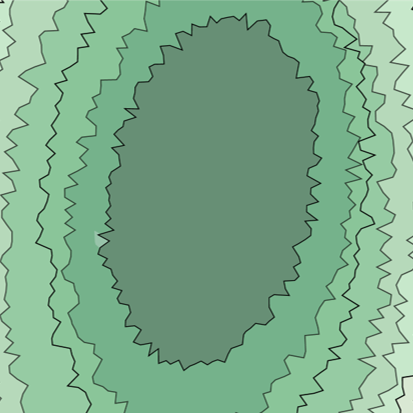}
        \caption{BCL: Tail Classes}
        \label{fig:loss_bcl_t}
    \end{subfigure}
    \hspace{0.04\textwidth}
    \begin{subfigure}[t]{0.28\textwidth}
        \centering
        \includegraphics[width=\linewidth]{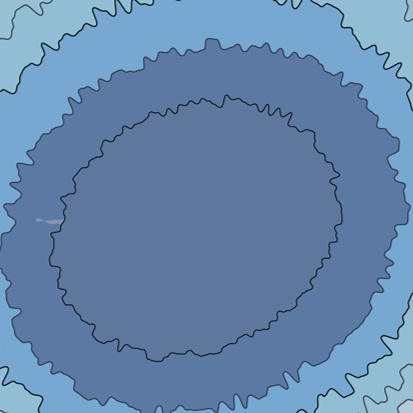}
        \caption{BCL+OFBD: Tail Classes}
        \label{fig:loss_our_t}
    \end{subfigure}

    \caption{Visualization of loss landscape for head and tail classes of ResNet models trained with CE, BCL, and BCL+OFBD on CIFAR-100 LT respectively.}
    \label{fig:LossLand}
\end{figure*}

Fig.~\ref{fig:LossLand} visualizes the loss landscapes of head and tail classes for ResNet models trained with CE, BCL, and BCL+OFBD on CIFAR-100-LT. 
The first row corresponds to head classes, and the second row corresponds to tail classes. 
Compared with CE and BCL, BCL+OFBD tends to produce a flatter and smoother loss landscape for both class groups. 
Such a tendency is more pronounced for tail classes, where the loss landscape under long-tailed training is usually sharper and more irregular. 
The results indicate that OFBD helps stabilize the optimization process and mitigates the unfavorable optimization behavior caused by data imbalance.

\begin{table*}[t]
\centering
\scriptsize
\renewcommand{\arraystretch}{1.05}


\begin{minipage}[t]{0.49\textwidth}
\centering

\captionof{table}{Comparison with attention modules on CIFAR-100-LT ($r=100$).}
\label{tab:attention_comparison}

\setlength{\tabcolsep}{3pt}

\begin{tabular}{lcccc}
\toprule
\textbf{Method}
&
\textbf{Overall}$\uparrow$
&
\textbf{Many}$\uparrow$
&
\textbf{Med.}$\uparrow$
&
\textbf{Few}$\uparrow$
\\
\midrule

CBAM~\cite{woo2018cbam}
&50.43&67.80&48.90&31.90\\

BAM~\cite{park2018bam}
&50.88&68.70&49.30&32.00\\

Coordinate Attention~\cite{hou2021coordinate}
&50.83&\textbf{69.40}&49.10&31.30\\

Triplet Attention~\cite{zhou2021triplet}
&50.67&67.90&50.00&31.40\\

\midrule

\rowcolor{gray!18}
BFR (Ours)
&\textbf{52.97}
&66.96
&\textbf{53.65}
&\textbf{35.86}\\

\bottomrule
\end{tabular}

\end{minipage}
\hfill
\begin{minipage}[t]{0.49\textwidth}
\centering

\captionof{table}{Robustness of $B_u$ under input perturbations.}
\label{tab:bu_robustness}

\setlength{\tabcolsep}{5pt}

\begin{tabular}{lcc}
\toprule
\textbf{Input}
&
\textbf{BG AUROC}$\uparrow$
&
\textbf{$\Delta$ vs. Clean}
\\
\midrule

Clean
&0.7182&0.0000\\

Brightness, factor 1.2
&0.7054&-0.0128\\

Contrast, factor 1.2
&0.7051&-0.0131\\

Gaussian noise, std 0.05
&0.7038&-0.0144\\

Gaussian blur, $k=5,\sigma=1.0$
&0.7016&-0.0166\\

\bottomrule
\end{tabular}

\end{minipage}

\vspace{2mm}


\begin{minipage}[t]{0.49\textwidth}
\centering

\captionof{table}{Integration with additional long-tailed methods.}
\label{tab:integration_methods}

\setlength{\tabcolsep}{4pt}

\begin{tabular}{lcccc}
\toprule

\textbf{Method}
&
\textbf{Overall}$\uparrow$
&
\textbf{Many}$\uparrow$
&
\textbf{Med.}$\uparrow$
&
\textbf{Few}$\uparrow$
\\
\midrule

MDCS~\cite{zhao2023mdcs}
&53.15&67.09&\textbf{56.09}&33.47\\

\rowcolor{gray!18}
MDCS + OFBD
&\textbf{54.39}
&\textbf{68.31}
&55.26
&\textbf{37.14}\\

\midrule

SADE~\cite{zhang2022self}
&49.20
&\textbf{61.31}
&51.14
&32.80\\

\rowcolor{gray!18}
SADE + OFBD
&\textbf{51.01}
&57.89
&\textbf{53.49}
&\textbf{40.10}\\

\bottomrule
\end{tabular}

\end{minipage}
\hfill
\begin{minipage}[t]{0.49\textwidth}
\centering

\captionof{table}{Feature-layer analysis for the Region Selector.}
\label{tab:selector_layer}

\setlength{\tabcolsep}{2.2pt}

\begin{tabular}{lcccccc}
\toprule
\textbf{Layer}
&
\textbf{Res.}
&
\textbf{S-FG}$\uparrow$
&
\textbf{S-BG}$\downarrow$
&
\textbf{AUROC}$\uparrow$
&
\textbf{Overall}$\uparrow$
&
\textbf{Few}$\uparrow$
\\
\midrule

layer1
&32$\times$32
&0.1618
&0.1441
&0.7285
&53.98
&41.00\\

layer2
&16$\times$16
&0.1694
&0.1280
&0.7375
&54.04
&40.97\\

\rowcolor{gray!18}
layer3
&8$\times$8
&\textbf{0.1709}
&\textbf{0.1026}
&\textbf{0.8644}
&\textbf{54.12}
&\textbf{41.15}\\

\bottomrule
\end{tabular}

\end{minipage}

\vspace{2mm}


\begin{minipage}[t]{0.49\textwidth}
\centering

\captionof{table}{Sensitivity to the number of selected proposals $K$.}
\label{tab:proposal_number}

\setlength{\tabcolsep}{10pt}

\begin{tabular}{cc}
\toprule
\textbf{$K$}
&
\textbf{Overall Acc. (\%)}$\uparrow$
\\
\midrule

2 & 53.87 $\pm$ 0.34\\
3 & 53.93 $\pm$ 0.22\\
4 & 53.94 $\pm$ 0.20\\
5 & 53.89 $\pm$ 0.19\\

\rowcolor{gray!18}
6 & \textbf{53.96 $\pm$ 0.19}\\

\bottomrule
\end{tabular}

\end{minipage}
\hfill
\begin{minipage}[t]{0.49\textwidth}
\centering

\captionof{table}{Masking analysis based on background-aware score $B_u$.}
\label{tab:bu_masking}

\setlength{\tabcolsep}{8pt}

\begin{tabular}{lcc}
\toprule
\textbf{Masking}
&
\textbf{Overall Acc. (\%)}$\uparrow$
&
\textbf{$\Delta$ Acc.}
\\
\midrule

None
&58.475
&0.000\\

Random 20\%
&53.850
&-4.625\\

High-$B_u$ 20\%
&56.655
&-1.820\\

\rowcolor{gray!18}
Low-$B_u$ 20\%
&45.140
&\textbf{-13.335}\\

\bottomrule
\end{tabular}

\end{minipage}

\end{table*}

\subsection{Comparison with Attention Modules}

We compare BFR with representative attention modules under the same BCL-based setting on CIFAR-100-LT ($r=100$). As shown in Table~\ref{tab:attention_comparison}, BFR achieves better overall and Few-shot performance, demonstrating its effectiveness in background bias.

\subsection{Robustness of Background Score under Input Perturbations}

We evaluate the robustness of $B_u$ under different image perturbations on ImageNet-LT. As shown in Table~\ref{tab:bu_robustness}, BG AUROC remains above 0.70 across all perturbations, indicating stable background-aware estimation under appearance variations.

\subsection{Integration with Multi-Expert Long-tailed Recognition Frameworks}

We integrate OFBD into representative multi-expert long-tailed recognition frameworks while keeping their original architectures and training objectives unchanged. As shown in Table~\ref{tab:integration_methods}, OFBD consistently improves different multi-expert baselines, demonstrating its compatibility and effectiveness across existing long-tailed recognition frameworks.

\subsection{Region Selector Feature Layer Analysis}

We analyze the effect of different feature layers used by the Region Selector. As shown in Table~\ref{tab:selector_layer}, deeper layers provide stronger foreground-background separation, and layer3 achieves the best discrimination capability.

\subsection{Sensitivity Analysis of Region Proposal Number}

We evaluate the sensitivity of $K$ in the Region Selector. As shown in Table~\ref{tab:proposal_number}, OFBD maintains stable performance across different $K$ values, demonstrating robustness to proposal selection.

\subsection{Masking Analysis of Background-aware Score}

We perform masking experiments based on $B_u$ to analyze the semantic meaning of high-score regions. As shown in Table~\ref{tab:bu_masking}, masking high-$B_u$ regions causes limited degradation, while masking low-$B_u$ regions leads to significant performance drops, validating the effectiveness of BFR.

\section{Limitation}
\label{sec:limitations}

A key consideration of OFBD lies in its reliance on accurate foreground localization and feature separation. While our current FG-CutMix and BFR designs work effectively on standard visual benchmarks, scenarios with severe occlusion, extremely small objects, or highly entangled foreground-background cues may benefit from enhanced region selection strategies or adaptive rectification. Addressing these cases represents a natural direction for extending OFBD, highlighting opportunities for future work without undermining its demonstrated effectiveness on long-tailed datasets.

\end{document}